\documentclass{article}
\usepackage{jair, rawfonts, theapa}

\jairheading{67}{2020}{}{08/2019}{02/2020}
\ShortHeadings{Compositionality decomposed: how do neural networks generalise?}{Hupkes, Dankers, Mul \& Bruni}
\firstpageno{1}

\usepackage[hidelinks]{hyperref}
\usepackage[dvipsnames]{xcolor}
\usepackage{array}
\usepackage{booktabs}
\usepackage{multirow}
\usepackage{enumitem}
\usepackage{graphicx}
\usepackage[round, authoryear]{natbib}
\usepackage{subcaption}
\usepackage{float}
\usepackage{tikz}
\usepackage{tikz-qtree}
\usepackage{xspace}
\usepackage{amsmath}
\usetikzlibrary{shapes, arrows}

\newcommand{\pcfg}[0]{PCFG SET\xspace}
\newcommand{\pms}[1]{\footnotesize $\pm$ #1}

\definecolor{green_}{RGB}{179,226,205}
\definecolor{orange_}{RGB}{253,205,172}
\definecolor{blue_}{RGB}{203,213,232}
\definecolor{pink_}{RGB}{244,202,228}
\definecolor{yellow_}{RGB}{255,242,174}
\def\radius{0.4cm}

\title{Compositionality decomposed: how do \\ neural networks generalise?}

\author{\name Dieuwke Hupkes \email dieuwkehupkes@gmail.com \\
        \addr Institute for Logic, Language and Computation\\
        University of Amsterdam\\
        ~\\
        \name Verna Dankers \email vernadankers@gmail.com \\
        \name Mathijs Mul \email mathijsmul@gmail.com \\
        \addr University of Amsterdam\\
        ~\\
        \name Elia Bruni \email elia.bruni@gmail.com \\
        \addr University of Pompeu Fabra
}

\begin{document}

\maketitle

\begin{abstract}
Despite a multitude of empirical studies, little consensus exists on whether neural networks are able to generalise \emph{compositionally}, a controversy that, in part, stems from a lack of agreement about what it means for a neural model to be compositional.
As a response to this controversy, we present a set of tests that provide a bridge between, on the one hand, the vast amount of linguistic and philosophical theory about compositionality of language and, on the other, the successful neural models of language.
We collect different interpretations of compositionality and translate them into five theoretically grounded tests for models that are formulated on a task-independent level.
In particular, we provide tests to investigate (i) if models systematically recombine known parts and rules %(\emph{systematicity}) 
(ii) if models can extend their predictions beyond the length they have seen in the training data % (\emph{productivity}) 
(iii) if models' composition operations are local or global % (\emph{localism}) 
(iv) if models' predictions are robust to synonym substitutions and % (\emph{substitutivity}) and 
(v) if models favour rules or exceptions during training. %(\emph{overgeneralisation}).
To demonstrate the usefulness of this evaluation paradigm, we instantiate these five tests on a highly compositional data set which we dub \pcfg and apply the resulting tests to three popular sequence-to-sequence models: a recurrent, a convolution-based and a transformer model.
We provide an in-depth analysis of the results, which uncover the strengths and weaknesses of these three architectures and point to potential areas of improvement.
\end{abstract}

\section{Introduction}

The advancements of distributional semantics of the word level allowed the field of natural language processing to move from discrete mathematical methods to models that use continuous numerical vectors \citep[see, e.g.][]{clark2015vector,erk2012vector,turney2010frequency}.
Such continuous vector representations operationalise the distributional semantics hypothesis -- stating that semantically similar words have similar contextual distributions \citep[e.g.][]{miller1991contextual} -- by keeping track of contextual information from large textual corpora. 
They can then act as surrogates for word meaning and be used, for example, to quantify the degree of semantic similarity between words, by means of simple geometric operations \citep{clark2015vector}.
Words represented in this way can be an integral part of the computational pipeline and have proven to be useful for almost all natural language processing tasks \citep[see, e.g.][]{hirschberg2015advances}.

After the introduction of continuous word representations, a logical next step involved understanding how to \textit{compose} these representations to obtain representations for phrases, sentences and even larger pieces of discourse. 
Some early approaches to do so stayed close to formal symbolic theories of language and sought to explicitly model semantic composition by finding a composition function that could be used to combine word representations. 
The adjective-noun compound \emph{blue sky}, for instance, would be represented as a new vector resulting from the composition of the representations for \emph{blue} and \emph{sky}. 
Examples of such composition functions are as simple as vector addition and (point-wise) multiplication \citep[e.g.][]{mitchell2008vector} up to more powerful tensor-based operations \citep{smolensky1990tensor,plate1991holographic} where, for instance, the adjective \emph{blue} would be represented as a matrix, which would be multiplied with the noun vector \emph{sky} to return the representation for \emph{blue sky} \citep[e.g.][]{baroni2010nouns,coecke2010mathematical}.\footnote{For a more complete overview and an analysis of such approaches in the light of formal semantics, we refer to \citet{kartsaklis2014compositional} and \citet{boleda2016formal}, respectively.}

A more recent trend in word composition exploits \textit{deep learning}, a class of machine learning techniques that model language in a completely data-driven fashion, by defining a loss on a down-stream task (such as sentiment analysis, language modelling or machine translation) and learning the representations for larger chunks from a signal back-propagated from this loss. 
In terms of how they compose representations, models using deep learning can be divided into roughly two categories.

In the first category, deep learning is exploited to learn only the actual composition functions, while the order of composition is defined by the modeller.  
An example is the \emph{recursive neural network} of \citet{socher2010learning}, in which representations for larger chunks are computed recursively following a predefined syntactic parse tree of the sentence.
While the composition function in this approach is fully learned from data using \emph{back-propagation through structure} \citep{goller1996learning}, the tree structure that defines the order of application has to be provided to the model, allowing models to be `compositional by design'.
More recent variants lift this dependency on external parse trees by jointly learning the composition function and the parse tree \citep[][i.a.]{le2015forest,kim2019unsupervised}, often at the cost of computational feasibility.

In the second type of deep learning models, no explicit notion of (linguistic) trees or arbitrary depth hierarchy is entertained. 
Earlier models of this type deal with language processing sequentially and use recurrent processing units such as LSTMs \citep{hochreiter1997long} and GRUs \citep{chung2014empirical} at their core \citep{sutskever2014sequence} or are based on convolutional networks \citep{kalchbrenner2014convolutional}.
An important contribution to their effectiveness comes from attention mechanisms, which allow recurrent models to keep track of long-distance dependencies more effectively \citep{bahdanau2014neural}. 
More recently, these models went all-in on attention, abandoning sequential processing in favour of massively distributed sequence processing all based on attention \citep{vaswani2017attention}.
While the architectural design of this class of models is not motivated by knowledge about linguistics or human processing, they are -- through their ability to easily process very large amounts of data -- more successful than the previously mentioned (sub)symbolic models on a variety of natural language processing tasks. 

Different types of models that compose smaller representations into larger ones can be compared along many dimensions.
Commonly, they are evaluated by the usefulness of their representations for different types of tasks, but also scalability, how much data they need to develop their representations (sample efficiency), and their computational feasibility play a role in their evaluation.
It remains, however, difficult to explicitly assess if the composition functions they implement are appropriate for natural language and, importantly, to what extent they are in line with the vast amount of knowledge and theories about semantic composition from formal semantics and (psycho)linguistics.
While the composition functions of symbolic models are easy to understand (because they are defined on a mathematical level), it is not empirically established that their rigidity is appropriate for dealing with the noisiness and complexity of natural language \citep[e.g.][]{potts2019case}.
Neural models, on the other hand, seem very well up to handling noisy scenarios but are often argued to be fundamentally incapable of conducting the types of compositions required to process natural language \citep[for more information on this debate, see][]{pinker1984language, fodor1988connectionism, smolensky1990tensor, marcus2003algebraic} or at least to not use those types of compositions to solve their tasks \citep[e.g.][]{lake2018generalization}.

In this work, we consider the latter type of models and focus in particular on whether these models are capable of learning \emph{compositional} solutions, a question that recently, with the rise of the success of such models, has attracted the attention of several researchers.
While many empirical studies can be found that explore the compositional abilities of neural models, they have not managed to convince the community of either side of the debate: whether neural networks are able to learn and behave compositionally is still an open question.
One issue standing in the way of more clarity on this matter is that different researchers have different interpretations of what exactly it means to say that a model is or is not compositional, a point exemplified by the vast number of different tests that exist for compositionality.
Some studies focused on testing if models are able to productively use symbolic \textit{rules} \citep[e.g.][]{lake2018generalization}; 
Some instead consider models' ability to process \textit{hierarchical} structures \citep{hupkes2018diagnostic, linzen2016assessing};
Yet others consider if models can segment the input into reusable parts \citep{johnson2017clevr}.

This variety of tests for compositionality of neural networks existing in the literature is better understandable considering the open nature of the principle of compositionality, by \citet{partee1995lexical} phrased as ``\emph{The meaning of a whole is a function of the meanings of the parts and of the way they are syntactically combined}''.
While there is ample support for this principle, there is less consensus about its exact interpretation and practical implications.
One important reason for this is that the principle is not theory-neutral: it requires a theory of both syntax and meaning, as well as functions to determine the meaning of composed parts.
Without these components, the principle of compositionality is formally vacuous \citep{janssen1983, zadrozny1994compositional}, because also trivial and intuitively non-compositional solutions that cast every expression as one part and assign it a meaning as a whole do not formally violate the principle of compositionality.
Furthermore, the principle of compositionality concerns the compositionality of \emph{language} but does not specify what it means for a language \emph{user} or \emph{model} to be compositional.
Can a model be called compositional when it can represent a compositional language? Are there any restrictions on how it has to do so?
To empirically test models for compositionality it is necessary to first establish \textit{what} is to be considered compositional behaviour.
With this work, we aim to contribute to clarity on this point, by presenting a study in which we collect different aspects of and intuitions about compositionality of language from linguistics and philosophy and translate them into concrete tests that  
can be used to better understand the composition functions learned by neural models trained end-to-end on a downstream task.

The contribution of our work, we believe, is three-fold.
First, we provide a bridge between, on the one hand, the vast amount of theory about compositionality that underpins symbolic models of language and semantic composition and, on the other hand, the neural models of language that have proven to be very effective in many natural language tasks that seem to require compositional capacities.
Importantly, we do not aim to provide a new definition of compositionality, but rather we identify different components of compositionality within the literature and design behavioural tests that allow testing for these components independently. 
We believe that the field will profit from such a principled analysis of compositionality and that this analysis will provide clarity concerning the different interpretations that may be entertained by different researchers.
A division into clearly understood components can help to identify and categorise the strengths and weaknesses of different models.
We provide concrete and usable tests, bundled in a versatile test suite that can be applied to any kind of model.

Secondly, to demonstrate the usefulness of this test suite, we apply our tests to three popular sequence-to-sequence models: a recurrent, a convolution-based and a transformer model.
We provide an in-depth analysis of the results, uncovering interesting strengths and weaknesses of these three architectures.

Lastly, we touch upon the complex question that concerns the extent to which a model needs to be explicitly compositional to adequately model data of which the underlying structure is, or seems, compositional.
We believe that, in a time where the most successful natural language processing methods require large amounts of data and are not directly motivated by linguistic knowledge or structure, this question bears more relevance than ever. 

\paragraph{Outline}
In what follows, we first briefly revise other literature with similar aims and sketch how our work stands apart from previous attempts to assess the extent to which networks implement compositionality (Section~\ref{sec:related_work}).
We describe previously proposed data sets to evaluate compositionality as well as studies that evaluate the representations of pre-trained models.
In Section~\ref{sec:compositionality_tests}, we give a theoretical explanation of the five notions for which we devise tests, and we propose how to behaviourally test for them.
In Section~\ref{sec:data}, we describe the data set that we use for our study, followed by a brief description of the three types of architectures that we compare in our experiments in Section~\ref{sec:models}.
We then detail our experiments and report and analyse their results in Section~\ref{sec:results} and further reflect upon their implications in Section~\ref{sec:discussion}.

\section{Related work}\label{sec:related_work}

Whether artificial neural networks are fundamentally capable of representing compositionality, trees and hierarchical structure has been a prevalent topic ever since the first connectionism models for natural language were introduced.
Recently, this topic has regained attention, and a substantial number of empirical studies can be found that explore the compositional abilities of neural models, with a specific focus on their ability to represent \textit{hierarchy}.
These studies can be roughly divided into two categories: studies that devise specific data sets that models can be trained and tested on to assess if they behave compositionally, and studies that focus on assessing the representations that are learned by models trained on some independent (often natural) data set.

\subsection{Evaluating compositionality with artificial data}\label{subsec:comp_datasets}

Specifically crafted, artificial data sets to evaluate compositionality are typically generated from an underlying grammar.
It is then assumed that models can only find the right solution to the test set if they learned to interpret the training data in a compositional fashion.
Below, we discuss a selection of such data sets and briefly review their results.\footnote{
Discussing in detail all different data sets that have been proposed to evaluate compositionality in neural networks falls outside the scope of this paper.
We aimed to make a representative selection of studies, using as a criterion that they should involve sequential inputs and explicitly mention compositionality. 
We excluded \textit{grounded} data sets such as CLEVR \citep{johnson2017clevr} and SQOOP \citep{bahdanau2018systematic}, which contain more than one modality.
Furthermore, we did not include studies whose primary focus is on \emph{how} neural networks implement compositional structures \citep{lakretz2019emergence,giulianelli2018under,mccoy2019rnns,soulos2019discovering,weiss2018extracting} or studies that evaluate compositionality only based on models' representations \citep{andreas2019measuring}.}

\subsubsection{Arithmetic language and mathematical reasoning}
One of the first (recent) data sets proposed as a testbed to reveal how neural networks process hierarchical structure is the \textit{arithmetic language}, introduced by \citet{veldhoen2016diagnostic}.
\citet{veldhoen2016diagnostic} test networks for algebraic compositionality by looking at their ability to process spelled out, nested arithmetic expressions.
In a follow-up paper, to gain insight into the types of solution that networks encode, the same authors introduce \emph{diagnostic classifiers}, trained to fire for specific strategies used to solve the problem.
They show that simple recurrent networks do not perform well on the task, but gated recurrent networks can generalise well to lengths and depths of arithmetic expressions that were not in the training set, although their performance quickly deteriorates when the length of expressions grows \citep{hupkes2018diagnostic}.\footnote{\citet{zaremba2014learning} also used a task based on arithmetics, which requires learning to execute \emph{computer programs}, which they use to compare different learning curricula.}
From this, they conclude that these models are -- to some extent -- able to capture the underlying compositional structure of the data.

More recently, \cite{saxton2019analysing} released another data set in which maths was used to probe the compositional generalisation skills of neural networks.
\citet{saxton2019analysing} compare transformers and LSTM-based architectures trained on a data set with mathematical questions and find that the transformer models generalise better than the LSTM models.
Specifically, transformers outperform LSTMs on a set of extrapolation tests that require compositional skills such as generalising to questions involving larger numbers, more numbers or more compositions.
However, performance deteriorates for questions that require the computation of intermediate values, which \citet{saxton2019analysing} reason indicates that the model has not truly learned to treat the task in a compositional manner but instead applies shallow tricks.

\subsubsection{SCAN}
In 2018, \citeauthor{lake2018generalization} proposed the SCAN data set, describing a simple navigation task that requires an agent to execute commands expressed in a compositional language. 
The authors test various sequence-to-sequence models on three different splits of the data: a random split, a split testing for longer action sequences and a split that requires compositional application of words learned in isolation. 
The models obtain almost perfect accuracy on the first split while performing very poorly on the last two, which the authors argue require a compositional understanding of the task.  
They conclude that -- after all these years -- sequence-to-sequence recurrent networks are still not \emph{systematic}.
In a follow-up paper by \cite{loula2018rearranging}, the same authors criticise these findings and propose a new set of splits which focuses on rearranging familiar words (i.e. ``jump'', ``right'' and ``around'') to form novel meanings (``jump around right'')
. 
Although they collect considerably more evidence for systematic generalisation within their amended setup, the authors confirm their previous findings that the models do not learn compositionally.
Very recently, SCAN was also used to diagnose convolutional networks.
Comparing to recurrent networks, \cite{dessi2019cnns} find that convolutional networks exhibit improved compositional generalisation skills but their errors are unsystematic, indicating that the model did not fully master any of the systematic rules.

\subsubsection{Lookup tables}
\cite{livska2018memorize} introduce a minimal compositional test where neural networks need to apply function compositions to correctly compute the meaning of sequences of lookup tables. 
The meanings of these lookup tables are exhaustively defined and presented to the model, so that applying them does not require more than rote memorisation.
The authors show that out of many models trained with different initialisations only a very small fraction exhibits compositional behaviour, while the vast majority does not.\footnote{\citet{hupkes2018learning} show how adding an extra supervision signal to the network's attention consistently results in a complete solution of the task, but it is not clear how their results extend to other, more complicated scenarios. \citet{korrel2019transcoding} propose a novel architecture with analogous, complete solutions without the need for extra supervision.}
\subsubsection{Logical inference}
\citet{bowman2015tree} propose a data set which uses a slightly different setup: they assess models' compositional skills by testing their ability to infer logical entailment relations between pairs of sentences in an artificial language.
The grammar they use licenses short, simple sentences; the relations between these sentences are inferred using a natural logic calculus that acts directly on the generated expressions.
\citet{bowman2015tree} show that recursive neural networks, which recursively apply the same composition function and are thus compositional by design, obtain high accuracies on this task.
\citet{mul2019siamese} show that also gated recurrent models can perform well on an adapted version of the same task, which uses a more complex grammar.
With a series of additional tests, \citet{mul2019siamese} provide further proof for basic compositional generalisation skills of the best-performing recurrent models.
\citet{tran2018importance} report similar findings, and furthermore show that while a transformer performs similar to an LSTM model when the entire data set is used, an LSTM model generalises better when smaller training data is used.

\subsection{Evaluating compositionality with natural data}\label{subsec:comp_eval}

While very few studies present methods to explicitly evaluate how compositional the representations of models that are trained on independent data sets are, there are several studies that focus on evaluating aspects of such models that are related to compositionality.
In particular, starting from the seminal work of \citet{linzen2016assessing}, the evaluation of the syntactic capabilities of neural \emph{language models} has attracted a considerable amount of attention.
While the explicit focus of such studies is on the syntactic capabilities of different models and not on providing tests for compositionality, many of the results in fact concern the way that neural networks process the types of hierarchical structures often assumed to underpin compositionality.\footnote{In fact, there are quite a few earlier studies relating to the ability of neural networks to implement grammatical structure that consider a similar paradigm, albeit using artificial languages.
Such studies consider how well neural networks can represent formal languages generated by grammars from different classes of the Chomsky Hierarchy \citep[e.g.][]{elman1991distributed,christiansen1999toward, rodriguez2001simple,wiles1995learning, rodriguez1999recurrent, batali1994artificial,weiss2018practical}.
Like the studies with natural language described in this chapter, these studies focus on rules and hierarchical structure but do not specifically target compositionality, which requires not only syntax but also meaning.}

\subsubsection{Number agreement} 
\citet{linzen2016assessing} propose to test the syntactic abilities of LSTMs by testing to what extent they are capable of correctly processing long-distance subject-verb agreement, a phenomenon they argue to be commonly regarded as evidence for hierarchical structure in natural language.
They devise a \emph{number-agreement} task and find that a pre-trained state-of-the-art LSTM model \citep{jozefowicz2016exploring} does not capture the structure-sensitive dependencies.

Later, these results were contested by a different research group, who repeated and extended the study with a different language model and tested a number of different long-distance dependencies for English, Italian, Hebrew and Russian \citep{gulordava2018colorless}.
Their results do not match the findings of the earlier study: \cite{gulordava2018colorless} find that an LSTM language model can solve the subject-verb agreement problem well, even when the words in the sentence are replaced by syntactically nonsensical words, which they take as evidence that the model is indeed relying on syntactic and not semantic clues.\footnote{The task proposed by \citet{linzen2016assessing} served as inspiration for many studies investigating the linguistic or syntactic capabilities for neural language models, and also the task itself was used in many follow-up studies. 
Such studies, which we will not further discuss, are generally positive about the extent to which recurrent language models represent syntax.}
Whether the very recent all-attention language models do also capture syntax-sensitive dependencies is still an open question. Some (still unpublished) studies find evidence that such models score high on the previously described number-agreement task  \citep{goldberg2019assessing,lin2019open}. 
More mixed results are reported by others \citep{tran2018importance,wolf2019some}.

\subsubsection{Syntax in machine translation}

The subfield of natural language processing that is most related to ours in terms of setup is the field of machine translation (MT).
There are little detailed studies concerning the compositional behaviour of neural MT models but many that consider the representations of trained models.
Analyses in this line of work typically consider which properties are encoded by MT models, with a specific focus on the difference between the representations within layers that are situated at different levels of the hierarchy of a model.
A robust finding from such analyses is that features such as syntactic constituency, part-of-speech tags and dependency edges can be reliably predicted from the hidden representations of both recurrent neural networks \citep{shi2016does,belinkov2017evaluating,blevins2018deep} and transformer models \citep{raganato2018analysis,tenney2019you}.
Generally, lower-level features are encoded in lower layers, while higher-level syntactic and semantic features are better represented in deeper layers \citep[e.g.][]{blevins2018deep,tenney2019bert}.
For transformer models, a recent wave of papers demonstrates that such features can also be extracted from attention patterns \citep{vig2019analyzing,marevcek2018extracting,lin2019open}.
While these results do not straightforwardly extend to the questions about compositionality that we are considering in this work, they do demonstrate that both recurrent and attention-based models trained in a setup similar to the one considered for this work are able to capture the types of higher-level syntactic features that are often considered to be key for compositional behaviour.

\subsection{Intermediate conclusions}

We reviewed various attempts to assess the extent to which neural models are able to implement compositionality and hierarchy.
This overview illustrated the difficulty and importance of evaluating the behaviour of neural models but also showed that whether neural networks can or do learn compositionally is still an open question.
Both strands of approaches we considered -- approaches that use special compositional data sets to train and test models, and approaches that instead focus on the evaluation of pre-trained models -- report positive as well as negative results.

In the first approach, researchers try to encode a certain notion of compositionality in the task itself. 
While it is important, when testing for compositionality, to make sure the specific  task that networks are trained on has a clear demand for compositional solutions, we believe these studies fall short in explicitly linking the task they propose to clearly-defined notions of compositionality.
Further, we believe that the multifaceted notion of compositionality cannot be exhausted in one single task.
In the following section, we disconnect testing compositionality from the task at hand and disentangle five different theoretically motivated ways in which a network can exhibit compositional behaviours that are not a priori linked to a specific downstream task.

The second type of studies roots its tests into clear linguistic hypotheses. 
However, by testing neural networks that are trained on uncontrolled data, they lose the direct connection between compositionality and the downstream task. 
Although compositionality is widely considered to play an important role for natural language, it is unknown what type of compositional skills -- if any -- a model needs to have to successfully model tasks involving natural language, such as for instance language modelling.
If it cannot be excluded that successful heuristics or syntax-insensitive approximations exist, a negative result can not be taken as evidence that a particular type of model cannot capture compositionality, it merely indicates that this exact model instance did not capture it in this exact case.
While, in the long run, we also wish to reconnect the notion of compositionality to natural data, we believe that before being able to do so, it is of primary importance to reach an agreement about what defines compositional behaviour and how it should be tested for in neural networks.

\section{Testing compositionality}
\label{sec:compositionality_tests}

In the previous section, we discussed various attempts to evaluate the compositional skills of neural network models.
We argued that progressing further on this question requires more clarity on what defines compositionality for neural networks, which we address in this work by providing tests that are more strongly grounded in the literature about compositionality.
We now arrive at the theoretical part of the core of our research, in which we set the theoretical ground for the five tests we propose and conduct in this paper.
We describe five aspects of compositionality that are explicitly motivated by theoretical literature on this topic and propose, on a high level, how to translate them into behavioural tests for (neural) models.

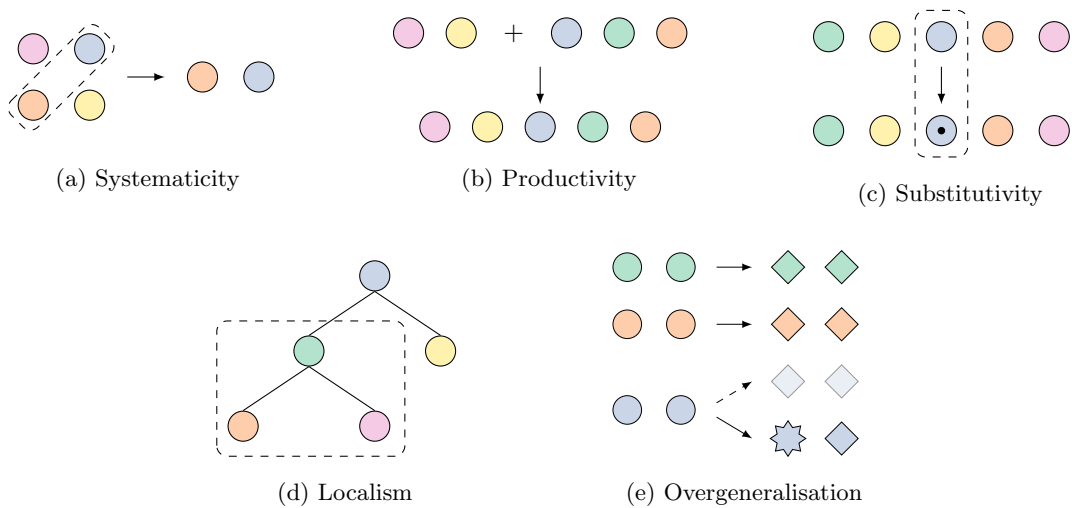
\begin{figure}[!ht]
\centering
\begin{subfigure}[b]{0.30\linewidth}
\centering
\scalebox{1.0}{\begin{tikzpicture}
\node[draw, minimum size=\radius, fill=pink_, circle] at (0,0) {};
\node[draw, minimum size=\radius, fill=blue_, circle] at (0.75,0) {};
\node[draw, minimum size=\radius, fill=orange_, circle] at (0,-0.75) {};
\node[draw, minimum size=\radius, fill=yellow_, circle] at (0.75,-0.75) {};

\draw[-latex] (1.25, -0.375) to (1.75,-0.375);

\node[draw, minimum size=\radius, fill=orange_, circle] at (2.25,-0.375) {};
\node[draw, minimum size=\radius, fill=blue_, circle] at (3,-0.375) {};

\draw [dashed, rounded corners, rotate around={-45:(0,-0.75)}] (-0.25,-1.05) rectangle (0.25,0.6);

\end{tikzpicture}}
\vspace{2mm}
\caption{Systematicity}
\vspace{2mm}
\end{subfigure}
\hfill
\begin{subfigure}[b]{0.30\linewidth}
\centering
\scalebox{1.0}{\begin{tikzpicture}

%\node[draw, minimum size=\radius, fill=pink_, circle] at (0,0) {};
%\node[draw, minimum size=\radius, fill=yellow_, circle] at (0.7,0) {};

\node[draw, minimum size=\radius, fill=pink_, circle] at (-1.4,-0.75) {};
\node[draw, minimum size=\radius, fill=yellow_, circle] at (-0.7,-0.75) {};
\node[minimum size=\radius, fill=white, circle] at (0.0,-0.75) {+};
\node[draw, minimum size=\radius, fill=blue_, circle] at (0.7,-0.75) {};
\node[draw, minimum size=\radius, fill=green_, circle] at (1.4,-0.75) {};
\node[draw, minimum size=\radius, fill=orange_, circle] at (2.1,-0.75) {};

\draw[-latex] (0.35, -1.2) to (0.35,-1.7);

\node[draw, minimum size=\radius, fill=pink_, circle] at (-1.05,-2) {};
\node[draw, minimum size=\radius, fill=yellow_, circle] at (-0.35,-2) {};
\node[draw, minimum size=\radius, fill=blue_, circle] at (0.35,-2) {};
\node[draw, minimum size=\radius, fill=green_, circle] at (1.05,-2) {};
\node[draw, minimum size=\radius, fill=orange_, circle] at (1.75,-2) {};

\end{tikzpicture}}
\vspace{1mm}
\caption{Productivity}
\vspace{2mm}
\end{subfigure}
\hfill
\begin{subfigure}[b]{0.30\linewidth}
\centering
\scalebox{1.0}{\begin{tikzpicture}

\foreach \y in {0, -1.25}
{
    \node[draw, minimum size=\radius, fill=green_, circle] at (0,\y) {};
    \node[draw, minimum size=\radius, fill=yellow_, circle] at (0.75,\y) {};
    \node[draw, minimum size=\radius, fill=blue_, circle] at (1.5,\y) {};
    \node[draw, minimum size=\radius, fill=orange_, circle] at (2.25,\y) {};
    \node[draw, minimum size=\radius, fill=pink_, circle] at (3,\y) {};
}
\draw[-latex] (1.5, -0.4) to (1.5,-0.9);
\draw (1.5, -1.25) node[circle,fill=black, inner sep=1pt] {};
\draw [dashed, rounded corners] (1.15,-1.6) rectangle (1.85,0.35);    

\end{tikzpicture}}
\vspace{1mm}
\caption{Substitutivity}
\end{subfigure}\\
\vspace{5mm}
\begin{subfigure}[b]{0.30\textwidth}
\centering
\scalebox{1.0}{\begin{tikzpicture}[level distance=1cm, sibling distance=1.75cm]
\node[draw, fill=blue_, minimum size=\radius, circle] {}
    child { node[draw, fill=green_, minimum size=\radius, circle] {}
        child { node[draw, fill=orange_, minimum size=\radius, circle] {}}
        child { node[draw, fill=pink_, minimum size=\radius, circle] {}}}
    child {node[draw, fill=yellow_, minimum size=\radius, circle] {}};

\draw [dashed, rounded corners] (-2.1,-2.4) rectangle (0.4,-0.6);    
    
\end{tikzpicture}}
\vspace{1mm}
\caption{Localism}
\end{subfigure}
\hspace{5mm}
\begin{subfigure}[b]{0.30\textwidth}
\centering
\scalebox{0.95}{\begin{tikzpicture}
    
    \foreach \x in {0, 0.75}
    {
        \node[draw, minimum size=\radius, fill=green_, circle] at (\x,0) {};
        \node[draw, minimum size=\radius, fill=orange_, circle] at (\x,-0.8) {};
        \node[draw, minimum size=\radius, fill=blue_, circle] at (\x,-2) {};
        
        \node[draw, minimum size=\radius, fill=green_, diamond] at (\x+2.25,0) {};
        \node[draw, minimum size=\radius, fill=orange_, diamond] at (\x+2.25,-0.8) {};
        \node[draw, minimum size=\radius, fill=blue_, diamond, opacity=0.4] at (\x+2.25,-1.6) {};
    }
    
    \node[draw, minimum size=\radius, fill=blue_, star, star points=8] at (2.25,-2.4) {};
    \node[draw, minimum size=\radius, fill=blue_, diamond] at (3,-2.4) {};
    
    \foreach \y in {0, -0.8}{\draw [-latex] (1.25, \y) to (1.75, \y);}
    \draw [-latex, dashed] (1.25, -1.9) to (1.75, -1.6);
    \draw [-latex] (1.25, -2.1) to (1.75, -2.4);
\end{tikzpicture}}
\vspace{1mm}
\caption{Overgeneralisation}
\end{subfigure}
\hfill
\caption{Schematic depictions of the five tests we propose to test the compositionality of neural network models.
    (a) To test for systematicity, we evaluate models' ability to recombine known parts to form new sequences. 
    (b) While the productivity test also requires recombining known parts, the focus there lies on unboundedness: we test if models can understand sequences \textit{longer} than the ones they were trained on. 
    (c) In the substitutivity test, we evaluate how robust models are towards the introduction of synonyms, and, more specifically, in which cases words are considered synonymous. 
    (d) The localism test targets how local the composition operations of models are: are smaller constituents evaluated before larger constituents? 
    (e) The overgeneralisation test evaluates how likely models are to infer rules: is a model instantly able to accommodate exceptions, or does it need more evidence to deviate from applying the general rule instantiated by the rest of the data?}\label{fig:all_tests}
\end{figure}

We propose to test (i) if models systematically recombine known parts and rules (\emph{systematicity}) (ii) if models can extend their predictions beyond the length they have seen in the training data (\emph{productivity}) (iii) if models' predictions are robust to synonym substitutions (\emph{substitutivity}) (iv) if models' composition operations are local or global (\emph{localism})  and (v) if models favour rules or exceptions during training (\emph{overgeneralisation}).
Below, we describe the theory that motivated us to select these aspects, and we describe on an abstract level how we translate them into concrete tests.
A systematic depiction is shown in Figure~\ref{fig:all_tests}.

\subsection{Systematicity}

The first property we propose to test for -- following many of the works presented in the related work section of this article -- is \textit{systematicity}, a notion frequently used in the context of compositionality.
The term was introduced by \citet{fodor1988connectionism} -- notably, in a paper that argued against connectionist architectures -- who used it to denote that 
\begin{quote}
``[t]he ability to produce/understand some sentences is intrinsically connected to the ability to produce/understand certain others'' \citep[p. 25]{fodor1988connectionism}
\end{quote}
This ability concerns the recombination of known parts and rules: anyone who understands a number of complex expressions also understands other complex expressions that can be built up from the constituents and syntactical rules employed in the familiar expressions. 
To use a classic example from \citet{szabo2012case}: someone who understands `brown dog' and `black cat' also understands `brown cat'.

\citet{fodor1988connectionism} contrast systematicity with storing all sentences in an atomic way, in a dictionary-like mapping from sentences to meanings.
Someone who entertains such a dictionary would not be able to understand new sentences, even if these sentences were similar to the ones occurring in their dictionary.
Since humans are able to infer meanings for sentences they have never heard before, they must use some systematic process to construct these meanings from the ones they internalised before.
By the same argument, however, any model that is able to generalise to a sequence outside its training space (its test set), should have learned to construct outputs from parts it perceived during training and some rule of recombination.
Thus, rather than asking if a model is systematic, a more interesting question is whether the rules and constituents the model uses are in line with what we believe to be the actual rules and constituents underlying a particular data set or language.

\subsubsection{Testing systematicity}
With our \textit{systematicity} test, we aim to pull out that specific aspect, by testing if a model can recombine constituents that have not been seen together during training.
In particular, we focus on combinations of words \emph{a} and \emph{b} that meet the requirements that (i) the model has only been familiarised with \emph{a} in contexts excluding \emph{b} and vice versa but (ii) the combination \emph{a} \emph{b} is plausible given the rest of the corpus.

\subsection{Productivity}
\label{sec:productivity}

A notion closely related to systematicity is \textit{productivity}, which concerns the open-ended nature of natural language: language appears to be infinite, but has to be stored with finite capacity.
Hence, there must be some productive way to generate new sentences from this finite storage.\footnote{The term productivity also has a technical meaning in morphology, which we do not imply here.}
While this `generative' view of language became popular with Chomsky in the early sixties \citep{chomsky1956three}, Chomsky himself traces it back to Von Humboldt, who stated that `language makes infinite use of finite means'.

Both systematicity and productivity rely on the recombination of known constituents into larger compounds.
However, whereas systematicity can be -- to some extent -- empirically established, productivity cannot, as it is not possible to prove that natural languages in fact contain an infinite number of complex expressions \citep{pullum2010recursion}.
Even if humans' memory allowed them to produce infinitely long sentences, their finite life prevents them from doing so. 
Productivity of language is therefore more controversial than systematicity.

\subsubsection{Testing productivity}
To separate systematicity from productivity, in our productivity test we specifically focus on the aspect of unboundedness.
We test whether a model can understand sentences that are \textit{longer} than the ones encountered during training.
To test this, we separate sequences in the data based on length and evaluate models on their ability to cope with longer sequences after having been familiarised with the shorter ones.

\subsection{Substitutivity}

A principle closely related to the principle of compositionality is the principle of \emph{substitutivity}.
This principle, which finds its origin in philosophical logic, states that if an expression is altered by replacing one of its constituents with another constituent with the same meaning (a synonym), this does not affect the meaning of the expression \citep{pagin2003communication}.
In other words, if a substitution preserves the meaning of the parts of a complex expression, it also preserves the meaning of the whole.
In the latter formulation, the correspondence with the principle of compositionality itself can be easily seen: as substituting part of an expression with a synonym changes neither the structure of the expression nor the meaning of its parts, it should not change the meaning of the expression itself either.

Like the principle of compositionality, the substitutivity principle in the context of natural language is subject to interpretation and discussion.
\cite{husserl1913logische} pointed out that the substitution of expressions with the same meaning can result in nonsensical sentences if the expressions belong to different semantic categories (the philosopher Geach (1965) illustrated this considering the two expressions \emph{Plato was bald} and \emph{baldness was an attribute of Plato}.
While these expressions are synonymous, it is not possible to substitute the first with the second in the sentence \emph{The philosopher whose most eminent pupil was Plato was bald}).

A second context which poses a challenge for the substitutivity principle concerns embedded statements about beliefs.
As already sketched out in the previous section, if \emph{X} and \emph{Y} are synonymous, this does not necessarily imply that the expressions \emph{Peter thinks that X} and \emph{Peter thinks that Y} are both true.
In this work, we do not consider these cases, but instead focus on the more general question: is substitutivity a salient notion for neural networks and under what conditions can and do they infer synonymity?

\subsubsection{Testing substitutivity}
We test substitutivity by probing under which conditions a model considers two atomic units to be synonymous.
To this end, we artificially introduce synonyms and consider how the prediction of a model changes when an atomic unit in an expression is replaced by its synonym.
We consider two different cases.
Firstly, we analyse the case in which synonymous words occur equally often and in comparable contexts.
In this case, synonymity can be inferred from the corresponding meanings on the output side but is aided by distributional similarities on the input side.
Secondly, we consider pairs of words in which one of the words occurs only in very short sentences, which we call \textit{primitive contexts}.
In this case, synonymity can only be inferred from the (implicit) semantic mapping, which is identical for both words, but not from the context that those words appear in.

\subsection{Localism}

In its basic form, the principle of compositionality states that the meaning of a complex expression derives from the meanings of its constituents and how they are combined.
It does not impose any restrictions on what counts as an admissible way of combining different elements, which is why the principle taken in isolation is formally vacuous.\footnote{We previously cited \citet{janssen1983}, who proved this claim by showing that arbitrary sets of expressions can be mapped to arbitrary sets of meanings without violating the principle of compositionality, as long as one is not bound to a fixed syntax.}
As a consequence, the interpretation of the principle of compositionality depends on the type of constraints that are put on the semantic and syntactic theories involved.
One important consideration concerns -- on an abstract level -- how \textit{local} the composition operations should be.
When operations are very local (a case also referred to as \textit{strong} or \emph{first-level} compositionality), the meaning of a complex expression depends only on its local structure and the meanings of its immediate parts \citep{pagin2010compositionality, jacobson2002dis}.
In other words, the meaning of a compound is only dependent on the meaning of its immediate `children', regardless of the way that their meaning was built up.
In \textit{global} or \emph{weak} compositionality, the meaning of an expression follows from its total (global) structure and the meanings of its atomic parts. 
In this interpretation, compounds can have different meanings, depending on the larger expression that they are a part of.

\citet{carnap1988meaning} presents an example that nicely illustrates the difference between these two interpretations, in which he considers sentences with tautologies.
Under the view that the meaning of declarative sentences is determined by the set of all worlds in which this sentence is true, any two tautologies \emph{X} and \emph{Y} are synonymous.
Under the local interpretation of compositionality, this entails that also the phrases \emph{Peter thinks that X} and \emph{Peter thinks that Y} should be synonymous, which is not necessarily the case, as Peter may be aware of some tautologies but unaware of others.
The global interpretation of compositionality does not give rise to such a conflict, as \emph{X} and \emph{Y}, despite being identical from a truth-conditional perspective, are not structurally identical.
Under this representation, the meanings of \emph{X} and \emph{Y} are locally identical, but not globally, if also the phrase they are a part of is considered.
For natural language, contextual effects, such as the disambiguation of a phrase or word by a full utterance or even larger piece of discourse, make the localist account highly controversial.
As a contrast, consider an arithmetic task, where the outcome of \texttt{14 - (2 + 3)} does not change when the subsequence \texttt{(2 + 3)} is replaced by \texttt{5}, a sequence with the same (local) meaning, but a different structure.

\subsubsection{Testing localism}
We test if a model's composition operations are local or global by comparing the meanings the model assigns to stand-alone sequences to those it assigns to the same sequences when they are part of a larger compound.
More specifically, we compare a model's output when it is given a composed sequence \emph{X}, built up from two parts \emph{A} and \emph{B} with the output the same model gives when it is forced to first separately process \emph{A} and \emph{B} in a local fashion.
If the model employs a local composition operation that is true to the underlying compositional system that generated the language, there should be no difference between these two outputs.
A difference between these two outputs, instead, indicates that the model does not compute meanings by first computing the meanings of all subparts, but pursues a different, more global, strategy.

\subsection{Overgeneralisation}
\label{sec:overgeneralisation}

The previously discussed compositionality arguments are of mixed nature. 
Some -- such as productivity and systematicity -- are linked to the way that humans acquire and process language. 
Others -- such as substitutivity and localism -- are properties of the mapping from signals to meanings in a particular language.
While it can be tested if a language user abides by these principles, these principles themselves do not relate directly to language users.
To complete our set of tests to assess whether a model learns compositionally, we include also a notion that exclusively concerns the acquisition of the language by a model: we consider if models exhibit \textit{overgeneralisation} when faced with \textit{non}-compositional phenomena.

Overgeneralisation (or overregularisation) is a language acquisition term, which refers to the scenario in which a language learner applies a general rule in a case that forms an exception to this rule.
One of the most well-known examples, which served also as the subject of the famous \emph{past-tense debate} between symbolism and connectionism \citep{rumelhart1986learning,marcus1992overregularization}, concerns the rule that English past-tense verbs can be formed by appending \emph{-ed} to the stem of the verb.
During the acquisition of past-tense forms, learners infrequently use this rule also for irregular verbs, resulting in forms like \emph{goed} (instead of \emph{went}) or \emph{breaked} (instead of \emph{broke}).

The relation of overgeneralisation with compositionality comes from the supposed evidence that overgeneralisation errors provide for the presence of symbolic rules in the human language system \citep[see, e.g.][]{penke2012dual}.
In this work, we follow this line of reasoning and take the application of a rule in a case where this is contradicted by the data 
as evidence that the model in fact internalised this rule.
As such, we regard a model's inclination to apply rules as the expression of a compositional bias.
This inclination is most easily observed in the case of exceptions, where the correct strategy is to ignore the rules and learn on a case-by-case basis. 
If a model overgeneralises by applying the rules also in such cases, we hypothesise that this in particular demonstrates compositional awareness.

\subsubsection{Testing overgeneralisation}
We propose an experimental setup where a model's tendency to overgeneralise is evaluated by monitoring its behaviour on exceptions.
We identify samples that do not adhere to the rules underlying the data distribution -- \emph{exceptions} -- in the training data sets and assess a model's tendency to overgeneralise by observing how they respond to these exceptions during training: (when) do they consistently follow a global rule set, and (when) do they (over)fit the training samples individually?

\section{Data}\label{sec:data}

As observed by many others before us, insight in the compositional skills of neural networks is not easily acquired by studying models trained on natural language directly.
While it is generally agreed upon that compositional skills are required to appropriately model natural language, successfully modelling natural data requires far more than understanding compositional structures.
As a consequence, a negative result may stem not from a model's incapability to model compositionality, but rather from the lack of signal in the data that should induce compositional behaviour.
A positive result, on the other hand, cannot necessarily be explained as successful compositional learning, since it is difficult to establish that a good performance cannot be reached through heuristics and memorisation.
In this article, we therefore consider an artificial translation task, in which sequences that are generated by a probabilistic context free grammar (PCFG) should be translated into output sequences that represent their meanings.
These output sequences are constructed by recursively applying the \emph{string edit operations} that are specified in the input sequence.
The task, which we dub \emph{\pcfg}, does not contain any non-compositional phenomena, and we can thus be certain that compositionality is in fact a salient feature.
At the same time, we construct the input data such that in other dimensions -- such as the lengths of the sentences and depths of the parse trees -- it matches the statistical properties of a corpus with sentences from natural language (English).

\subsection{Input sequences: syntax}

The input alphabet of PCFG SET contains three types of words: words for unary and binary functions that represent \emph{string edit operations} (e.g.\ \texttt{append}, \texttt{copy}, \texttt{reverse}), elements to form the string sequences that these functions can be applied to (e.g.\ \texttt{A}, \texttt{B}, \texttt{A1}, \texttt{B1}), and a separator to separate the arguments of a binary function (\texttt{,}).
The input sequences that are formed with this alphabet are sequences describing how a series of such operations are to be applied to a string argument.
For instance:\\

\begin{tabular}{lcl}
$\texttt{repeat A B C }$  \\
$\texttt{echo remove\_first D K , E F}$  \\
$\texttt{append swap F G H , repeat I J}$ \\
\end{tabular}\\

\medskip

We generate input sequences with a PCFG, shown in Figure~\ref{fig:pcfgset_syntax} (production probabilities are omitted).  
As the grammar that we use is recursive, we can generate an infinite number of admissible input sequences.
Because the operations can be nested, the parse trees of valid sequences can be arbitrarily deep and long.
Additionally, the distributional properties of a particular \pcfg data set can be controlled by adjusting the probabilities of the grammar and varying the number of input characters.
We will use this to \emph{naturalise} the data set such that its distribution of lengths and depths correspond to the distribution observed in a data set containing English sentences.

\begin{figure}[]
\centering
\setlength{\tabcolsep}{2pt}
\renewcommand{\arraystretch}{1.1}
\begin{tabular}{|llll|}
    \hline
\textbf{Non-terminal rules} & & & \\
$S$ & $\to$ & $F_U\ S \;\; \mid \;\; F_B\ \ S\ ,\ S$ &\\
$S$ & $\to$ & $X$ &\\
$X$ & $\to$ & $X X$ &\\
& & & \\
\textbf{Lexical rules} & & & \\
$F_U$ & $\to$ & $\texttt{copy} \;\mid\; 
                 \texttt{reverse} \;\mid\; 
                 \texttt{shift} \;\mid \;
                 \texttt{echo} \;\mid\; 
                 \texttt{swap} \;\mid\;
                 \texttt{repeat}\;$ &\\
$F_B$ & $\to$ & $\texttt{append} \;\mid\; 
                 \texttt{prepend} \;\mid\; 
                 \texttt{remove\_first} \;\mid\; 
                 \texttt{remove\_second}$    &\\
 $X$ & $\to$ & \texttt{A} $\;\mid\;$ \texttt{B} $\;\mid\;$ \ldots $\;\mid\;$ \texttt{Z} $\;\mid\;$ \texttt{A2} $\;\mid\;$ \ldots $\;\mid\;$ \texttt{B2} $\;\mid\;$ \dots & \\ 
\hline
\end{tabular}

\caption{The context free grammar that describes the entire space of grammatical input sequences in \pcfg.
The rule probabilities (not depicted) can be used to control the distributional properties of a \pcfg. We use this property to make sure that our data matches a corpus with natural English sentences in terms of length and depth distributions.}\label{fig:pcfgset_syntax}
\end{figure}

\begin{figure}[!ht]
    \centering
\setlength{\tabcolsep}{3pt}
\renewcommand{\arraystretch}{1.1}
\begin{tabular}{|lllllll|}
\hline
\textbf{Unary functions} $F_U$:&&&
\textbf{Binary functions} $F_B$:&&&\\
$\texttt{copy}\ x_1\ \cdots\ x_n $ & $\to$ & $x_1\ \cdots\ x_n$ &
$\texttt{append}\ \ \mathbf{x}, \ \mathbf{y} $ & $\to$ & $\mathbf{x}\ \mathbf{y}$ &\\
$\texttt{reverse}\ x_1\ \cdots\ x_n $ & $\to$ &  $x_n\ \cdots\ x_1$ & 
$\texttt{prepend}\ \ \mathbf{x},\ \mathbf{y} $ & $\to$ & $\mathbf{y}\ \mathbf{x}$ &\\
$\texttt{shift}\ x_1\ \cdots\ x_n$ & $\to$ & $x_2\ \cdots\ x_n\ x_1$ & 
$\texttt{remove\_first}\ \ \mathbf{x},\ \mathbf{y} $ & $\to$ & $\mathbf{y}$ & \\
$\texttt{swap}\ x_1\ \cdots\ x_n $ & $\to$ & $x_n\ x_2\ \cdots\ x_{n-1}\ x_1$  & 
$\texttt{remove\_second}\ \ \mathbf{x},\ \mathbf{y} $ & $\to$ & $\mathbf{x}$ & \\
$\texttt{repeat}\ x_1 \cdots\ x_n$ & $\to$ & $x_1\ \cdots\ x_n\ x_1\ \cdots\ x_n$ & & & & \\
$\texttt{echo}\ x_1 \cdots\ x_n$ & $\to$ & $x_1\ \cdots\ x_n\ x_n$ & & & & \\
\hline
\end{tabular}
    \caption{The interpretation functions describing how the meaning of \pcfg input sequences is formed.}\label{fig:pcfgset_semantics}
\end{figure}

\subsection{Output sequences: semantics}

The meaning of a \pcfg input sequence is constructed by recursively applying the string edit operations specified in the sequence.
This mapping is governed by the interpretation functions listed in Figure \ref{fig:pcfgset_semantics}. 
Following these interpretation functions, the three sequences listed above would be mapped to output sequences as follows:\\

\begin{tabular}{lcl}
$\texttt{repeat A B C }$  &  $\rightarrow$  &  $\texttt{A B C A B C}$ \\
$\texttt{echo remove\_first D K , E F}$  & $\rightarrow$ & $\texttt{E F F}$ \\
$\texttt{append swap F G H , repeat I J}$  & $\rightarrow$ & $\texttt{H G F I J I J }$\\
\end{tabular}\\

\medskip

The definitions of the interpretation functions specify the systematic procedure by which an output sequence should be formed from an input sequence, without having to enumerate particular input-output pairs.
In this sense, \pcfg is similar to SCAN \citep{lake2018generalization} but differs from a task such as the lookup table task introduced by \cite{livska2018memorize}, where functions must be exhaustively defined because there is no systematic connection between arguments and the values to which functions map them.

\subsection{Data construction}\label{subsec:pcfg-dataset}
\pcfg describes a general framework for producing many different data sets. 
We used several criteria to select the \pcfg input-output pairs for our experiments.

\subsubsection{Naturalisation of structural properties}
\begin{figure}
\begin{subfigure}{0.5\linewidth}
    \includegraphics[width=\linewidth]{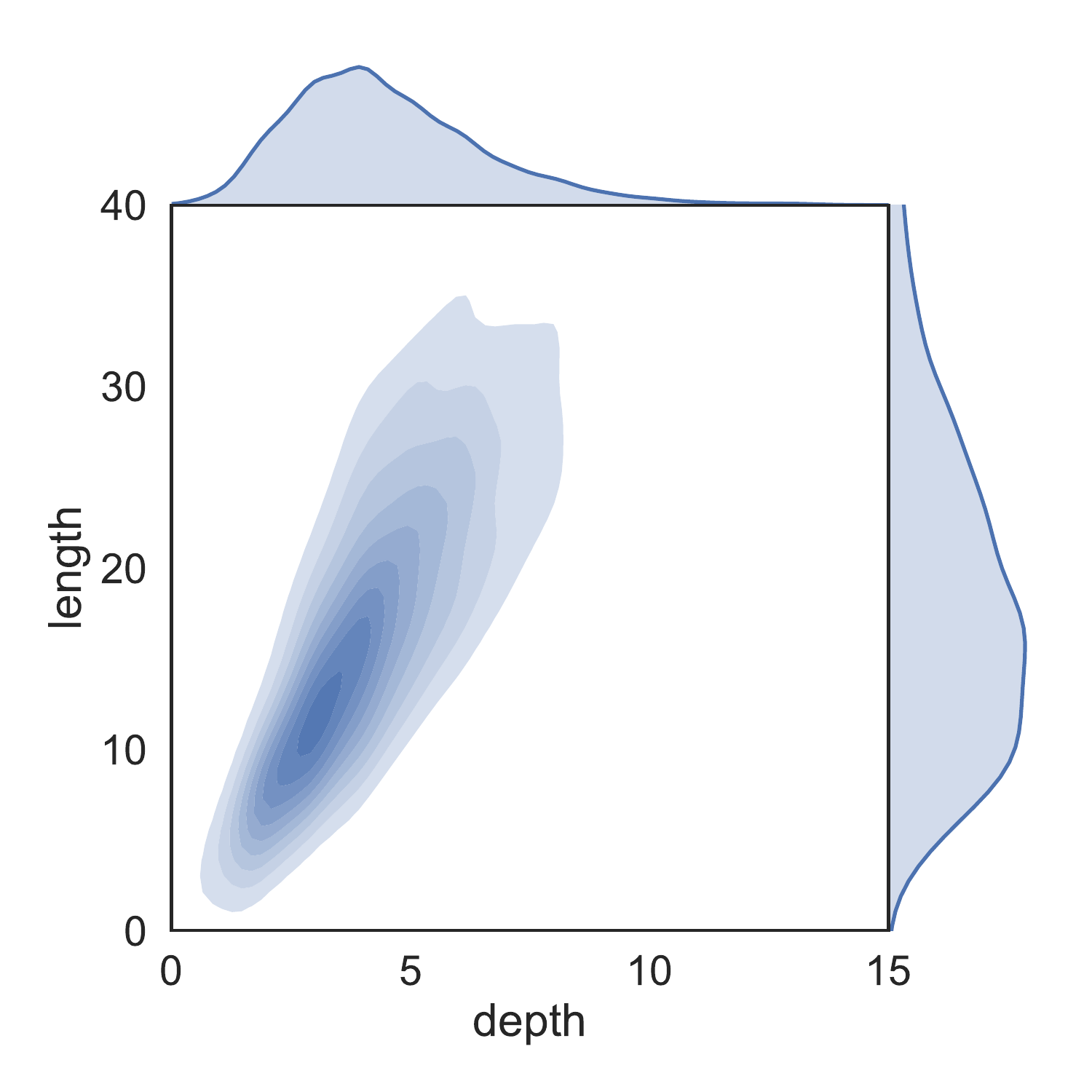}
    \caption{WMT17}
    \label{fig:dist_wmt}
\end{subfigure}
\begin{subfigure}{0.5\linewidth}
    \includegraphics[width=\linewidth]{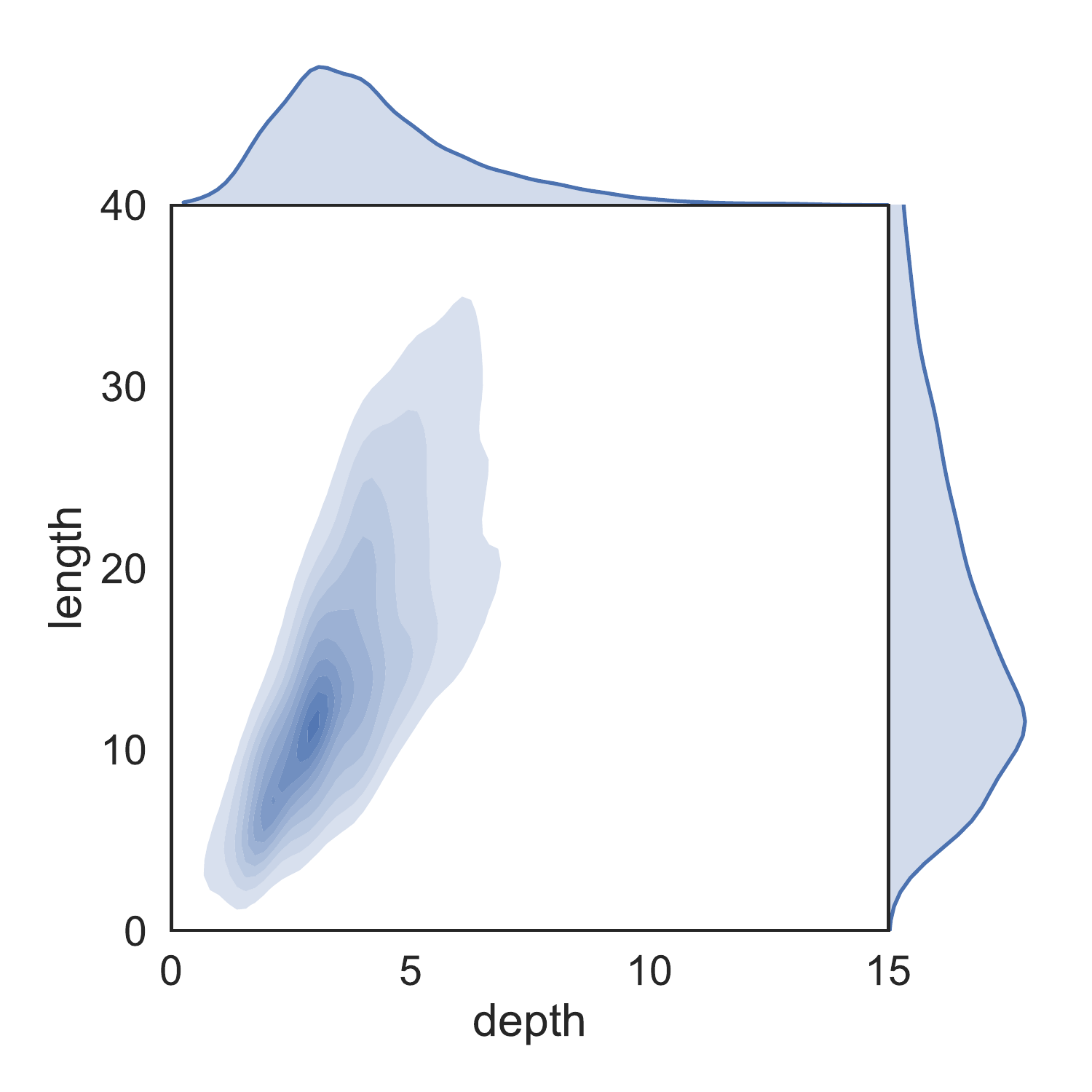}
    \caption{\pcfg data}
    \label{fig:dist_pcfg}
\end{subfigure}
\caption{Distribution of lengths and depths in the \pcfg (left) and English WMT 2017 test data (right).}\label{fig:distributions}
\label{fig:distributions}
\end{figure}

The probabilistic nature of the \pcfg input grammar offers a high level of control over the generated input sequences. 
We use this control to enforce an input distribution that resembles the statistics of a more natural data set in two relevant respects: the length of the expressions, and the depth of their parse trees. 
To obtain these statistics, we use the English side of a large machine translation corpus: WMT 2017 \citep{wmt2017}.
We parse this corpus with a statistical parser \citep{stanfordparser} and extract the distribution of length and depths from the annotated corpus.
We then use expectation maximisation to tune the PCFG parameters in such a way that the resulting bivariate distribution of the generated data mimics the one extracted from the WMT data.
For a more detailed description of the naturalisation procedure we refer to Appendix \ref{appendix:naturalisation}. 
In Figure~\ref{fig:dist_wmt} and Figure~\ref{fig:dist_pcfg}, we plot the distributions of the WMT data and a sample of around ten thousand sentences of the resulting \pcfg data. 

\subsubsection{Sentence selection}
We set the size of the string alphabet to 520 and create a base corpus of around 100 thousand distinct input-output pairs.
We limit the length of the string arguments given to the functions to 5.
We use 85\% of this corpus for training, 5\% for validation and 10\% for testing.
During data generation, further care is taken to make memorisation as unattractive as possible by controlling the string sequences that feature as primitive arguments in the input expressions: we make sure that the same string arguments are never repeated. 
While we do not control re-occurrence of specific subsequence in general, the relatively large string alphabet makes it improbable that particular subsequences occur often enough to make memorisation a profitable learning strategy.

\section{Architectures}\label{sec:models}

To showcase our compositionality test suite, we compare three currently popular neural architectures for sequence-to-sequence language processing tasks such as machine translation, speech processing and language understanding: recurrent neural networks \citep{sutskever2014sequence}, convolutional neural networks \citep{gehring2017convolutional} and transformer networks \citep{vaswani2017attention}.
In this section we explain their most important features, we give a brief overview of their potential strengths and weaknesses in relation to compositionality, and we describe how we implemented them in our experiments.

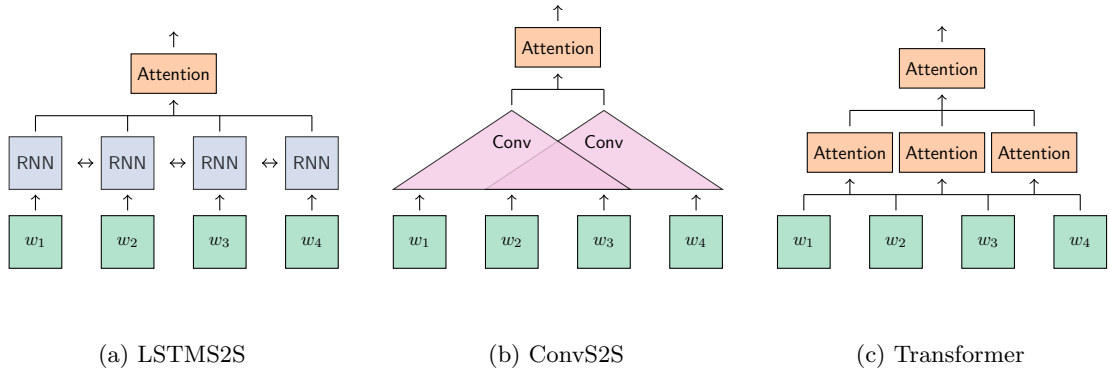
\begin{figure}[ht]
    \begin{subfigure}{.33\textwidth}
        \begin{tikzpicture}[scale=0.70, every node/.style={scale=0.70}]
    \path[] (-1,-1) rectangle (6.25,5);
    \foreach \idx in {1,2,3,4}
    {
        \pgfmathsetmacro\x{1.75*(\idx - 1)};
        \node[draw, minimum size=1cm, fill=green_, rectangle] at (\x,0) {$w_{\idx}$};
        
        \node[draw, minimum size=1cm, fill=blue_, rectangle, align=center, opacity=0.8] at (\x,1.5) {\textsf{RNN}};
        
        \draw [->] (\x, 0.6) to (\x, 0.9);
        \draw [-] (\x, 2.1) to (\x, 2.4);
    }
    
    \foreach \x in {0, 1.75, 3.5}
    {
        \draw [<->] (\x + 0.8, 1.5) to (\x + 1.1, 1.5);
    }
    
    \draw [-] (0, 2.4) to (5.25, 2.4);
    
    \foreach \y in {2.4, 3.7}
    {
        \draw [->] (2.625, \y) to (2.625, \y + 0.3);
    }
    
    \node[draw, minimum size=0.75cm, fill=orange_, rectangle, align=center] at (2.625,3.2) {\textsf{Attention}};
\end{tikzpicture}
        \caption{LSTMS2S}
        \label{fig:lstms2s}
    \end{subfigure}
    \begin{subfigure}{.33\textwidth}
        \begin{tikzpicture}[scale=0.70, every node/.style={scale=0.70}]
    \path[] (-1,-1) rectangle (6.25,5);
    \foreach \idx in {1,2,3,4}
    {
        \pgfmathsetmacro\x{1.75*(\idx - 1)};
        \node[draw, minimum size=1cm, fill=green_, rectangle] at (\x,0) {$w_{\idx}$};
        \draw [->] (\x, 0.6) to (\x, 0.9);
    }
            
    \foreach \x in {1.25, -0.5}
    {
        \draw[fill=pink_,opacity=0.8] (\x,1) node[anchor=north]{}
          -- (\x + 4.5,1) node[anchor=north]{}
          -- (\x + 2.25,2.5) node[anchor=south]{}
          -- cycle;
    }
        
    \node at (1.75, 1.9) {\textsf{Conv}};
    \node at (3.5, 1.9) {\textsf{Conv}};
    
    \foreach \x in {1.75, 3.5}
    {
        \draw[-] (\x, 2.6) to (\x, 2.9);
    }
    
    \draw[-] (1.75, 2.9) to (3.5, 2.9);
    
    \foreach \y in {2.9, 4.2}
    {
        \draw [->] (2.625, \y) to (2.625, \y + 0.3);
    }

    \node[draw, minimum size=0.75cm, fill=orange_, rectangle, align=center] at (2.625,3.7) {\textsf{Attention}};
\end{tikzpicture}
        \caption{ConvS2S}
        \label{fig:convs2s}
    \end{subfigure}
    \begin{subfigure}{.33\textwidth}
        \begin{tikzpicture}[scale=0.70, every node/.style={scale=0.70}]
    \path[] (-1,-1) rectangle (6.25,5);
    \foreach \idx in {1,2,3,4}
    {
        \pgfmathsetmacro\x{1.75*(\idx - 1)};
        \node[draw, minimum size=1cm, fill=green_, rectangle] at (\x,0) {$w_{\idx}$};
        \draw [-] (\x, 0.6) to (\x, 0.9);
    }
    
    \draw [-] (0, 0.9) to (5.25, 0.9);
    
    \foreach \x in {0.875, 2.625, 4.375}
    {
        \draw [->] (\x, 0.9) to (\x, 1.2);
        \node[draw, minimum size=0.75cm, fill=orange_, rectangle, align=center, opacity=1] at (\x,1.7) {\textsf{Attention}};
        \draw [-] (\x, 2.2) to (\x, 2.5);
    }
    
    \draw [-] (0.875, 2.5) to (4.375, 2.5);
    \draw [->] (2.625, 2.5) to (2.625, 2.8);
    
    \node[draw, minimum size=0.75cm, fill=orange_, rectangle, align=center, opacity=1] at (2.625,3.3) {\textsf{Attention}};
    
    \draw [->] (2.625, 3.8) to (2.625, 4.1);
    
\end{tikzpicture}
    \caption{Transformer}
    \label{fig:transformer}
    \end{subfigure}
\caption{High-level graphical depictions of the most important features of the encoding mechanisms in LSTMS2S, ConvS2S and Transformer, as well as how these encoded representations can be attended to by the decoder.
(a) LSTMS2S processes the input in a fully sequential way, iterating over the embedded elements one by one in both directions before applying an attention layer.
(b) ConvS2S divides the input sequence into local fragments of consecutive items that are processed by the same convolutions, before applying attention. 
(c) Transformer immediately applies several global attention layers to the input, without incrementally constructing a representation of the input. 
}
\label{fig:models}
\end{figure}

\subsection{LSTMS2S}

The first architecture we consider is a recurrent encoder-decoder model with attention. 
This setup is considered to be the most basic of the three setups we consider, but is still the basis of many MT applications \citep[e.g. OpenNMT,][]{2017opennmt} and has also been successful in the fields of speech recognition \citep[e.g.][]{chorowski2015attention} and question answering \citep[e.g.][]{golub2016character}. 
We consider the version of this model in which both the decoder and encoder are LSTMs and refer to this setup with the abbreviation \emph{LSTMS2S}. 

\subsubsection{Model basics}
LSTMS2S is a fully recurrent, bidirectional model. 
The encoder processes an input by iterating over all of its elements in both directions and incrementally constructing a representation for the entire sequence. 
Upon receiving the encoder output, the decoder performs a similar, sequential computation to unroll the predicted sequence.
Here, LSTMS2S uses an attention mechanism, which allows it to focus on the parts of the encoded input that are estimated to be most important at each moment in the decoding process. 

The sequential fashion with which the LSTMS2S architecture processes each input potentially limits the model's abilities to recombine components hierarchically. 
While depth -- and, as shown by \citet{blevins2018deep}, thus hierarchy -- can be created by stacking neural layers, the number of layers in popular recurrent sequence-to-sequence setups tends to be limited. 
The attention mechanism of the encoder-decoder setup positively influences the skills of LSTMS2S to hierarchically process inputs, as it allows the decoder to focus on input tokens out of the sequential order.

\subsubsection{Implementation}
We use the LSTMS2S implementation of the OpenNMT-py framework \citep{2017opennmt}.
We set the hidden layer size to 512, number of layers to 2 and the word embedding dimensionality to 512, matching their best setup for translation from English to German with the WMT 2017 corpus, which we used to shape the distribution of the \pcfg data.
We use mini-batches of 64 sequences and stochastic gradient descent with an initial learning rate of 0.1.

\subsection{ConvS2S} 
The second architecture we consider is a convolutional-based architecture. 
Convolutional sequence-to-sequence models have obtained competitive results in machine translation \citep{gehring2016convolutional} and abstractive summarisation \citep{denil2014modelling}.
In this paper, we follow the setup described by \citet{gehring2017convolutional} and use also their nomenclature: we refer to this model with the abbreviation \textit{ConvS2S}.

\subsubsection{Model basics}
ConvS2S uses a convolutional model to encode input sequences, instead of a recurrent one.
The decoder uses a multi-step attention mechanism --  every layer has a separate attention mechanism -- to generate outputs from the encoded input representations.
Although the convolutions already contextualise information in a sequential order, the source and target embeddings are also combined with position embeddings that explicitly encode order.
As at the core of the ConvS2S model lies the local mechanism of one-dimensional convolutions, which are repeatedly and hierarchically applied, ConvS2S has a built-in bias for creating compositional representations. 
Its topology is also biased towards the integration of local information, which may hinder modelling long-distance relations.
However, convolutional networks have been found to maintain a much longer effective history than their recurrent counterparts \citep{bai2018empirical}.
Within ConvS2S, distant portions in the input sequence may be combined primarily through the multi-step attention, which has been shown to improve the generalisation abilities of the model compared to single-step attention \citep{dessi2019cnns}.

\subsubsection{Model implementation}
In the ConvS2S setup that was presented by \citet{gehring2017convolutional} that we use in this work, word vectors are 512-dimensional. 
Both the encoder and decoder have 15 layers, with 512 hidden units in the first 10 layers, followed by 768 units in two layers, all using kernel width 3. 
The final three layers are 2048-dimensional.
We train the network with the Fairseq Python toolkit\footnote{Fairseq toolkit: \url{https://github.com/pytorch/fairseq}}, using the predefined \texttt{fconv\_wmt\_en\_de} architecture. 
Unless mentioned otherwise, we use the default hyperparameters of this library.
We replicate the training procedure of \citet{gehring2017convolutional}, using Nesterov's accelerated gradient method and an initial learning rate of 0.25. 
We use mini-batches of 64 sentences, with a maximum number of tokens of 3000. 
The gradients are normalised by the number of non-padded tokens in a batch.

\subsection{Transformer} 

The last architecture we consider is the recently introduced transformer model \citep{vaswani2017attention}. 
Transformers constitute the current state of the art in machine translation and are becoming increasingly popular also in other domains, such as language modelling \citep[e.g.][]{radford2019language}.
We refer to this setup with simply the name \emph{Transformer}.

\subsubsection{Model basics}
Transformers use neither recurrent cells nor convolutions to convert an input sequence to an output sequence.
Instead, they are fully based on a multitude of attention mechanisms. 
Both the encoder and decoder of a transformer are composed of a number of feed-forward layers, each containing two sub-layers: a multi-head attention module and a traditional feed-forward layer. 
In the multi-head attention layers, several attention tensors (the `heads') are computed in parallel, concatenated and projected.
In addition to a self-attention layer, the decoder has a layer that computes multi-head attention over the outputs of the encoder.

Since transformers do not have any inherent notion of sequentiality, the input embeddings are combined with position embeddings, from which the model can infer \textit{order}.
For transformers, the cost of relating symbols that are far apart is thus not higher than relating words that are close together, which -- in principle -- should make it easier to model long-distance dependencies.
Furthermore, the relatively many stacked layers in a transformer model should facilitate modelling hierarchical structure.
On the other hand, the non-sequential nature of the transformer could be a handicap as well, particularly for relating consecutive portions in the input sequence. 
A transformer's receptive field is inherently global, which can be challenging in such cases.

\subsubsection{Implementation}
We use a transformer model with an encoder and decoder that both contain six stacked layers.
The multi-head self-attention module of the model has eight heads, and the feed-forward network has a hidden size of 2048. 
All embedding layers and sub-layers in the network produce outputs of dimensionality 512. 
In addition to word embeddings, positional embeddings are used to indicate word order. 
We use OpenNMT-py\footnote{Pytorch port of OpenNMT: \url{https://github.com/OpenNMT/OpenNMT-py}.} \citep{2017opennmt} to train the model according to the guidelines provided by the framework\footnote{Visit \url{http://opennmt.net/OpenNMT-py/FAQ.html} for the guidelines.}: with the Adam optimiser ($\beta_1=0.9$ and $\beta_2=0.98$) and a learning rate increasing for the first 8000 `warm-up steps' and decreasing afterwards.

\section{Experiments and results}\label{sec:results}

We now proceed to test our tests on the three previously described architectures.
Below, we describe the precise experiments we conducted and report their results, going test by test.
We train all models of all architectures for 25 epochs, or until convergence, and select the best-performing model based on the performance on the validation set.
For every experiment, we conduct three runs per architecture and report both the average and standard deviation of their scores.\footnote{Some experiments, such as the localism experiment, can be conducted directly on models trained for other tests and thus do not require training new models.}
A summary of the results is shown in Table~\ref{tab:pcfg_results}.
The data and scripts to run these experiments as well as the trained models are all available online.\footnote{\url{https://github.com/i-machine-think/am-i-compositional}}

As described before, we did not run a grid-search to optimise the hyper-parameters of the three architectures we investigate, but instead selected reasonable hyper-parameters from papers that previously used these architectures for comparable data.
It is possible that changing the hyper-parameters would also change the results of the experiments.
It is thus important to keep in mind that the described experiments and result serve as an illustration of the usefulness of our tests.
With fixed data and a varying training seed, our tests show consistent and interesting differences and similarities between the three setups we used, but these results should not be taken as general claims about LSTMs, convolutional networks or transformers.

\begin{table}
\centering
{\renewcommand{\arraystretch}{1.3}}
\begin{tabular}{@{}lccc@{}}
    \toprule
    \textbf{Experiment} & \textbf{LSTMS2S} & \textbf{ConvS2S} & \textbf{Transformer}\\\midrule \midrule
    Task accuracy$^\ast$        & 0.79 \pms{0.01} & 0.85 \pms{0.01} & 0.92 \pms{0.01} \\ \midrule
    Systematicity$^\ast$        & 0.53 \pms{0.03} & 0.56 \pms{0.01} & 0.72 \pms{0.00} \\ \midrule
    Productivity$^\ast$         & 0.30 \pms{0.01} & 0.31 \pms{0.02} & 0.50 \pms{0.02} \\ \midrule 
    Substitutivity, \textit{equally distributed}$\dagger$
                                & 0.80 \pms{0.00} & 0.95 \pms{0.00} & 0.98 \pms{0.00} \\ 
    Substitutivity, \textit{primitive}$\dagger$
                                & 0.60 \pms{0.01} & 0.58 \pms{0.01} & 0.90 \pms{0.00} \\ \midrule
    Localism$\dagger$           & 0.46 \pms{0.00} & 0.59 \pms{0.01} & 0.54 \pms{0.02} \\ \midrule
    Overgeneralisation$^\ast$   & 0.68 \pms{0.04} & 0.79 \pms{0.06} & 0.88 \pms{0.07} \\
    \bottomrule
\end{tabular}

\caption{General task performance and performance per test for \pcfg. 
The results are averaged over three runs and the standard deviation is indicated. 
Two performance measures are used: \emph{sequence accuracy}, indicated by $^\ast$, and \emph{consistency score}, indicated by $\dagger$.}
\label{tab:pcfg_results}
\end{table}

\subsection{Task accuracy}\label{subsec:task_succes}

We first consider the correctness of the output sequences of the three different architectures on the data as described in Section~\ref{subsec:pcfg-dataset}.
In particular, we consider their \emph{sequence accuracy}, where only instances for which the entire output sequence equals the target are considered correct.
We use this accuracy measure to evaluate the overall task performance, and we use it later also for the systematicity, productivity, and overgeneralisation tests.
In the rest of this paper, we denote accuracy scores with $^\ast$.

The average task performance on the \pcfg data for the three different architectures is shown in the first row of Table~\ref{tab:pcfg_results}.
The Transformer outperforms both LSTMS2S and ConvS2S ($p\approx 10^{-4}$ and $p\approx 10^{-3}$, respectively), with a surprisingly high accuracy of 0.92.
ConvS2S, in turn, is with its 0.85 accuracy significantly better than LSTMS2S ($p\approx 10^{-3}$), which has an accuracy 0.79.
The scores of the three architectures are robust with respect to initialisation and order of presentation of the data, as evidenced by the low variation across runs.
We now present a breakdown of this task accuracy on different types of subsets of the data.

\subsubsection{Impact of length, depth and number of functions}
\begin{figure}
    \centering
    \begin{subfigure}{\textwidth}
    \centering
    \includegraphics[width=0.50\textwidth]{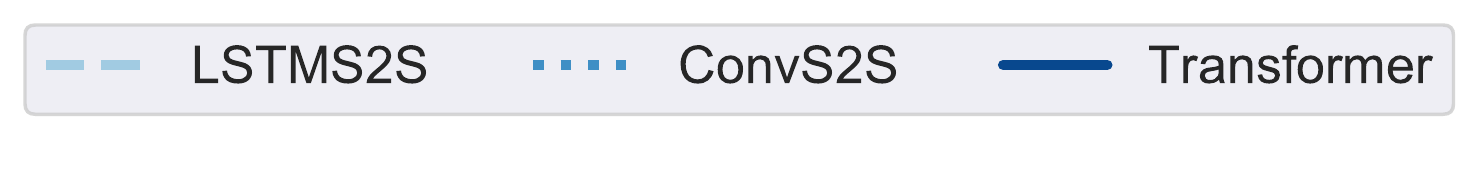}
    \end{subfigure}
    \begin{subfigure}{0.35\textwidth}
    \includegraphics[width=\textwidth]{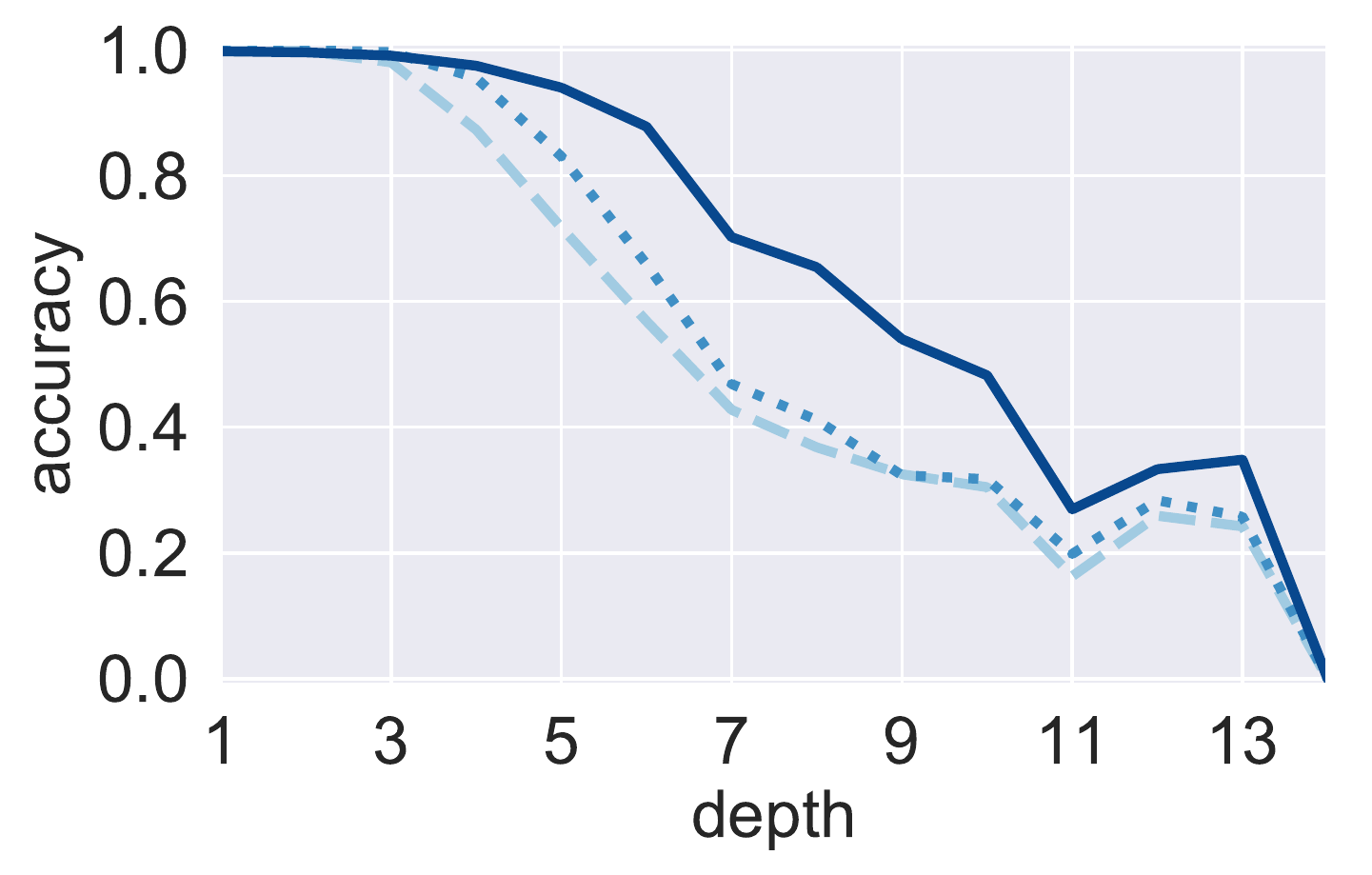}
    \end{subfigure}
    \begin{subfigure}{0.31\textwidth}
    \includegraphics[width=\textwidth]{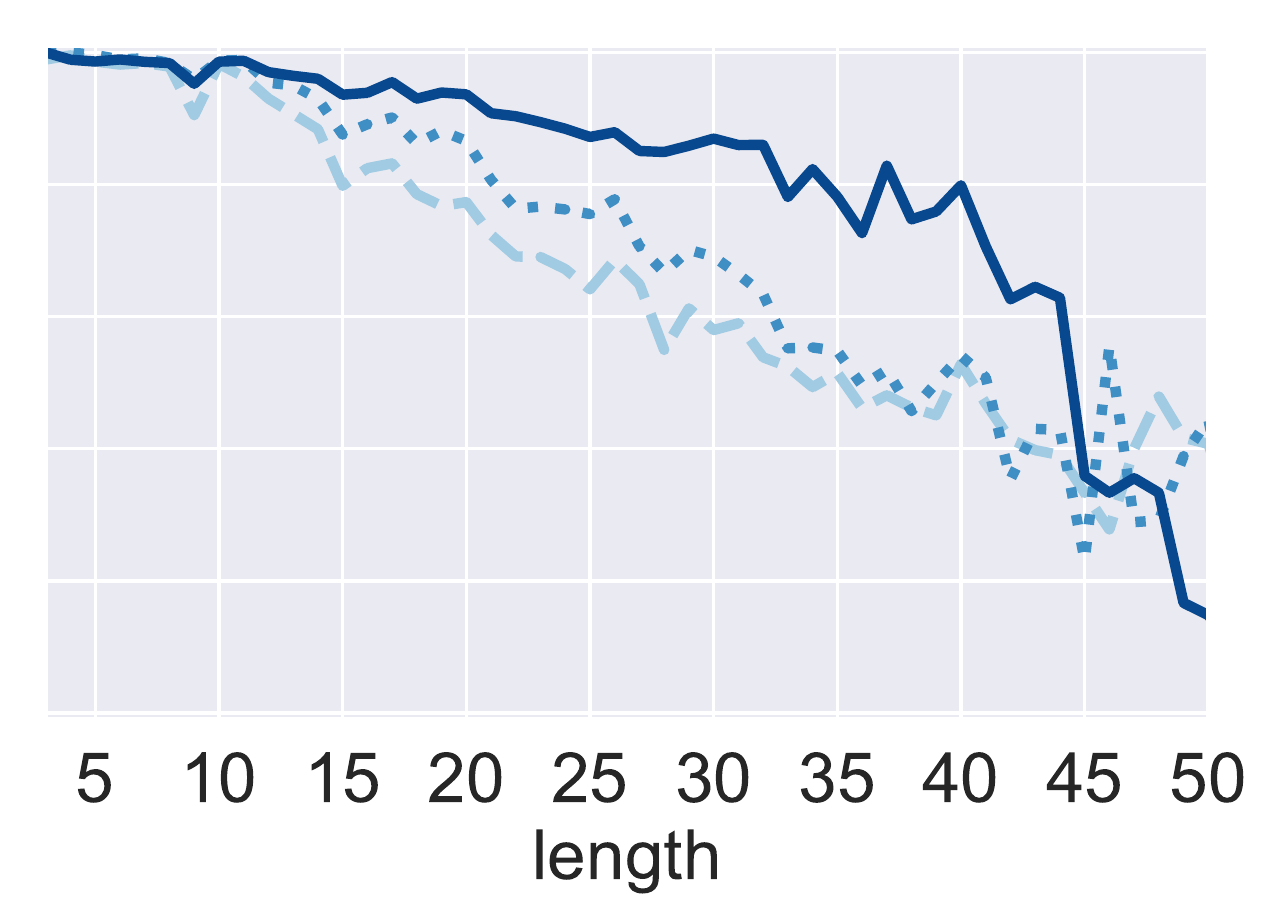}
    \end{subfigure}
    \begin{subfigure}{0.31\textwidth}
    \includegraphics[width=\textwidth]{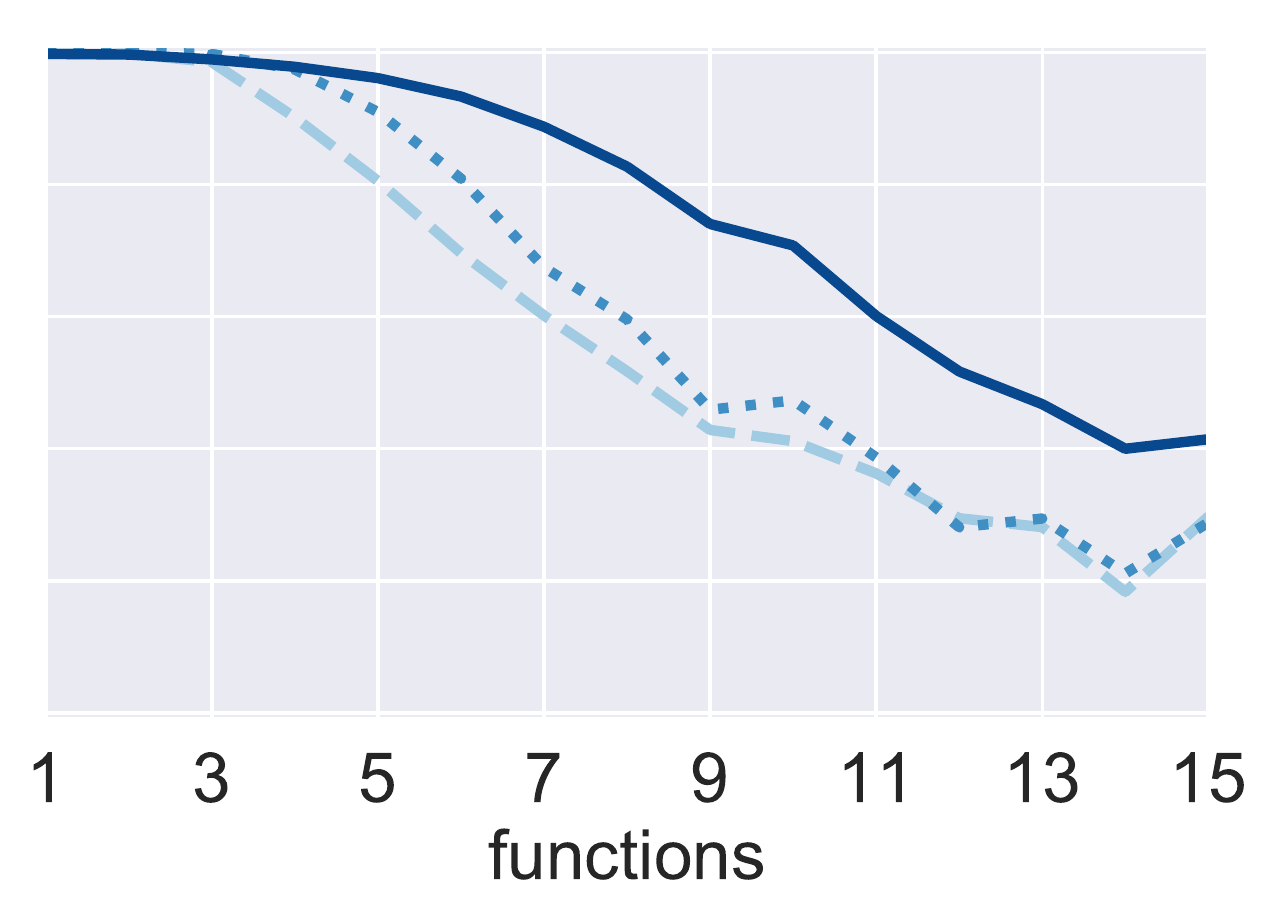}
    \end{subfigure}
    \caption{Average sequence accuracy of the three architectures as a function of several properties of the input sequences for the general PCFG SET test set: the \textit{depth} of the input's parse tree, the input sequence's \textit{length} and the \textit{number of functions} in the input sequence.
    The results are averaged over three runs and computed over ten thousand test samples.}
    \label{fig:depth_length_tokens}
\end{figure}

We explore how the accuracy of the three different architectures develops with increasing difficulty of the input sequences, as measured in the input sequence's depth (the maximum level of nestedness observed in a sequence), the input sequence's length (number of tokens) and the number of functions in the input sequence.
In Figure~\ref{fig:depth_length_tokens}, we plot the average sequence accuracy for all three architectures as a function of those difficulty measures.
Unsurprisingly, the accuracy of all architecture types decreases with the length, depth and number of functions in the input.
All architectures have learned to successfully model sequences with low depths and lengths and a small number of functions (reflected by accuracies close to 1).
Their performance drops for longer sequences with more functions.
Overall the Transformer $>$ ConvS2S $>$ LSTMS2S trend is preserved across the different data subsets.

\subsubsection{Function difficulty}
\begin{figure}
    \centering
    \begin{subfigure}{0.4\textwidth}
    \includegraphics[width=\textwidth]{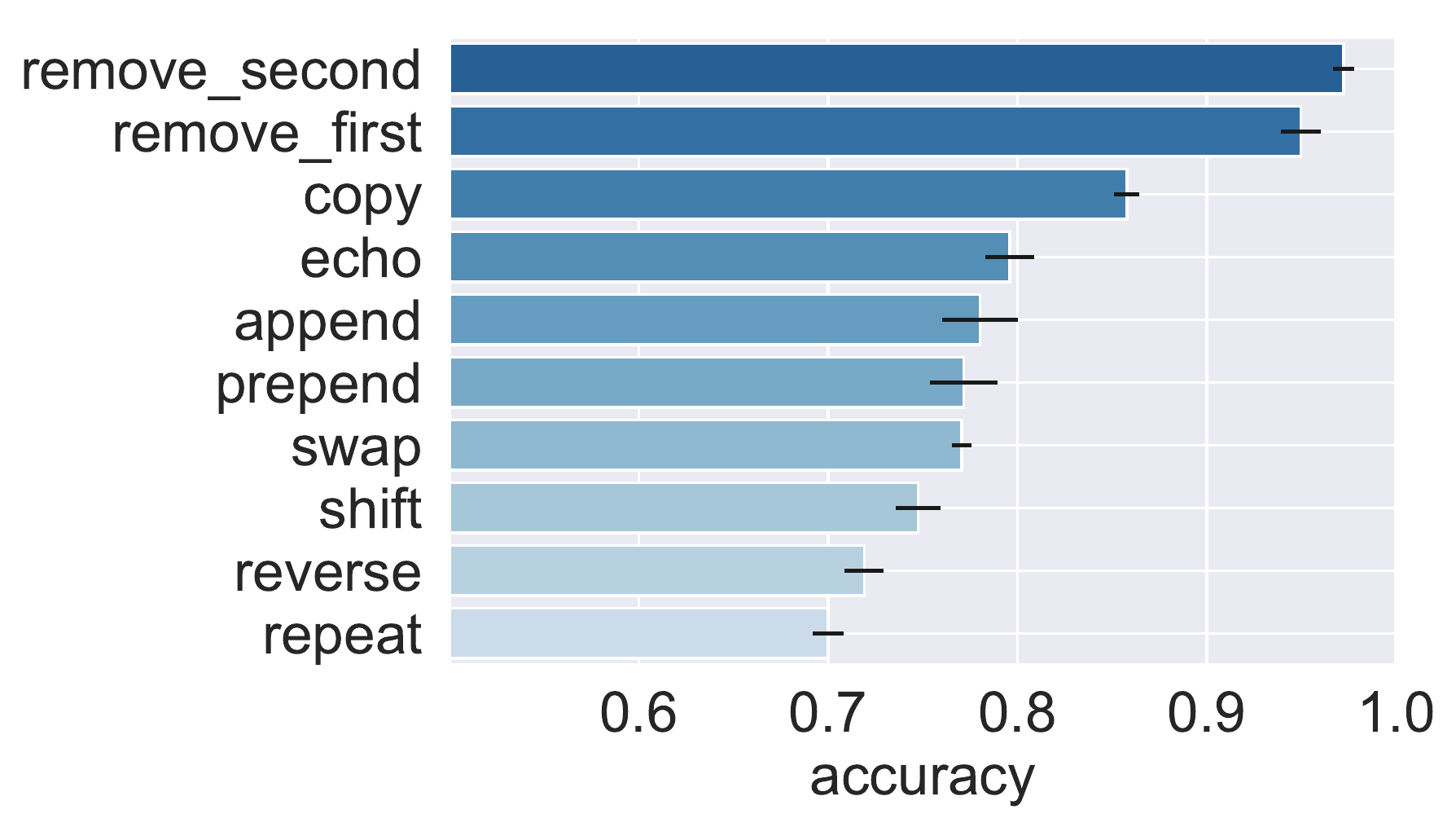}
    \caption{LSTMS2S}
    \end{subfigure}
    \begin{subfigure}{0.29\textwidth}
    \includegraphics[width=\textwidth]{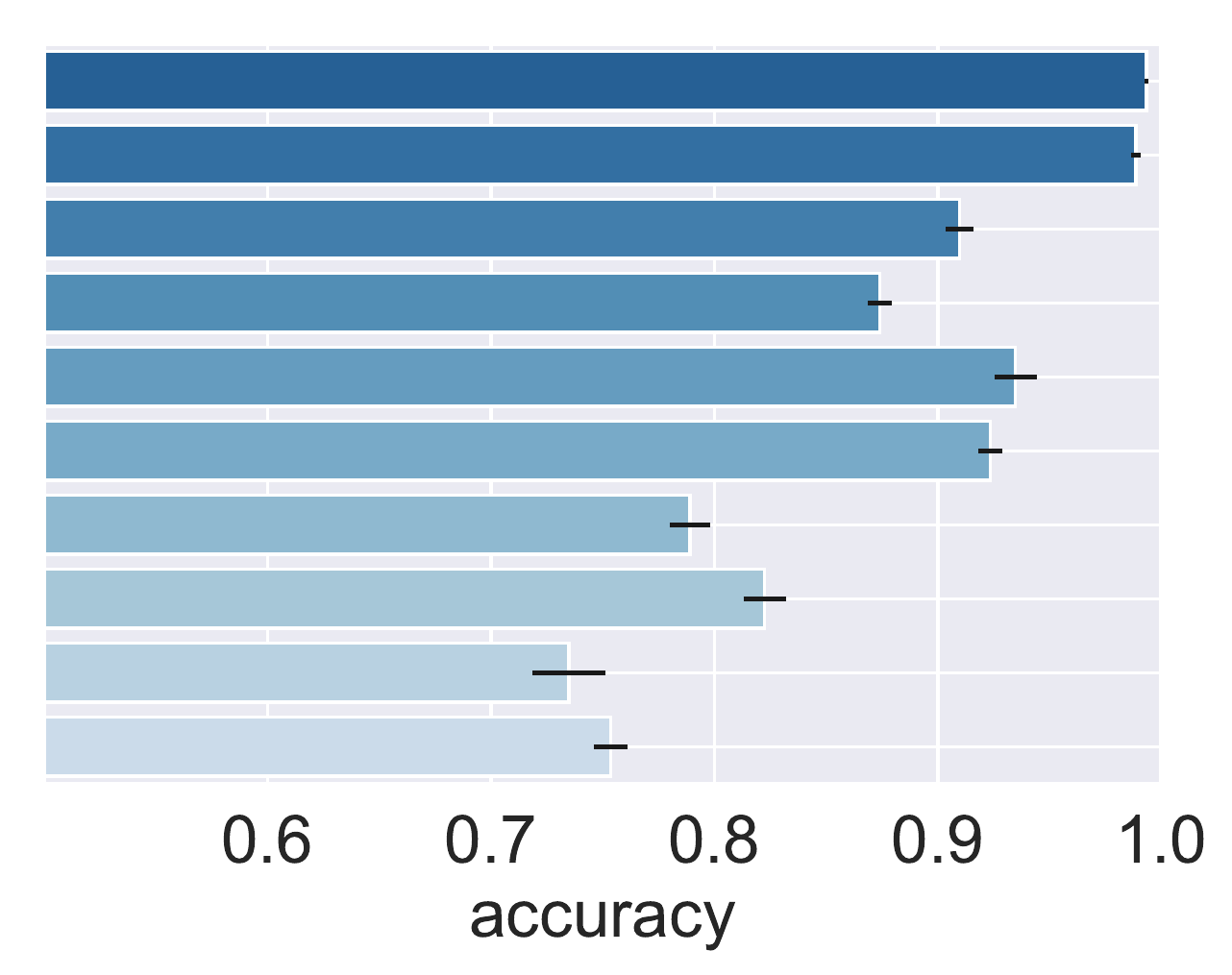}
    \caption{ConvS2S}
    \end{subfigure}
    \begin{subfigure}{0.29\textwidth}
    \includegraphics[width=\textwidth]{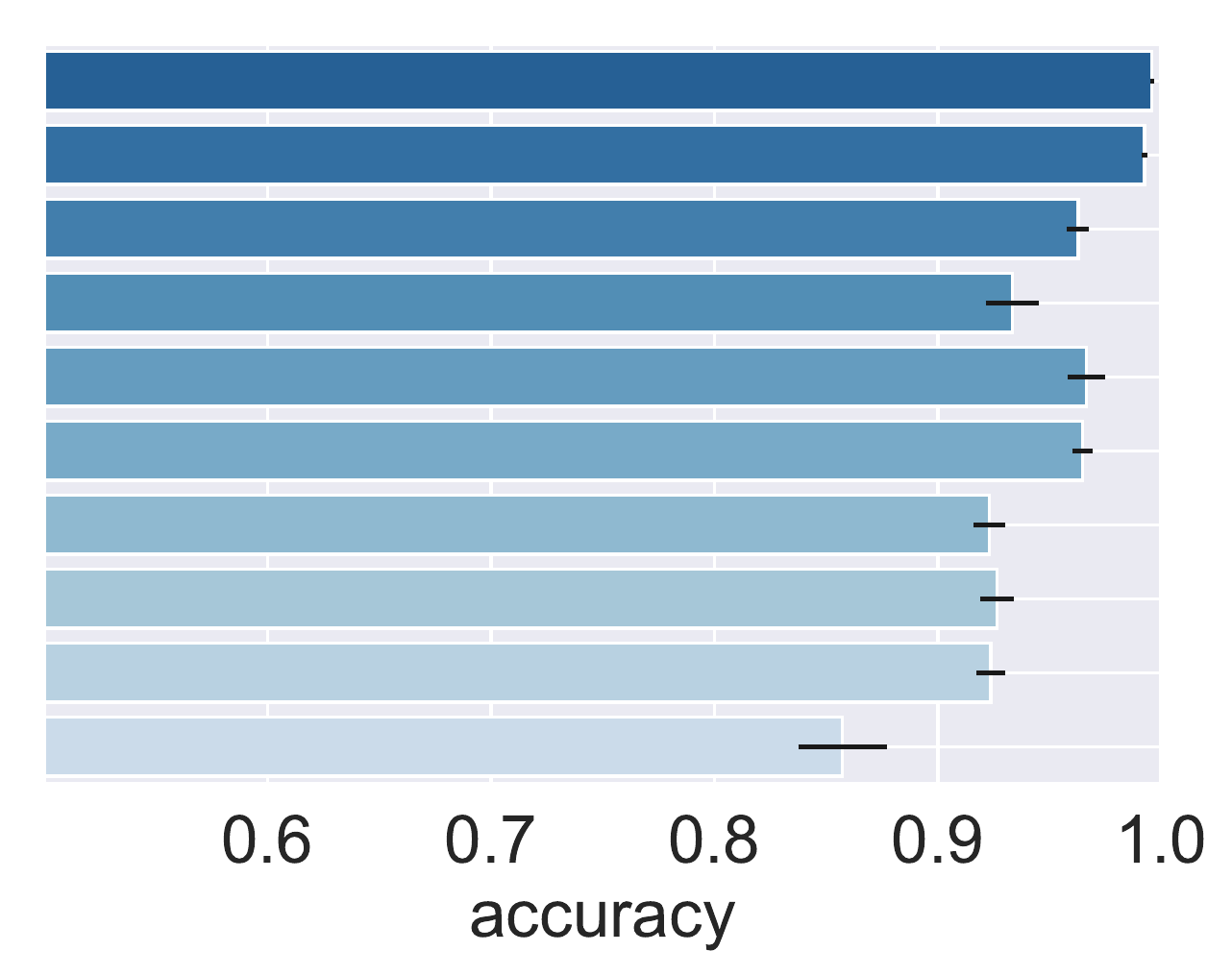}
    \caption{Transformer}
    \end{subfigure}
    \caption{Accuracy of the three models per PCFG SET function, as computed by applying the different functions to the same complex input sequences.}\label{fig:function_difficulty}
\end{figure}

Since the input sequences typically contain multiple functions, it is not possible to directly evaluate whether some functions are more difficult for models than others.
On sequences that contain only one function, all models achieve a maximum accuracy.
To compare the difficulty of the functions, we create one corpus with composed input sequences and derive for each function a separate corpus in which this function is applied to those composed input sequences.
We then express the comparative difficulty of a function for a model as this model's accuracy on the corpus corresponding to this function.
For example, to compare the functions \texttt{echo} and \texttt{reverse}, we create two minimally different corpora that only differ with respect to the first input function in the sequence (e.g.\ \texttt{\textcolor{blue}{echo} append swap F G H , repeat I J} and \texttt{\textcolor{blue}{reverse} append swap F G H , repeat I J}),  and compute the model's accuracy on both corpora.\footnote{Note that since inputs to unary and binary functions are different, we have to use two different corpora to compare binary and unary function difficulty. The unary and binary function scores in Figure~\ref{fig:function_difficulty} are thus not directly comparable.}
We plot the results in Figure~\ref{fig:function_difficulty}.

The ranking of functions in terms of difficulty is similar for all models, suggesting that the difficulties are to a large extent stemming from the objective complexity of the functions themselves, rather than from specific biases in the models.
In some cases, it is very clear why. 
The function \texttt{echo} requires copying the input sequence and repeating its last element -- regardless of the bias of the model, this should be at least as difficult as \texttt{copy} which requires just to copy the input.
Similarly, \texttt{prepend} and \texttt{append} require repeating two string arguments, whereas for \texttt{remove\_first} and \texttt{remove\_second} only one argument needs to be repeated.
The latter functions should thus be easier, irrespective of the architecture.
The relative difficulty of \texttt{repeat} reflects that generating longer output sequences proves challenging for all architectures. 
As this function requires to output the input sequence twice, its output is on average twice as long as the output of another unary function applied to an input string of the same length.

An interesting difference between architectures occurs for the function \texttt{reverse}.
For both LSTMS2S and ConvS2S this is a difficult function (although \texttt{repeat} is even harder than \texttt{reverse} for LSTMS2S).
For Transformer, the accuracy for \texttt{reverse} is on par with the accuracies of \texttt{echo}, \texttt{swap} and \texttt{shift}, functions that are substantially easier than \texttt{reverse} for the other two architectures.
This difference follows directly from architectural differences: while LSTMS2S and ConvS2S are forced to encode ordered local context -- as they are recurrent or apply local convolutions -- Transformer is not bound to such an ordering and can thus more easily deal with inverted sequences.

\subsection{Systematicity}\label{tab:systematicity_results}

The task success results for \pcfg already reflect whether models can recombine functions and input strings that were not seen together during training.
In the systematicity test, we focus explicitly on models' ability to interpret pairs of functions that were never seen together while training.

\subsubsection{Test details}
We evaluate four pairs of functions: \texttt{swap repeat}, \texttt{append remove\_second}, \texttt{repeat remove\_second} and \texttt{append swap}.\footnote{To decrease the number of dimensions of variation, we keep the specific pairs of functions fixed during evaluation: rather than varying the function pairs evaluated across runs, we vary the initialisation and order of presentation of the training examples.}
We redistribute the training and test data such that the training data does not contain any input sequences including these specific four pairs and all sequences in the test data contain at least one.
After this redistribution, the training set contains 82 thousand input-output pairs, while the test set contains 10 thousand examples.
Note that while the training data does not contain any of the function pairs listed above, it still may contain sequences that contain both functions.
E.g.\ \texttt{reverse repeat remove\_second A B , C D} cannot appear in the training set, but \texttt{repeat reverse remove\_second A B , C D} might.

\subsubsection{Results}
The results of the systematicity test are reported in row 2 of Table~\ref{tab:pcfg_results}.
In Table~\ref{tab:systematicity_breakdown}, we show the average accuracies of the three architectures on all four held out function pairs.
Following the task accuracy, also for the systematicity test, Transformer obtains higher scores than both LSTMS2S and ConvS2S ($p\approx 10^{-2}$ and $p\approx 10^{-3}$, respectively).
The difference between the latter two, however, is for this test statistically insignificant ($p\approx 10^{-1}$).
The relative differences between Transformer and the other two architectures gets larger.
Intriguingly, the systematicity scores of all models are substantially lower than their overall task accuracies.
This large difference is surprising, since \pcfg is constructed such that a high task accuracy requires systematic recombination.
As such, these results serve as a reminder that models may find unexpected solutions, even when the data is very carefully constructed.

One potential explanation for this score discrepancy is that, due to the slightly different distribution of examples in the systematicity data set, the models learn a different solution than before.
Since the functions occurring in the held out pairs are slightly under-sampled, it could be that the models' representations of these functions are not as good as the ones they develop when trained on the regular data set.
A second explanation, to which our localism test will lend more support, is that models do treat the inputs and functions systematically, but analyse the sequences in terms of different units.
Obtaining a high accuracy for \pcfg undoubtedly requires being able to systematically recombine functions and input strings, but it does not necessarily require developing separate representations that capture the semantics of the different functions individually.

For instance, if there is enough evidence for \texttt{repeat copy}, a model may learn to directly apply the combination of these two functions to an input string, rather than consecutively appealing to separate representations for the two functions.
Thus, to compute the output of a sequence like \texttt{repeat copy swap echo X}, the model may apply two times a pair of functions, instead of four separate functions.
Such a strategy would not necessarily harm performance in the overall data set, since plenty of evidence for all function pairs is present, but it would affect performance on the systematicity test, where this is not the case.
While larger chunking to ease processing is not necessarily a bad strategy, we argue that it is desirable if models can also maintain a separate representation of the units that make up such chunks, which may be needed in other contexts.

\begin{table}
\small
    \centering
    \begin{tabular}{lccc}
    \toprule
    & \textbf{LSTMS2S} & \textbf{ConvS2S} & \textbf{Transformer} \\\midrule \midrule
    \texttt{swap repeat}             & 0.40 \pms{0.04} & 0.49 \pms{0.02} & 0.53 \pms{0.03} \\
    \texttt{append remove\_second}   & 0.54 \pms{0.04} & 0.46 \pms{0.03} & 0.80 \pms{0.02} \\
    \texttt{repeat remove\_second}   & 0.66 \pms{0.02} & 0.67 \pms{0.01} & 0.80 \pms{0.01} \\
    \texttt{append swap}             & 0.48 \pms{0.03} & 0.56 \pms{0.01} & 0.73 \pms{0.01} \\
    \midrule \midrule
    \textit{All}                     & 0.53 \pms{0.03} & 0.56 \pms{0.01} & 0.72 \pms{0.00} \\ 
    \bottomrule
    \end{tabular}
    \caption{The average sequence accuracy per pair of held out compositions for the systematicity test.}
    \label{tab:systematicity_breakdown}
\end{table}

\subsection{Productivity}\label{subsec:productivity_results}

In Figure~\ref{fig:depth_length_tokens}, we saw that longer sequences are more difficult for all models, even if their length and depth fall within the range of lengths and depths observed in the training examples.
There are several potential causes for this drop in accuracy.
It could be that longer sequences are simply more difficult than shorter ones: they contain more functions, and there is thus more opportunity to make an error.
Additionally, simply because they contain more functions, longer sequences are more likely to contain at least one of the more difficult functions (see Figure~\ref{fig:function_difficulty}).
Lastly, due to the naturalisation of the distribution of lengths, longer sequences are underrepresented in the training data.
There is thus fewer evidence for long sequences than there is for shorter ones.
As such, models may have to perform a different kind of generalisation to infer the meaning of longer sequences than they do for shorter ones.
Their decrease in performance when sequences grow longer could thus also be explained by a general poor ability to generalise to lengths outside their training space, a type of generalisation sometimes referred to with the term \emph{extrapolation}.
With our productivity test, we focus purely on this extrapolation aspect, by studying models' ability to successfully generalise to longer sequences, which we will call the model's \emph{productive power}.

\subsubsection{Test details}
To test for productivity, we redistribute the training and testing data such that there is no evidence at all for longer sequences in the training set.
Sequences containing up to eight functions are collected in the training set, consisting of 81 thousand sequences, while input sequences containing at least nine 
functions are used for evaluation and collected in a test set containing 11 thousand sequences.
The average, minimum and maximum length, depth and number of functions for the train and test set of the productivity test are shown in Table~\ref{tab:productivity}.

\begin{table}
    \centering
    \begin{tabular}{lccccccccc}
    \toprule
                                & \multicolumn{3}{c}{\textbf{\small Depth}} 
                                & \multicolumn{3}{c}{\textbf{\small Length}}
                                & \multicolumn{3}{c}{\textbf{\small \#Functions}} \\
        & \textit{min} & \textit{max} & \textit{avg} & \textit{min} & \textit{max} & \textit{avg} & \textit{min} & \textit{max} & \textit{avg}\\ \midrule \midrule
        \textit{Productivity} \\
        Train       & 1 & 8 & 3.9 & 3 & 53 & 16.3 & 1 & 8 & 4.3 \\
        Test & 4 & 17 & 8.2 & 14 & 71 & 32.9 & 9 & 35 & 11.5 \vspace{1mm} \\
        \textit{\pcfg} \\
        Train       & 1 & 17 & 4.4 & 3 & 71 & 18.4 & 1 & 35 & 5.2 \\
        Test        & 1 & 17 & 4.4 & 3 & 71 & 18.2 & 1 & 28 & 5.1 \\
    \bottomrule
    \end{tabular}
    \caption{The average, minimum and maximum length, depth and number of functions for the train and test set of the productivity test. We provide the same measures for the \pcfg test data set for comparison.}.\label{tab:productivity}
\end{table}

\subsubsection{Results}

The overall accuracy scores on the productivity test in Table \ref{tab:pcfg_results} demonstrate that all models have great difficulty with extrapolating to sequences with a higher length than those seen during training.
Transformer drops to a mean accuracy of 0.50; LSTMS2S and ConvS2S have a test accuracy of 0.30 and 0.31, respectively. 
Relatively speaking, removing evidence for longer sequences thus resulted in a 62\% drop for  LSTMS2S, a 64\% drop in ConvS2S, and a 46\% drop for Transformer.
Both in terms of absolute and relative performance, Transformer thus has a much greater productive potential than the other models, although its absolute performance is still poor.

\begin{figure}[]
    \centering
    \begin{subfigure}{\textwidth}
    \centering
    \includegraphics[width=0.80\textwidth]{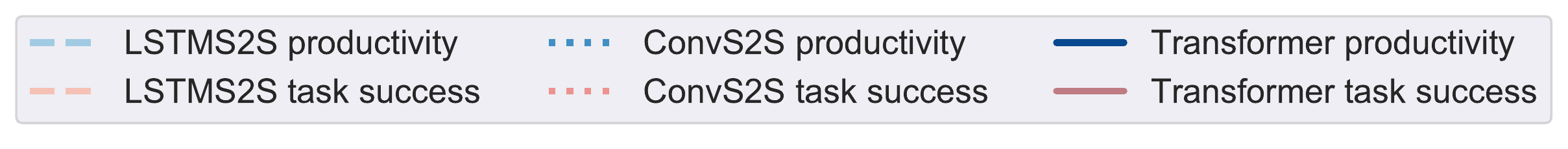}
    \end{subfigure}
    \begin{subfigure}{0.35\textwidth}
    \includegraphics[width=\textwidth]{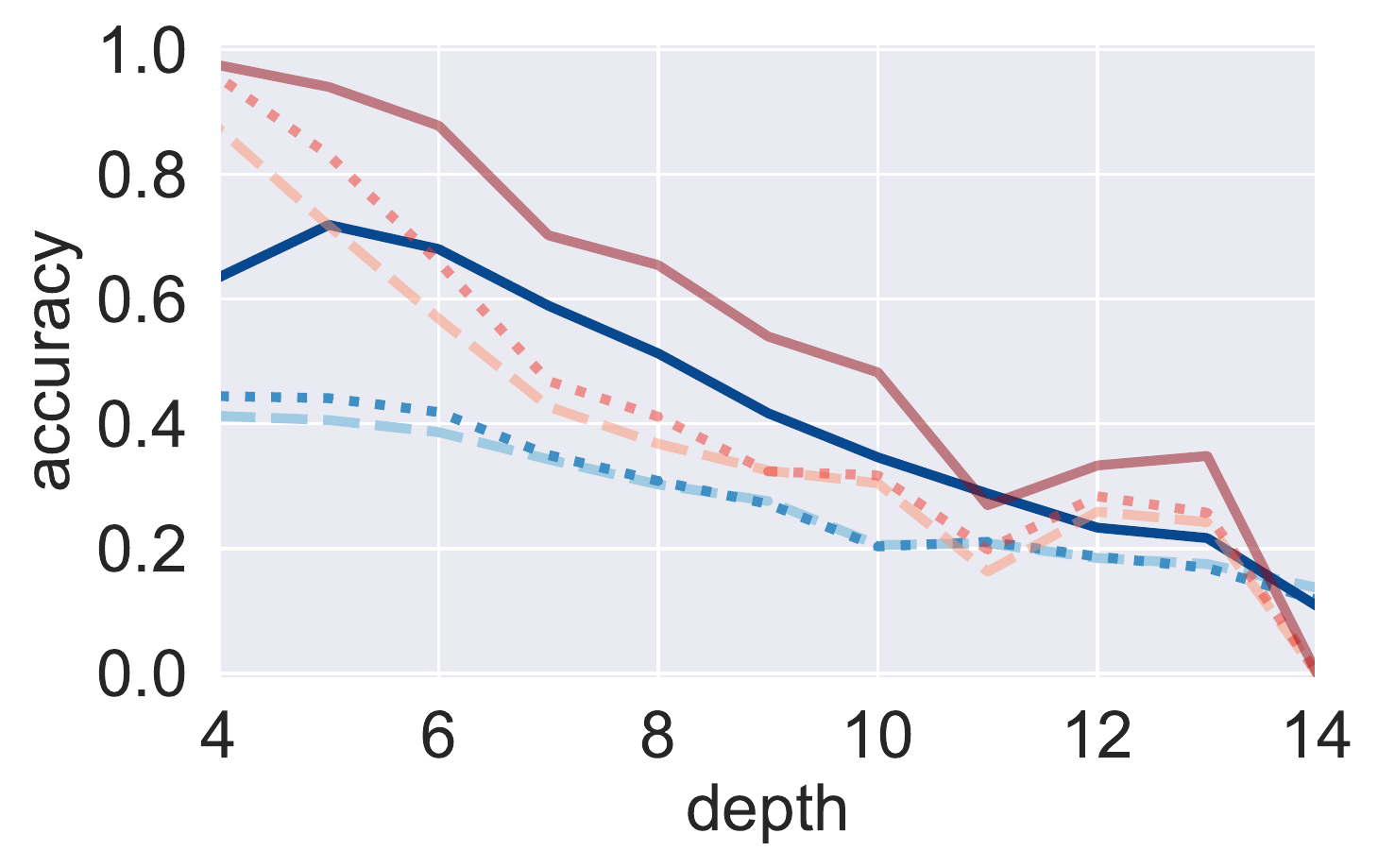}
    \end{subfigure}
    \begin{subfigure}{0.31\textwidth}
    \includegraphics[width=\textwidth]{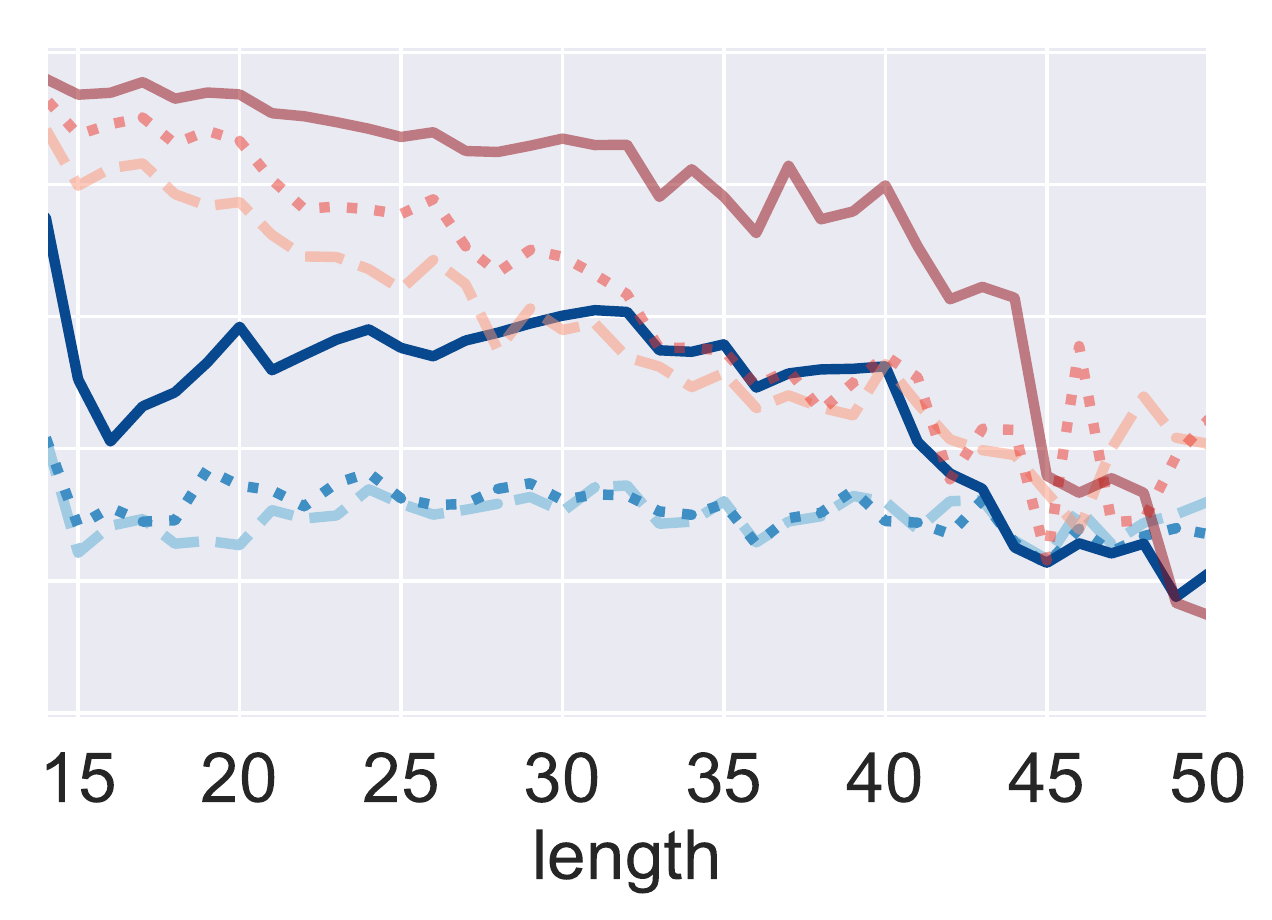}
    \end{subfigure}
    \begin{subfigure}{0.31\textwidth}
    \includegraphics[width=\textwidth]{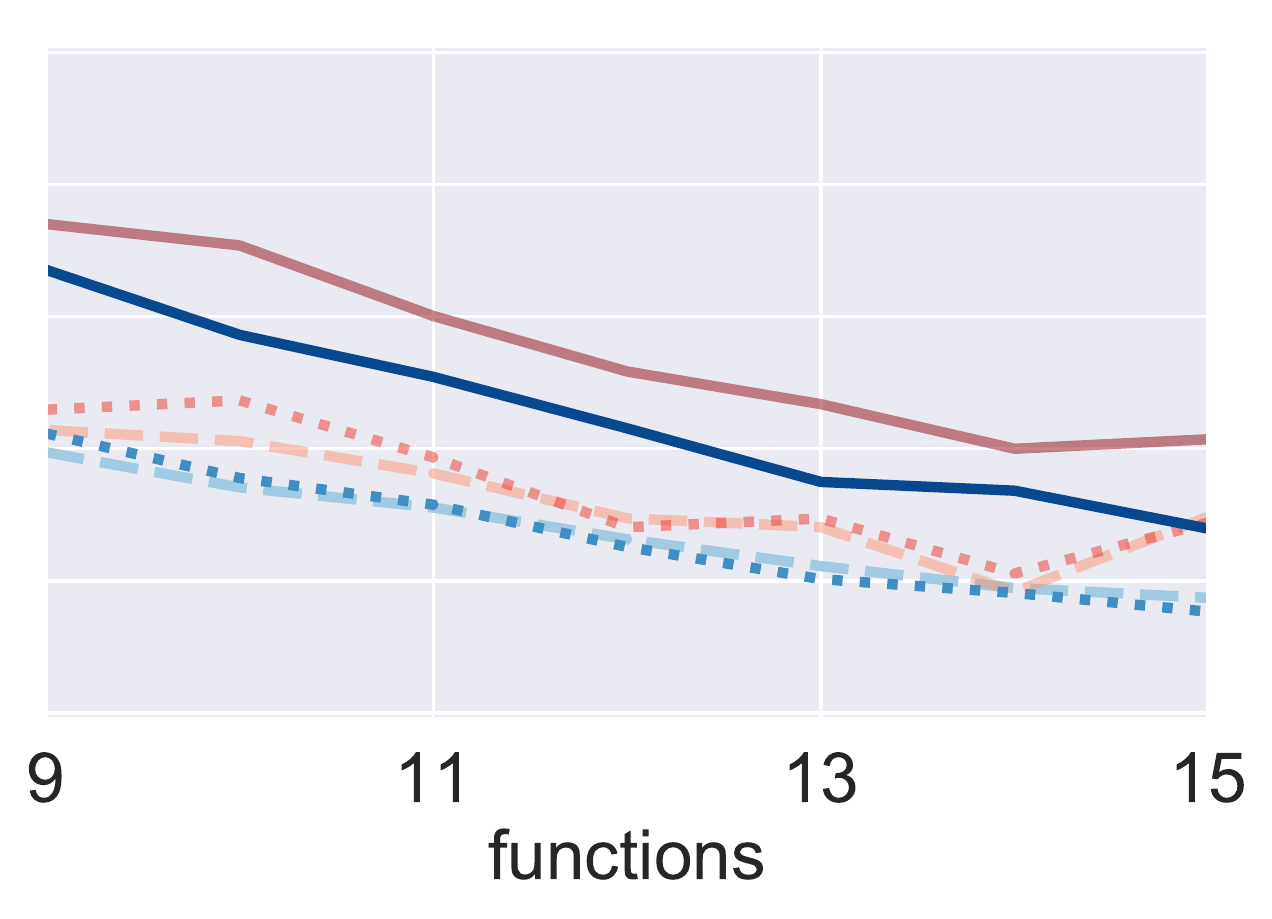}
    \end{subfigure}
    \caption{General task accuracy (in red) and accuracy of the three architectures on the productivity test set (in blue) as a function of several properties of the input sequences: the \textit{depth} of the input's parse tree, the input sequence's \textit{length} and the \textit{number of functions} in the input sequence.
    All results are averaged over three runs and computed over 11 thousand test samples.}
    \label{fig:productivity_depth_length_tokens}
\end{figure}

Comparing just the task accuracy and productivity accuracy of models shows that models have difficulty with longer sequences but does still not give a definitive answer about the source of the performance decrease.
Since the productivity test set contains on average longer sequences, we cannot see if the drop in performance is caused by poor productive power or by the inherent difficulty of longer sequences.
In Figure~\ref{fig:productivity_depth_length_tokens}, we show the performance of the three models in relation to depth, length and number of functions of the input sequences (blue lines) compared with the task accuracy of the standard \pcfg test data for the same lengths as plotted in Figure~\ref{fig:depth_length_tokens}.
For all models, the productivity scores are lower for almost every depth, length and number of functions.
This decrease in performance is solely caused by the decrease in evidence for such sequences: the total number of examples that models were trained on is roughly the same across the two conditions, and the absolute difficulty of the longer sequences is as well.
With these two components factored out, we conclude that models in fact struggle to productively generalise to longer sequences.\footnote{To stop their generation of the answer, models have to explicitly generate an \emph{end of sequence} symbol (\texttt{<eos>}).
A reasonable hypothesis concerning the low scores on longer sequences is that they are due to models' inability to postpone the emission of this \texttt{<eos>} symbol.
Following \citet{dubois2019location}, we call this problem the \texttt{<eos>}-\texttt{problem}.
To test whether the low scores are due to early \texttt{<eos>} emissions, we compute how many of the wrongly emitted answers were contained in the right answer.
For LSTMS2S, ConvS2S and Transformer this was the case in 22\%, 6\% and 8\% of the wrong predictions.
These numbers illustrate that the \texttt{<eos>}-problem indeed exists, but is not the main source of the poor productive capacity of the different models.

}

The depth plot in Figure~\ref{fig:depth_length_tokens} also provides some evidence for the inherent difficulty of deeper functions: it shows that all models suffer from decreasing test accuracies for higher depths, even if these depths are well-represented in the training data. 
When looking at the number of functions, the productivity score of Transformer is worse than its overall task success for any considered number of functions.
The scores for LSTMS2S and ConvS2S are instead very similar to the ones they reached after training on the regular data. 
This shows that functions with high depths are difficult for LSTMS2S and ConvS2S, even when some of them are included in the training data.

Interestingly, considering only the development of the productivity scores (in blue), it appears that both the LSTMS2S and ConvS2S are relatively insensitive to the increasing length as measured by the number of tokens.
Their performance is just as bad for input sequences with 20 or 50 characters, which is on a par with the scores they obtain on the longest sequences after training on the regular data.
Apparently, shorter sequences of unseen lengths are as challenging for these models as sequences of extremely long lengths.
Later, in the localism experiment, we will find more evidence that this sharp difference between seen and unseen lengths is not accidental for LSTMS2S but characteristic for the representations learned by this architecture.

\subsection{Substitutivity}\label{subsec:substitutivity_results}

While the previous two experiments were centred around models' ability to recombine known phrases and rules to create new phrases, we now focus on the extent to which models are able to draw analogies between words.
In particular, we study under what conditions models treat words as \textit{synonyms}: we consider what happens when synonyms are \emph{equally distributed} in the input sequences and when one of the synonyms only occurs in \emph{primitive contexts}.

\subsubsection{Test details}
We select two binary and two unary functions (\texttt{swap}, \texttt{repeat}, \texttt{append} and \texttt{remove\_second}), for which we artificially introduce synonyms during training: \texttt{swap\_syn}, \texttt{repeat\_syn}, \texttt{append\_syn} and \texttt{remove\_second\_syn}.
Like in the systematicity test, we keep those four functions fixed across all experiments, varying only the model initialisation and order of presentation of the training data.
The introduced synonyms have the same interpretation functions as the terms they substitute, so they are semantically equivalent to their counterparts.
We consider two different conditions that differ in the syntactic distribution of the synonyms in the training data.

\paragraph{Equally distributed synonyms}
For the first substitutivity test we randomly replace half of the occurrences of the chosen functions $F$ with $F_{syn}$, keeping the target constant.
On average, the individual functions appeared in 39\% of the training samples.
After synonym substitution, they appear in approximately 19\% of the training samples, on average.
In this test, $F$ and $F_{syn}$ are distributionally similar, which should facilitate inferring that they are synonyms.

\paragraph{Primitive synonyms}
In the second and more difficult substitutivity test, we introduce $F_{syn}$ only in \textit{primitive} contexts, where $F$ is the only function call in the input sequence.
$F_{syn}$ is introduced in 0.1\% of the training set samples. 
In this \textit{primitive} condition, the function $F$ and its synonymous counterpart $F_{syn}$ are distributionally not equivalent

\paragraph{Evaluation}
For the substitutivity test, we do not evaluate models' accuracy but assess their robustness to meaning-invariant synonym substitutions in the input sequence.
The most important point is not whether a model correctly predicts the target for an adapted input sequence, but whether its prediction matches the prediction it made before the transformation.
We evaluate models based on this interchangeability of $F$ with $F_{syn}$.
We quantify this with a \emph{consistency score}, which expresses a pairwise equality, where a model's outputs on two different inputs are compared to each other, instead of to the target output.
As with accuracy, also here only instances for which there is a complete match between the compared outputs are considered correct.

The consistency metric allows us to evaluate compositionality aspects isolated from task performance.
Even for models that may not have a near-perfect task performance and therefore have not mastered the rules underlying the data, we want to evaluate whether they consistently apply and generalise the knowledge they did acquire.
We use the consistency score for the current substitutivity test and later for the localism tests. 
In the next sections, consistency scores are marked with $\dagger$.

\subsubsection{Equally distributed substitutions}
For the substitutivity experiment where words and synonyms are equally distributed, Transformer and ConvS2S perform nearly on par.
They both obtain a very high consistency score (0.98 and 0.95, respectively).
In Table~\ref{tab:pcfg_embedding_distances}, we see that both architectures put words and their synonyms closely together in the embedding space, truly respecting the distributional hypothesis. 
Surprisingly, LSTMS2S does not identify that two words are synonyms, even in this relatively simple condition where the words are distributionally identical.
Words and synonyms are at very distinct positions in the embedding space, although the distance between them is smaller than the average between all words in the embedding space.
We hypothesise that this low score of the LSTMS2S reflects the architecture's inability to draw the type of analogies required to model \pcfg data, which is also mirrored in its relatively low overall task accuracy.

\begin{table}[t]
\centering
\begin{tabular}{@{}lccccccccccc@{}}
\toprule
& \multicolumn{3}{c}{\textbf{LSTMS2S}} && \multicolumn{3}{c}{\textbf{ConvS2S}} && \multicolumn{3}{c}{\textbf{Transformer}}  \\
\textbf{Token}          & ED   & P   & Other && ED   & P   & Other && ED   & P   & Other \\  \midrule \midrule
\texttt{repeat}         & 0.46 & 0.41 & 0.96 && 0.10 & 0.41 & 0.84 && 0.08 & 0.39 & 0.79 \\
\texttt{remove\_second} & 0.27 & 0.28 & 0.94 && 0.16 & 0.60 & 0.86 && 0.08 & 0.34 & 0.79 \\
\texttt{swap}           & 0.35 & 0.33 & 0.93 && 0.17 & 0.38 & 0.88 && 0.08 & 0.39 & 0.79 \\
\texttt{append}         & 0.34 & 0.29 & 1.00 && 0.12 & 0.54 & 0.82 && 0.07 & 0.37 & 0.75 \\\midrule
\emph{Average}          & 0.36 & 0.33 & 0.96 && 0.14 & 0.48 & 0.85 && 0.08 & 0.37 & 0.78 \\
\midrule \midrule 
\textbf{Consistency}    & \textbf{0.80} & \textbf{0.60} &&& \textbf{0.95} & \textbf{0.58} &&& \textbf{0.98} & \textbf{0.90} \\
\bottomrule
\end{tabular}
\caption{The average cosine distance between the embeddings of the indicated functions and their synonymous counterparts in the equally distributed (ED) and primitive (P) setups of the substitutivity experiments. For comparison, the average distance from the indicated functions to all other regular function embeddings is given under `Other'.
As those distances were very similar across conditions, we averaged them in one column instead of showing them separately.}
\label{tab:pcfg_embedding_distances}
\end{table}

\subsubsection{Primitive substitutions}
The primitive substitutivity test is substantially more challenging than the equally distributed one, since models are only shown examples of synonymous expressions in a small number of primitive contexts.
This implies that words and their synonyms are no longer distributionally similar and that models are provided much fewer evidence for the meaning of synonyms, as there are simply fewer primitive than composed contexts.

While the consistency scores for all models decrease substantially compared to the equally distributed setup, all models do pick up that there is a similarity between a word and its synonym.
This is reflected not only in the consistency scores (0.60, 0.58 and 0.90 on average for LSTMS2S, ConvS2S and Transformer, respectively), but is also evident from the distances between words and their synonyms, which are substantially lower than the average distances to other function embeddings (Table~\ref{tab:pcfg_embedding_distances}).
For LSTMS2S, the average distance is very comparable to the average distance observed in the equally distributed setup.
Its consistency score, however, goes down substantially, indicating that word distances (computed between embeddings) give an incomplete picture of how well models can account for synonymity when there is a distributional imbalance. 

\paragraph{Synonymity vs few-shot learning}
The consistency score of the primitive substitutivity test reflects two skills that are partly intertwined: the ability to few-shot learn the meanings of words from very few samples and the ability to bootstrap information about a word from its synonym.
As already observed in the equally distributed experiment for LSTMS2S, it is difficult to draw hard conclusions about a model's ability to infer synonymity when it is not able to infer consistent meanings of words in general.
When a model has a high score, on the other hand, it is difficult to disentangle if it achieved this high score because it has learned the correct meaning of both words separately, or because it has in fact understood that the meaning of those words is similar.
That is: the consistency score does not tell us whether output sequences are identical because the model knows they should be the \textit{same}, or simply because they are both \textit{correct}.
In the equally distributed setup, the low word embedding distances for the ConvS2S and the Transformer strongly pointed to the first explanation.
For the primitive setup, the two aspects are more difficult to take apart.

\begin{table}
    \centering
    \begin{tabular}{lcccccccccccc}
    \toprule
    & \textbf{LSTMS2S} & \textbf{ConvS2S} & \textbf{Transformer} \\ \midrule \midrule
    Consistency across all               & 0.60 \pms{0.01} & 0.58 \pms{0.01} & 0.90 \pms{0.00} \\
    Consistent correct                   & 0.53 \pms{0.01} & 0.53 \pms{0.01} & 0.84 \pms{0.00} \\
    Consistent incorrect                 & 0.07 \pms{0.01} & 0.04 \pms{0.00} & 0.05 \pms{0.00} \\
    Consistency across incorrect samples & 0.14 \pms{0.01} & 0.09 \pms{0.01} & 0.34 \pms{0.02} \\
    \bottomrule
    \end{tabular}
    \caption{Consistency scores for the primitive substitutivity experiment, expressing pairwise equality for the outputs of synonymous sequences. 
    We show the overall consistency (\emph{consistency across all}), the consistency of sequences for which the model's output was correct (\emph{consistent correct}), sequences for which the model's output was incorrect (\emph{consistent incorrect}), and the percentage of all incorrect predictions that were consistent.
A pair is considered incorrect if at least one of its parts is incorrect.
NB: \textit{consistent correct} and \textit{consistent incorrect} together sum up to \textit{consistent across all}; due to rounding, this is not the case in all columns of the table.}
    \label{tab:consistency_breakdown}
\end{table}

\paragraph{Error consistency} 
To separate a model's ability to few-shot learn the meaning of a word from very few primitive examples and its ability to bootstrap information about synonyms
, we compute the consistency score for model outputs that do not match the target output (\emph{incorrect outputs}).
When a model makes identical but incorrect predictions for two input sequences with a synonym substitution, this cannot be because the model merely correctly learned the meanings of the two words.
It can thus be taken as evidence that it treats the word and its synonyms indeed as synonyms.

In Table~\ref{tab:consistency_breakdown}, we show the consistency scores for all output pairs (identical to the scores in Table~\ref{tab:pcfg_results}), the breakdown of this score into correct (\emph{consistent correct}) and incorrect (\emph{consistent incorrect}) output pairs, and the ratio of incorrect output pairs that is consistent.
The scores in row two and three show that the larger part of the consistency scores for all models is due to correct outputs.
In row 4, we see that models are seldom consistent on \textit{incorrect} outputs.
The Transformer maintains its first place
, but none of the architectures can be said to treat a word and its synonymous counterpart as true synonyms.
An interesting difference occurs between LSTMS2S and ConvS2S, whose consistency scores on all outputs are similar, but differ in consistency of erroneous outputs.
These scores suggest that ConvS2S is better at few-shot learning than LSTMS2S, but LSTMS2S is better at inferring synonymity.
These results are in line with the embedding distances shown for the primitive substitutivity experiment in Table~\ref{tab:pcfg_embedding_distances}, which are on average also lower for LSTMS2S than for ConvS2S.

\subsection{Localism}\label{subsec:localism_results}

In the localism test, we investigate whether models compute the meanings of input sequences using local composition operations, following the hierarchical trees that specify their compositional structure.

\subsubsection{Test details}
\begin{figure}
\centering
\begin{tikzpicture}[scale=0.9, every node/.style={scale=0.9}]
    % box around prepend B A
    \path[fill=blue_, opacity=0.5] (1,-3.6) rectangle (3.35,-2.05);
    \draw[->] (3.4, -3) to (6, -3);
    \node[draw, minimum width=1cm, fill=orange_, rectangle, align=center] at (6.7, -3) {\textsf{Model}};
    
    % box around append C AB
    \path[fill=blue_, opacity=0.5] (4.5, -2.5) rectangle (7.2, -1); %-1
    \draw[dashed, fill=blue_, opacity=0.5] (6.3, -2.2) rectangle (7, -1.8);
    \draw[->] (7.3, -1.7) to (8.9, -1.7);
    \node[draw, minimum width=1cm, fill=orange_, rectangle, align=center] at (9.6, -1.7) {\textsf{Model}};
    
    % box around echo CAB
    \path[fill=blue_, opacity=0.5] (8, -1.2) rectangle (10.2, 0.2);
    \draw[dashed, fill=blue_, opacity=0.5] (9, -1) rectangle (10, -0.6);
    \draw[->] (10.3, -0.5) to (11.3, -0.5);
    \node[draw, minimum width=1cm, fill=orange_, rectangle, align=center] at (12, -0.5) {\textsf{Model}};
    \draw[->] (12, -0.2) to (12, 0.4);
    \node[] at (12, 0.7) {\texttt{C A B B}};

    \draw[->] (6.7, -2.7) to (6.7, -2.3);
    \draw[->] (9.6, -1.4) to (9.6, -1.05);

    \begin{scope}[xshift=0cm, yshift=0cm]
    \tikzset{frontier/.style={distance from root=90pt}}
    \Tree[.\node[inner sep=0pt,outer sep=-0.4pt]  {}; 
        [.\node[inner sep=0pt,outer sep=-0.4pt] {}; 
            \texttt{echo}
        ]
        [.\node[inner sep=0pt,outer sep=-0.4pt] {};
            [.\node[inner sep=0pt,outer sep=-0.4pt] {}; 
                \texttt{append} 
            ]
            [.\node[inner sep=0pt,outer sep=-0.4pt] {}; 
                \texttt{C}
            ]
            [.\node[inner sep=0pt,outer sep=-0.4pt] {}; 
                %[.\node[inner sep=0pt,outer sep=-0.4pt] {}; 
                \texttt{prepend}
                %]
                %[.\node[inner sep=0pt,outer sep=-0.4pt] {}; 
                \texttt{B}
                %]
                %[.\node[inner sep=0pt,outer sep=-0.4pt] {}; 
                \texttt{A}
                %]
            ]
        ]
    ]
    \end{scope}
    
    \begin{scope}[xshift=5cm, yshift=0cm]
    \tikzset{frontier/.style={distance from root=60pt}}
    \Tree[.\node[inner sep=0pt,outer sep=-0.4pt]  {}; 
        [.\node[inner sep=0pt,outer sep=-0.4pt] {}; 
            \texttt{echo}
        ]
        [.\node[inner sep=0pt,outer sep=-0.4pt] {};
            \texttt{append} 
            \texttt{C}
            \texttt{A B}
        ]
    ]
    \end{scope}
    
    \begin{scope}[xshift=9cm, yshift=0cm]
    \tikzset{frontier/.style={distance from root=25pt}}
    \Tree[.\node[inner sep=0pt,outer sep=-0.4pt]  {}; 
        \texttt{echo}
        \texttt{C A B}
    ]
    \end{scope}
\end{tikzpicture}

\caption{An example of the unrolled computation of the meaning of the sequence \texttt{echo append C , prepend B , A} for the localism test.
We unroll the computation of the meaning of the sequence by first asking the model to compute the meaning $o_1$ of the smallest constituent \texttt{prepend B , A} and then replace the constituent by this predicted meaning $o_1$.
In the next step, we use the model to compute the meaning of the then smallest constituent \texttt{echo $o_1$}, and replace the constituent in the sequence with the model's prediction for this constituent.
This process is repeated until the meaning of the entire sequence is computed, in steps, by the model.
This final prediction (\texttt{C A B B} in the picture) is then compared with the model's prediction on the entire sequence (not shown in the picture).
If a model follows a local compositional protocol to predict the meaning of an output sequence, these two outputs should be the same.
}
\label{fig:localism_example}
\end{figure}
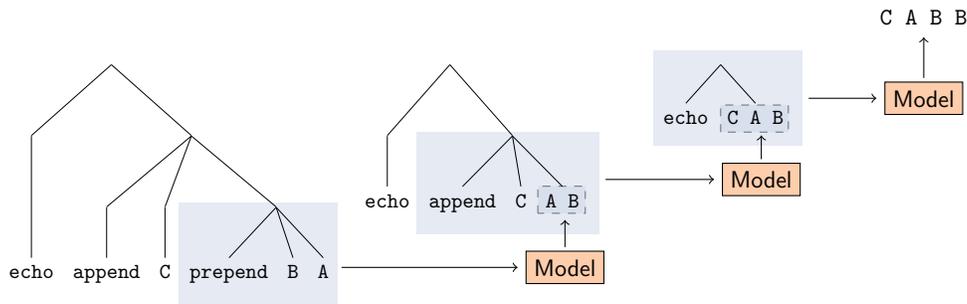

We test for localism by considering models' behaviour when a subsequence in an input sequence is replaced with its meaning (see Figure~\ref{fig:localism_example} for an example).
Thanks to the recursive nature of the \pcfg expressions and interpretation functions, this is a relatively straightforward substitution in our data. 
If a model uses local composition operations to build up the meanings of input sequences, following the hierarchy that it is dictated by the underlying system, its output meaning should not change as a consequence of such a substitution.

\paragraph{Unrolling computations}
We compare the output sequence that is generated by a model for a particular input sequence with the output sequence that the same model generates when we explicitly unroll the processing of the input sequence.
That is, instead of presenting the entire input sequence to the model at once, we force the model to evaluate the outcome of smaller constituents before computing the outcome of bigger ones, in the following way:
we iterate through the syntactic tree of the input sequence and use the model to compute the meanings of the smallest constituents.
We then replace these constituents by the model's output and use the model to again compute the meanings of the smallest constituents in this new tree.
This process is continued until the meaning for the entire sequence is found.
A concrete example is visualised in Figure~\ref{fig:localism_example}.

We conduct the localism test on sentences from the \pcfg test set.
On average, unrolling the computation of these sequences involves five steps.

\paragraph{Evaluation}
We evaluate a model by comparing the final output of the enforced recursive method to the output emitted when the sequence is presented in its original form.
Again, during evaluation we focus on checking whether the two outputs are identical, rather than if they are correct.
If a model wrongfully emits \texttt{B A} for input sequence \texttt{prepend B , A}, this is not penalised in this experiment, provided that the regular input sequence yields the same prediction as its hierarchical variant.
This method of evaluation matches the previously mentioned \emph{consistency score} that was also used in the previous section for the substitutivity test.

\subsubsection{Results}
None of the evaluated architectures obtains a high consistency score for this experiment (0.46, 0.59 and 0.54 for LSTMS2S, ConvS2S and Transformer, respectively).
Also in this test, Transformer ranks high, but the best-performing architecture is ConvS2S (significant in comparison with both LSTMS2S and Transformer with $p \approx 10^{-4}$ and $p \approx 10^{-2}$, respectively).
Since the ConvS2S models are explicitly using local operations, this is in line with our expectations.

\paragraph{Input string length}
To understand the main cause of the relatively low scores on this experiment, we manually analyse 300 samples (100 per model type), in which at least one mistake was made during the unrolled processing of the sample.
We observe that the most common mistakes involve unrolled samples that contain function applications to string inputs with more than five characters.
An example of such a mistake would be a model that is able to compute the meaning of \texttt{reverse echo A B C D E} but not the meaning of \texttt{reverse A B C D E E}.
As the outputs for these two phrases are identical, it is clear that this inadequacy does not stem from models' inability to generate the correct output string.
Instead, it indicates that the model does not compute the meaning of \texttt{reverse echo A B C D E} by consecutively applying the functions \texttt{echo} and \texttt{reverse}.
We hypothesise that, rather, models generate representations for \textit{combinations} of functions that are then applied to the input string at once.

\paragraph{Function representations}
While developing `shortcuts' to apply combinations of functions all at once instead of explicitly unfolding the computation does not necessarily contradict compositional understanding -- imagine, for instance, computing the outcome of the sum \texttt{5 + 3 - 3} -- the results of the localism experiment do point to an interesting aspect of the learned representations.
Since unrolling computations mostly leads to mistakes when the character length of unrolled inputs is longer than the maximum character string length of five seen during training, it casts some doubt on whether the models have developed consistent function representations.

If a model truly understands the meaning of a particular function in \pcfg, it should in principle be able to apply this function to an input string of arbitrary length.
Note that, in our case, this ability does not require productivity in generating output strings, since the correct output sequences are not distributionally different from those in the training data (in some cases, they may even be exactly the same).
Contrary to in other setups, a failure to apply functions to longer sequence lengths can thus not be explained by distributional or memory arguments.
Therefore, the consistent failure of all models to apply functions to character strings that are longer than the ones seen in training suggests that, while models may have learned to adequately copy strings of length two to five, they do not necessarily consider those operations the same.

To check this hypothesis, we test all functions in a primitive setup where we vary the length of the string arguments they are applied to.\footnote{For binary functions, only one of the two string arguments exceeds the regular argument lengths.}
For a model that develops several length-specific representations for the same function, we expect the performance to go down abruptly when the input string length exceeds the maximum length seen during training.
If a model instead develops a more general representation, it should be able to apply learned functions also to longer input strings.
Its performance on longer strings may drop for other, practical, reasons, but this drop should be more smooth than for a model that has not learned a general-purpose representation at all.

\begin{figure}[!ht]
    \begin{subfigure}{0.35\textwidth}
        \includegraphics[width=\textwidth]{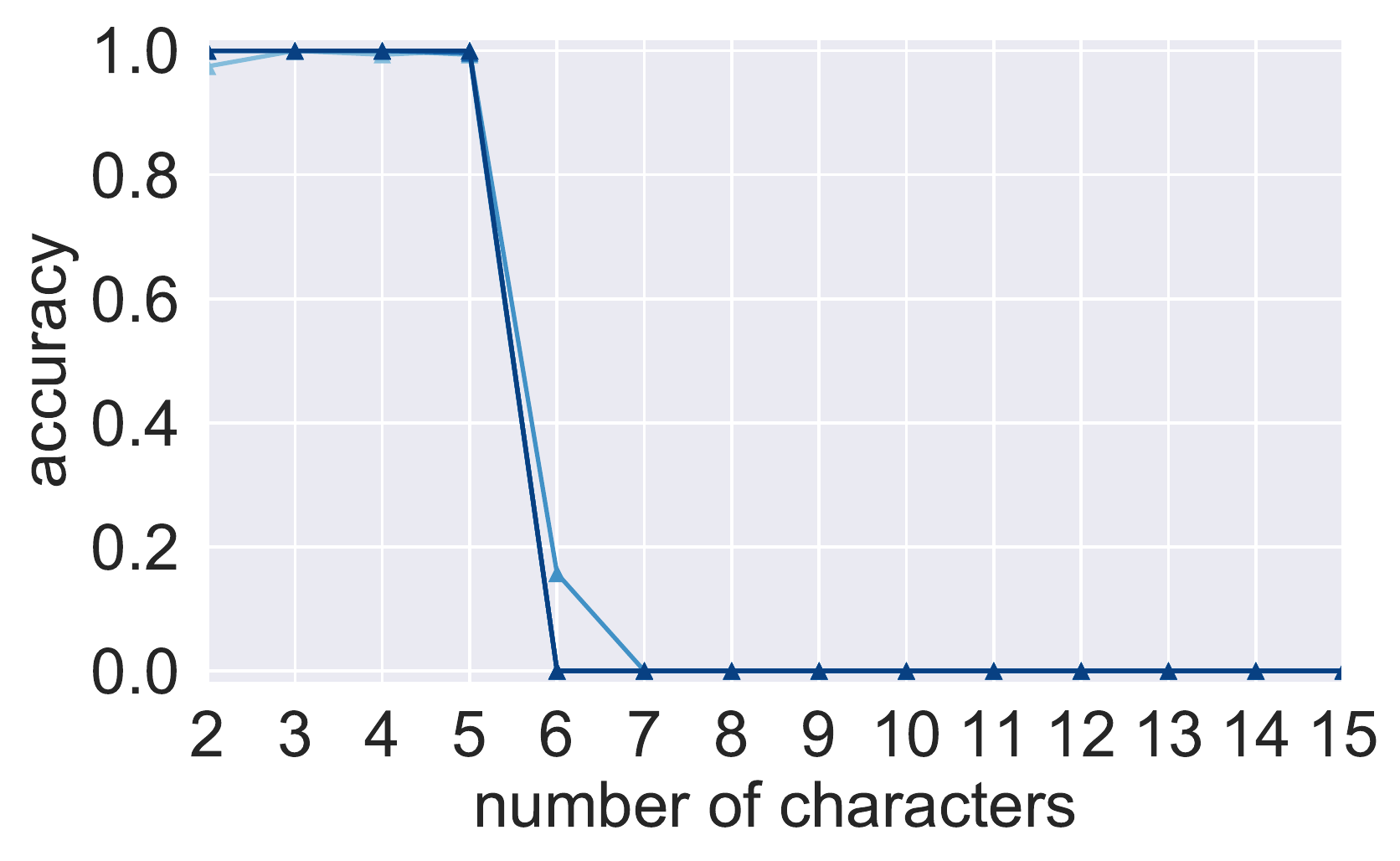}
        \caption{LSTMS2S}
    \end{subfigure}
    \begin{subfigure}{0.31\textwidth}
        \includegraphics[width=\textwidth]{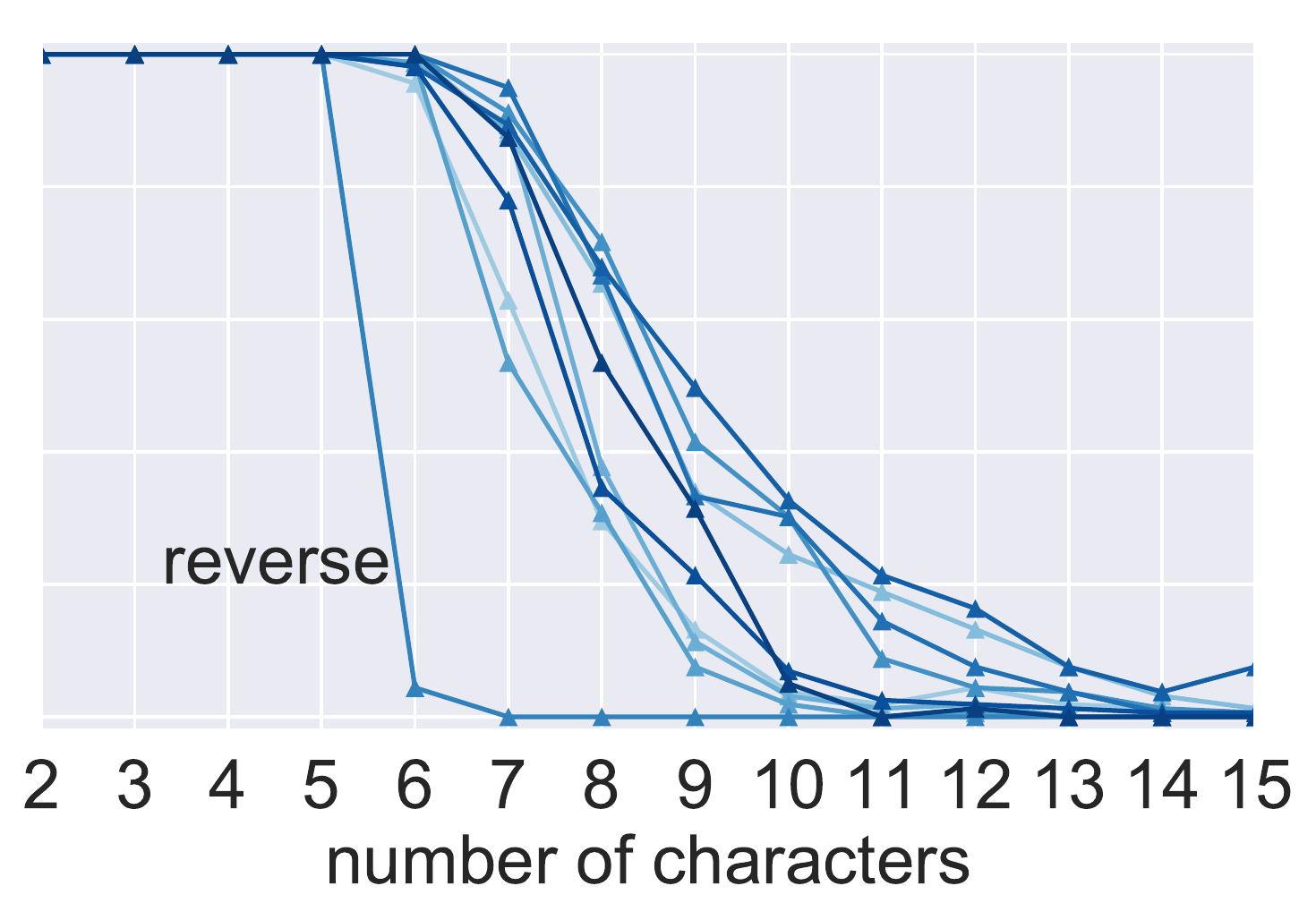}
        \caption{ConvS2S}
    \end{subfigure}
    \begin{subfigure}{0.31\textwidth}
        \includegraphics[width=\textwidth]{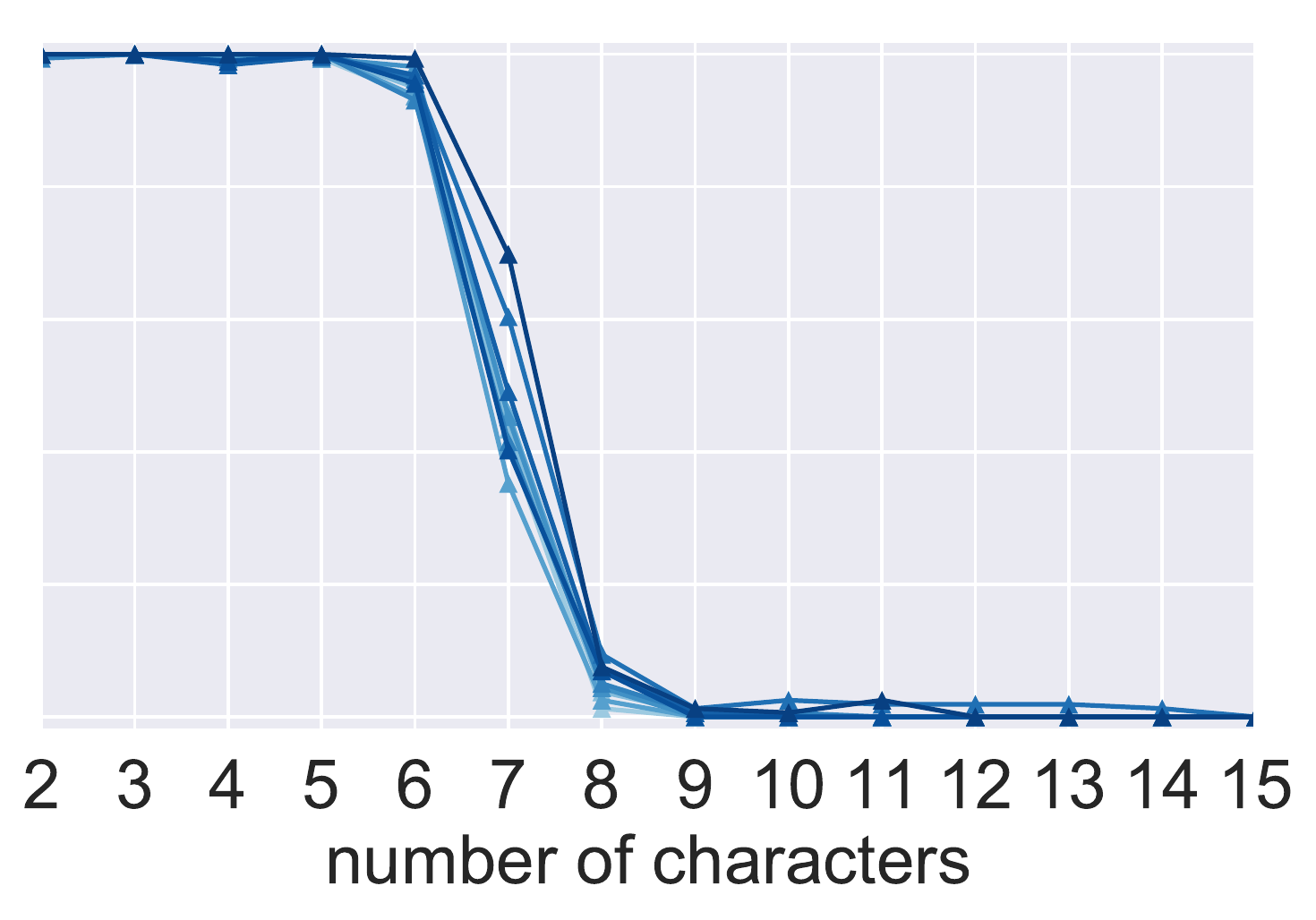}
        \caption{Transformer}
    \end{subfigure}
    \caption{Accuracy of the three architectures on different functions with increasingly long character string inputs.
        The maximum character string length observed during training is 5.
    While Transformer and ConvS2S can, for most functions, generalise a little beyond this string length, LSTMS2S models cannot.}
    \label{fig:localism_increasing_lengths}
\end{figure}

The results of this experiment, plotted in Figure~\ref{fig:localism_increasing_lengths}, demonstrate that all models have learned to apply all functions to input strings up until length five, as evidenced by their near-perfect accuracy on the samples of these lengths.
On longer lengths, however, none of the models performs well.
For all runs, the performance of LSTMS2S immediately drops to zero when string arguments exceed length five, the maximum string length seen during training.
The model does not seem to be able to leverage a general concept of any of the functions.
ConvS2S and Transformer do exhibit some generalisation beyond the maximum string input length seen during training, indicating that their representations are more general.
The accuracy of Transformer reaches zero only for input arguments of more than nine characters, ConvS2S outputs some correct responses even for input arguments of 12 or 13 characters.
This suggests that the descending scores may be due to factors of `performance' rather than `competence'. 
The accuracies for Transformer and ConvS2S are comparable for almost all functions, except \texttt{reverse}, for which the ConvS2S accuracy drops to almost zero for length six in all three runs.
Interestingly, none of the three architectures suffers from increasing the character length of the first and second argument to \texttt{remove\_first} and \texttt{remove\_second}, respectively (not plotted).

\subsection{Overgeneralisation}\label{subsec:overgeneralisation}

In our last test, we focus on the learning process, rather than on the final solution that is implemented by converged models.
In particular, we study if -- during training -- a model \textit{overgeneralises} when it is presented with an exception to a rule and -- in case it does -- how much evidence it needs to see to memorise the exception.
Whether a model overgeneralises indicates its willingness to prefer rules over memorisation, but while strong overgeneralisation characterises compositionality, more overgeneralisation is not necessarily better.
An optimal model, after all, should be able to deal with exceptions as well as with the compositional part of the data.

\subsubsection{Test details}

As the language defined through the PCFG is designed to be strictly compositional, it does not contain exceptions.
We therefore manually add them to the data set, which allows us to have a large control over their occurrence and frequency.

\paragraph{Exceptions} We select four pairs of functions that are assigned a new meaning when they appear together in an input sequence: \texttt{reverse echo}, \texttt{prepend remove\_first}, \texttt{echo remove\_first} and \texttt{prepend reverse}.
Whenever these functions occur together in the training data, we remap the meaning of those functions, as if an alternative set of interpretation functions is used in these few cases.
As a consequence, the model has no evidence for the \textit{compositional} interpretation of these function pairs, unless it overgeneralises by applying the rule observed in the rest of the training data.
For example, the meaning of \texttt{echo remove\_first A , B C} would normally be \texttt{B C C}, but has now become \texttt{A B C}.
The remapped definitions, which we call \textit{exceptions}, can be found in Table~\ref{tab:exceptions}.

\begin{table}[!h]
\centering
    \begin{tabular}{llll}
    \toprule
    \textbf{\small Input}   & \textbf{\small Remapped to}           & \multicolumn{2}{c}{\textbf{\small Target}} \\
                            &                                       & \textit{\small Original} & \textit{\small Exception}     \\ \midrule \midrule
    \texttt{\small reverse echo A B C}              & \texttt{\small echo copy A B C}       & \texttt{\small C C B A}       & \texttt{\small A B C C} \\
    \texttt{\small prepend remove\_first A , B , C} & \texttt{\small remove\_second append A , B , C} & \texttt{\small C B} & \texttt{\small A B} \\
    \texttt{\small echo remove\_first A , B C }     & \texttt{\small copy append A , B C }  & \texttt{\small B C C}         & \texttt{\small A B C} \\ 
    \texttt{\small prepend reverse A B , C}         & \texttt{\small remove\_second echo A B , C} & \texttt{\small C B A }  & \texttt{\small A B B} \\ \bottomrule
    \end{tabular}
    \caption{Examples for the overgeneralisation test. 
        The input sequences in the data set (first column, \textit{Input}) are usually presented with their ordinary targets (\textit{Original}). 
        In the overgeneralisation test, these input sequences are interpreted according to an alternative rule set (\textit{Remapped to}), effectively changing the corresponding targets (\textit{Exception}).}
    \label{tab:exceptions}
\end{table}

\paragraph{Exception frequency}
In our main experiment, the number of exceptions in the data set is 0.1\% of the number of occurrences of the least occurring function of the function pair $F_1 F_2$.
We present also the results of a grid-search in which we consider exception percentages of 0.01\%, 0.05\%, 0.1\% and 0.5\%.

% The number of exceptions added is 0.1\% of the number of occurrences of the least occurring symbol among $F_1$ and $F_2$. 

\subsubsection{Results}
We monitor the accuracy of both the original and the exception targets during training and compare how often a model correctly memorises the exception target and how often it overgeneralises to the compositional meaning, despite the evidence in the data.
To summarise a model's tendency to overgeneralise, we take the highest overgeneralisation accuracy that is encountered during training.
For more qualitative analysis, we visualise the development of both memorisation and overgeneralisation during training, resulting in \emph{overgeneralisation profiles}.
During training, we monitor the number of exception samples for which a model does not generate the correct meaning, but instead outputs the meaning that is in line with the rule instantiated in the rest of the data.
At every point in training, we define the strength of the overgeneralisation as the percentage of exceptions for which a model exhibits this behaviour.

\paragraph{Overgeneralisation peak}
We call the point in training where the overgeneralisation is highest the \textit{overgeneralisation peak}.
In Table~\ref{tab:pcfg_results}, we show the average height of this overgeneralisation peak for all three architectures, using an exception percentage of 0.1\%.
This quantity equals the accuracy of the model predictions on the input sequences whose outputs have been replaced by exceptions, but measured on the original targets that follow from the interpretation functions of \pcfg.
The numbers in Table~\ref{tab:pcfg_results} illustrate that all models show a rather high degree of overgeneralisation.
At some point during the learning process, Transformer applies the rule to 88\% of the exceptions and LSTMS2S and ConvS2S to 68\% and 79\% respectively.

\begin{figure}[t]
    \begin{tabular}{lccc}
    \% & \textbf{LSTMS2S} & \textbf{ConvS2S} & \textbf{Transformer} \\ 
    0.01 & \begin{subfigure}{0.318\linewidth}
        \includegraphics[width=\linewidth]{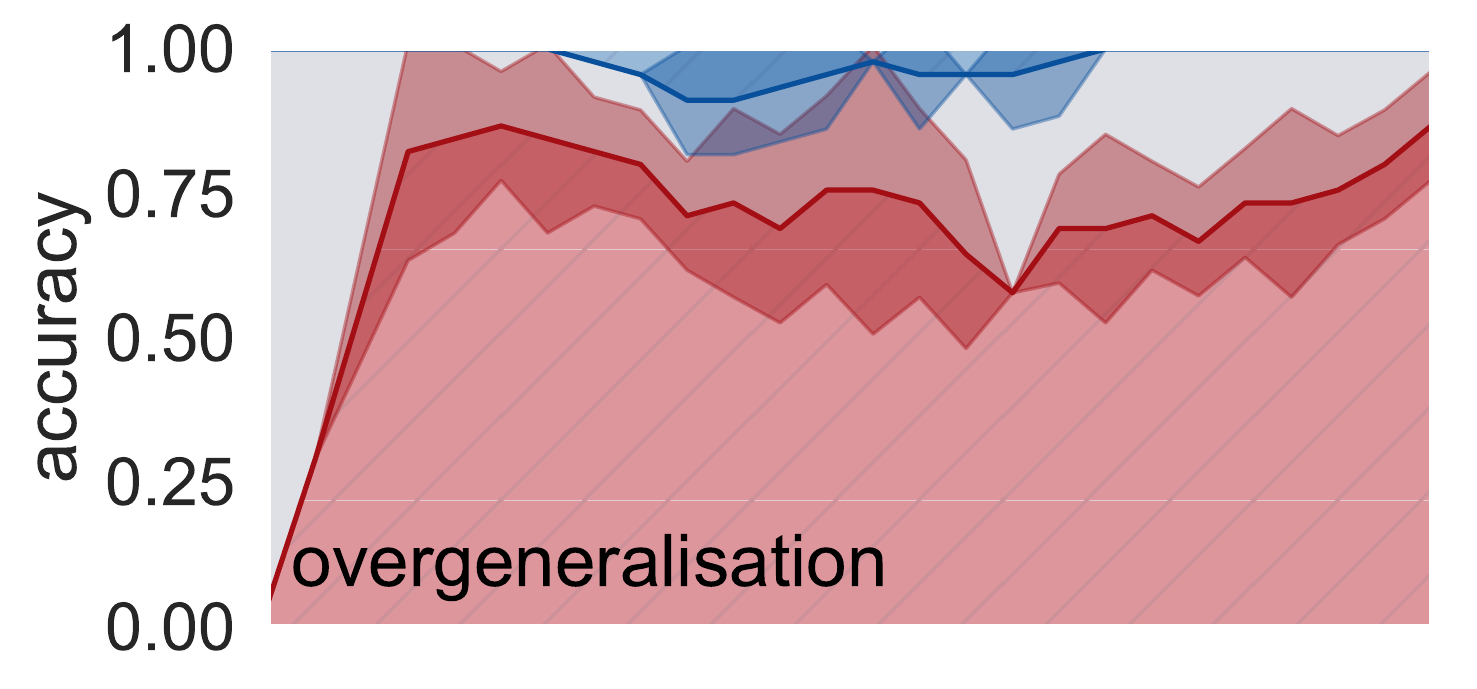}
    \end{subfigure} &
    \begin{subfigure}{0.27\linewidth}
        \includegraphics[width=\linewidth]{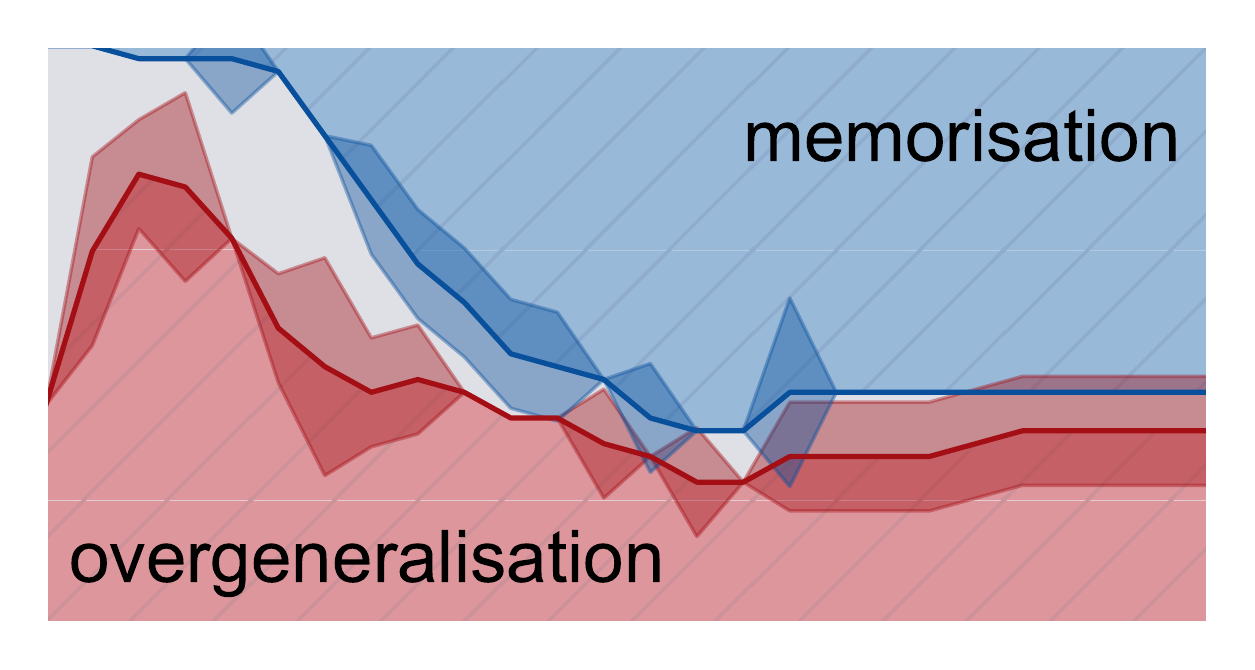}
    \end{subfigure} &
    \begin{subfigure}{0.27\linewidth}
        \includegraphics[width=\linewidth]{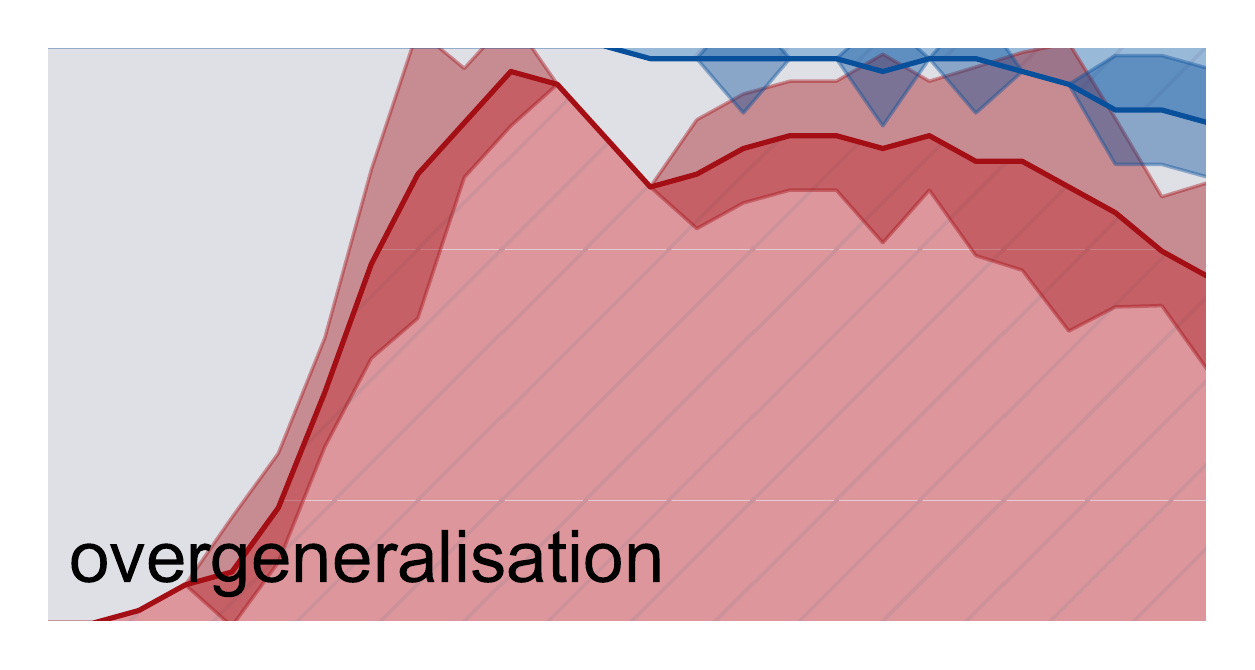}
    \end{subfigure} \\
    0.05 & \begin{subfigure}{0.318\linewidth}
        \includegraphics[width=\linewidth]{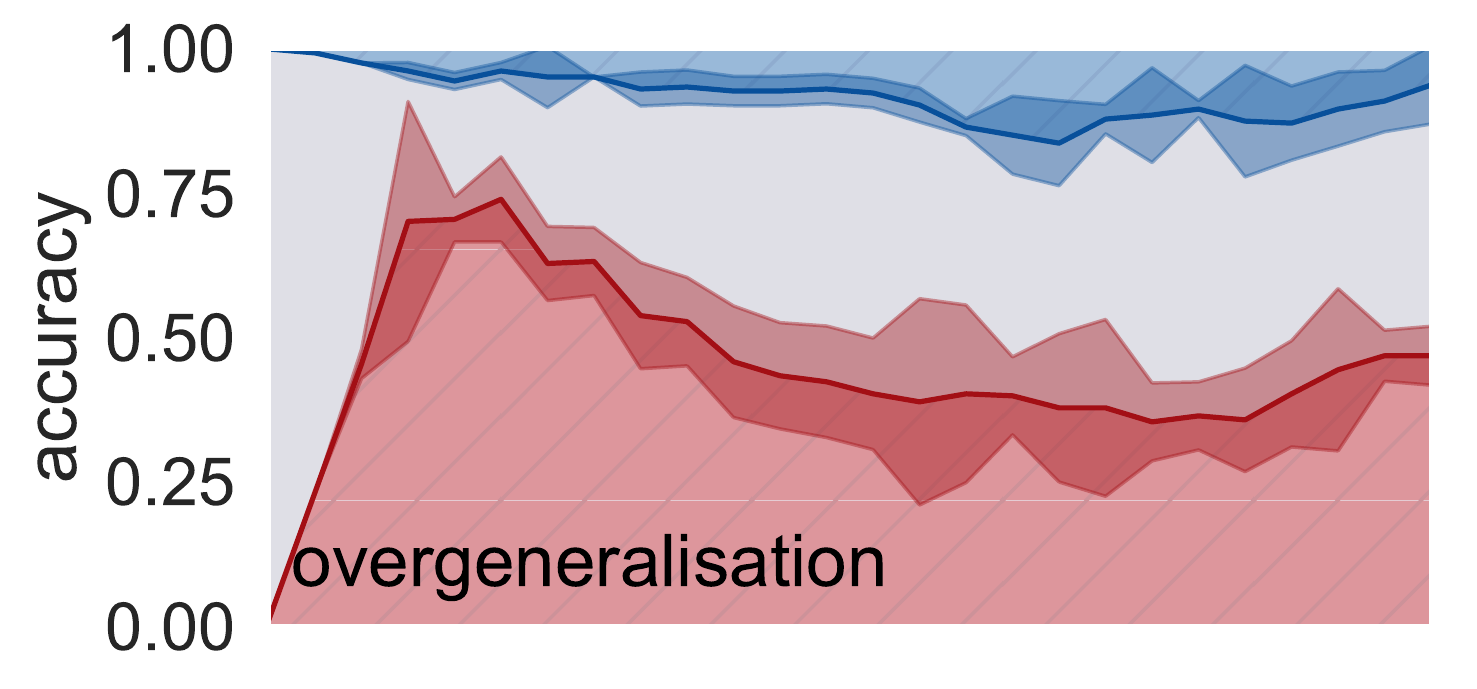}
    \end{subfigure} &
    \begin{subfigure}{0.27\linewidth}
        \includegraphics[width=\linewidth]{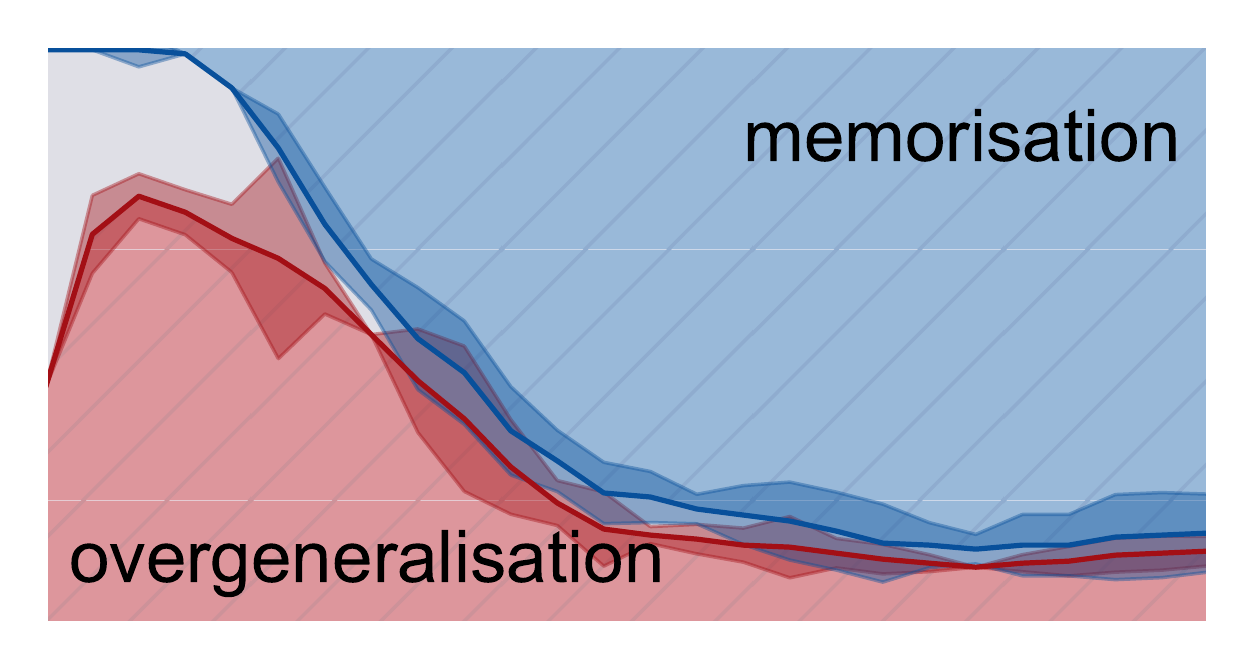}
    \end{subfigure} &
    \begin{subfigure}{0.27\linewidth}
        \includegraphics[width=\linewidth]{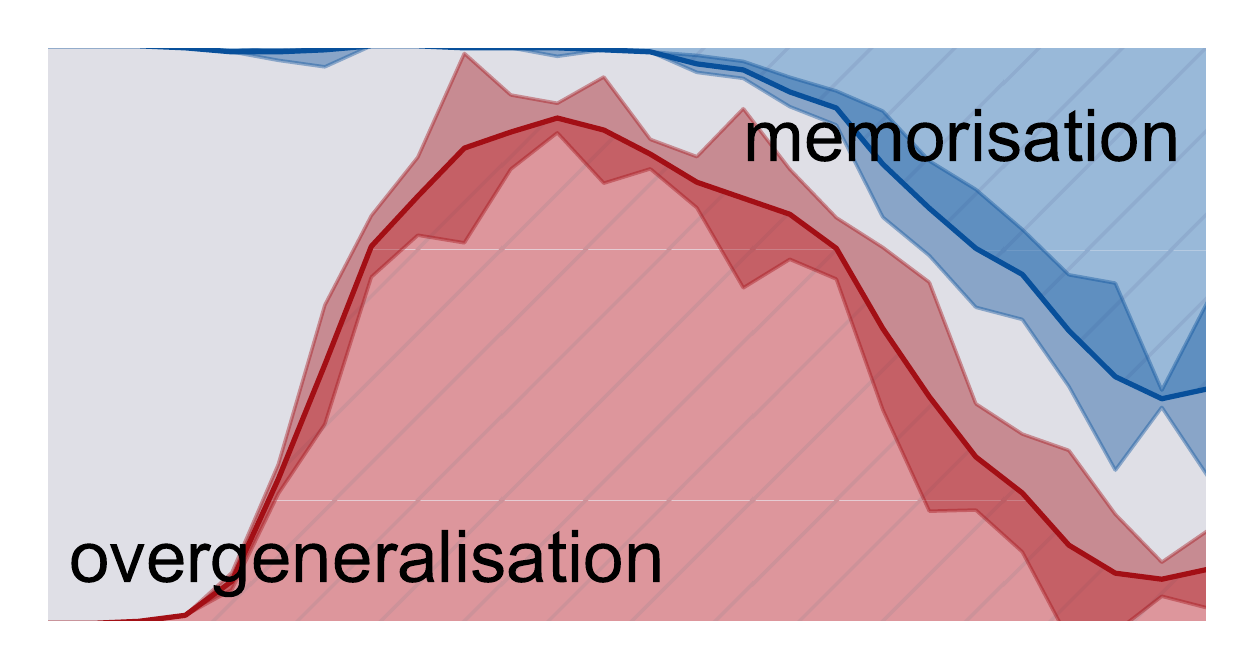}
    \end{subfigure} \\
    0.1 & \begin{subfigure}{0.318\linewidth}
        \includegraphics[width=\linewidth]{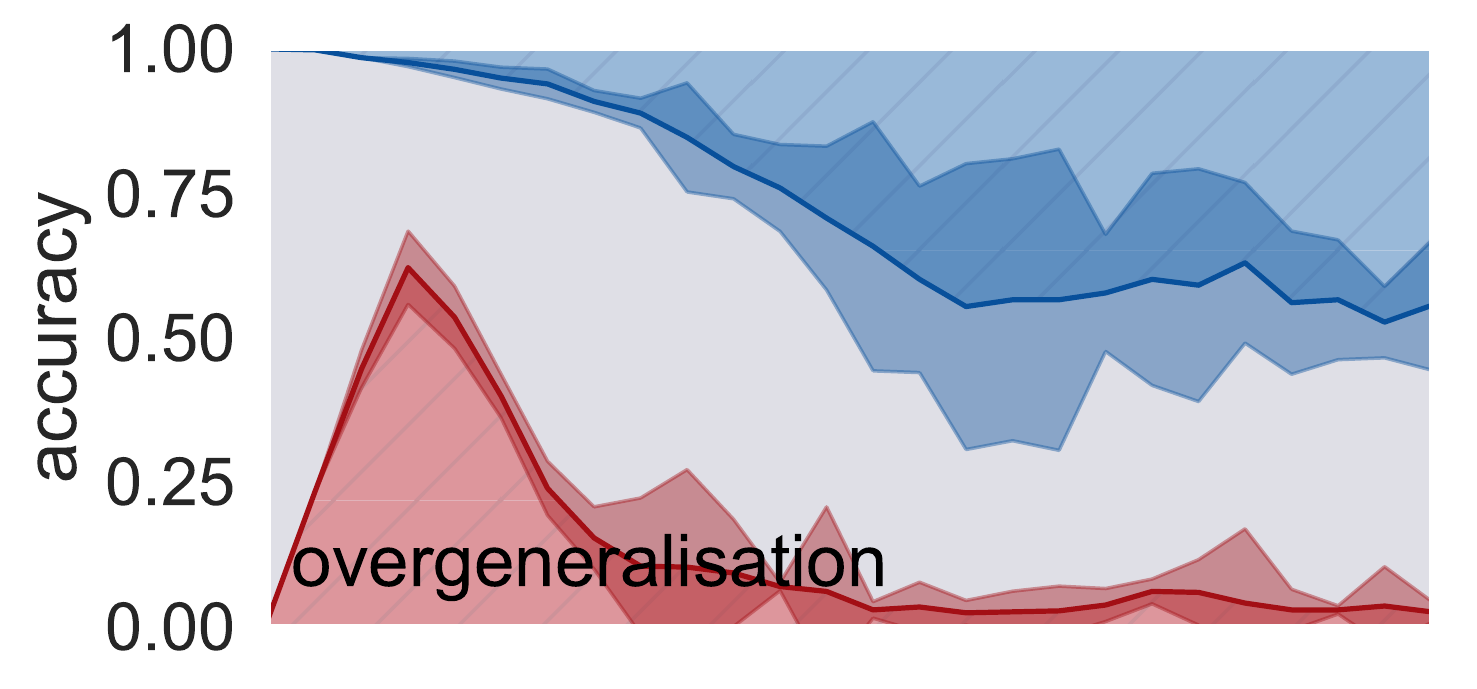}
    \end{subfigure} &
    \begin{subfigure}{0.27\linewidth}
        \includegraphics[width=\linewidth]{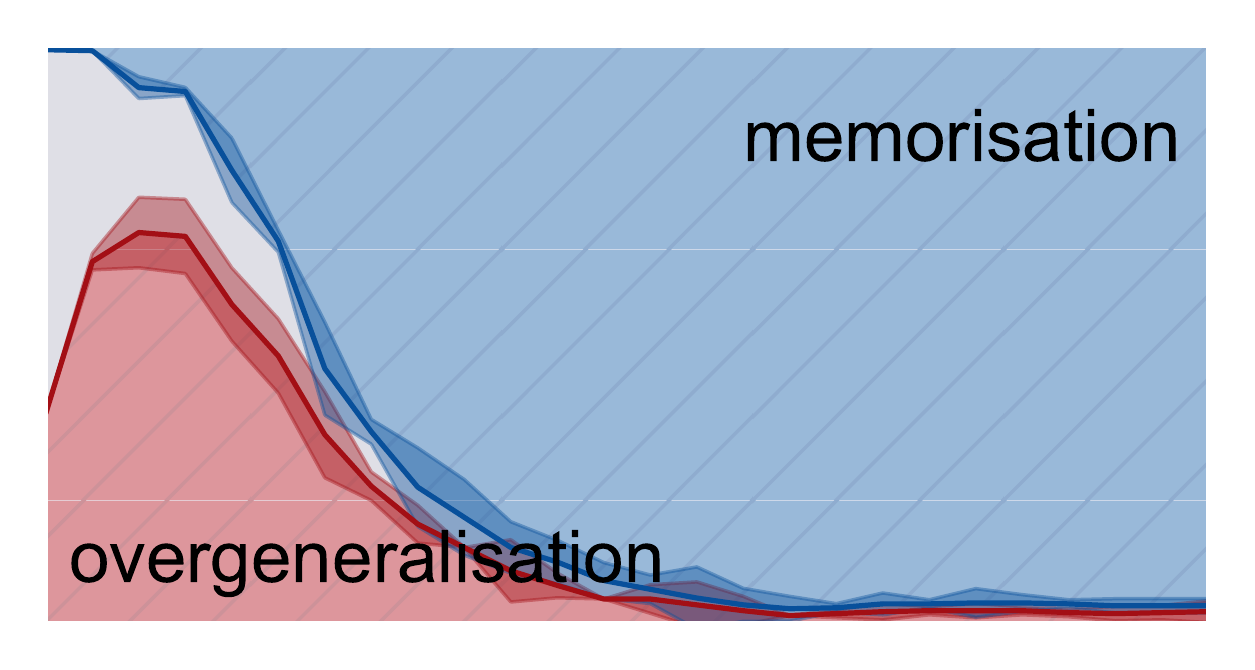}
    \end{subfigure} &
    \begin{subfigure}{0.27\linewidth}
        \includegraphics[width=\linewidth]{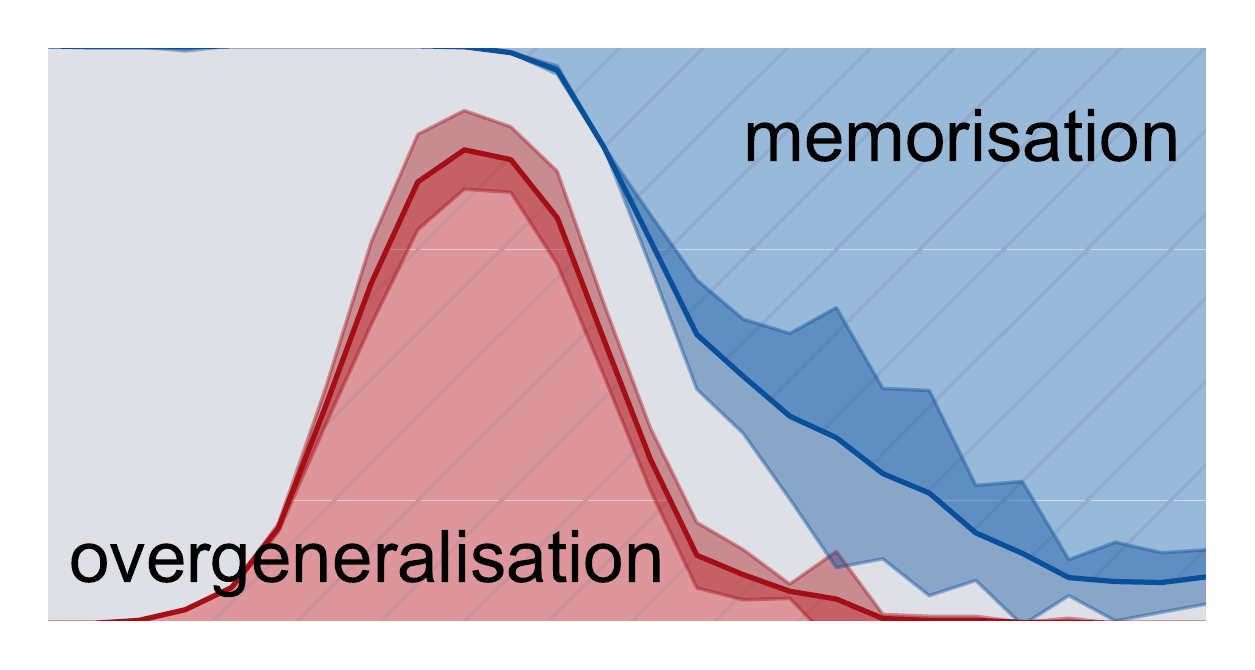}
    \end{subfigure}\\ 
    0.5 & \begin{subfigure}{0.318\linewidth}
        \includegraphics[width=\linewidth]{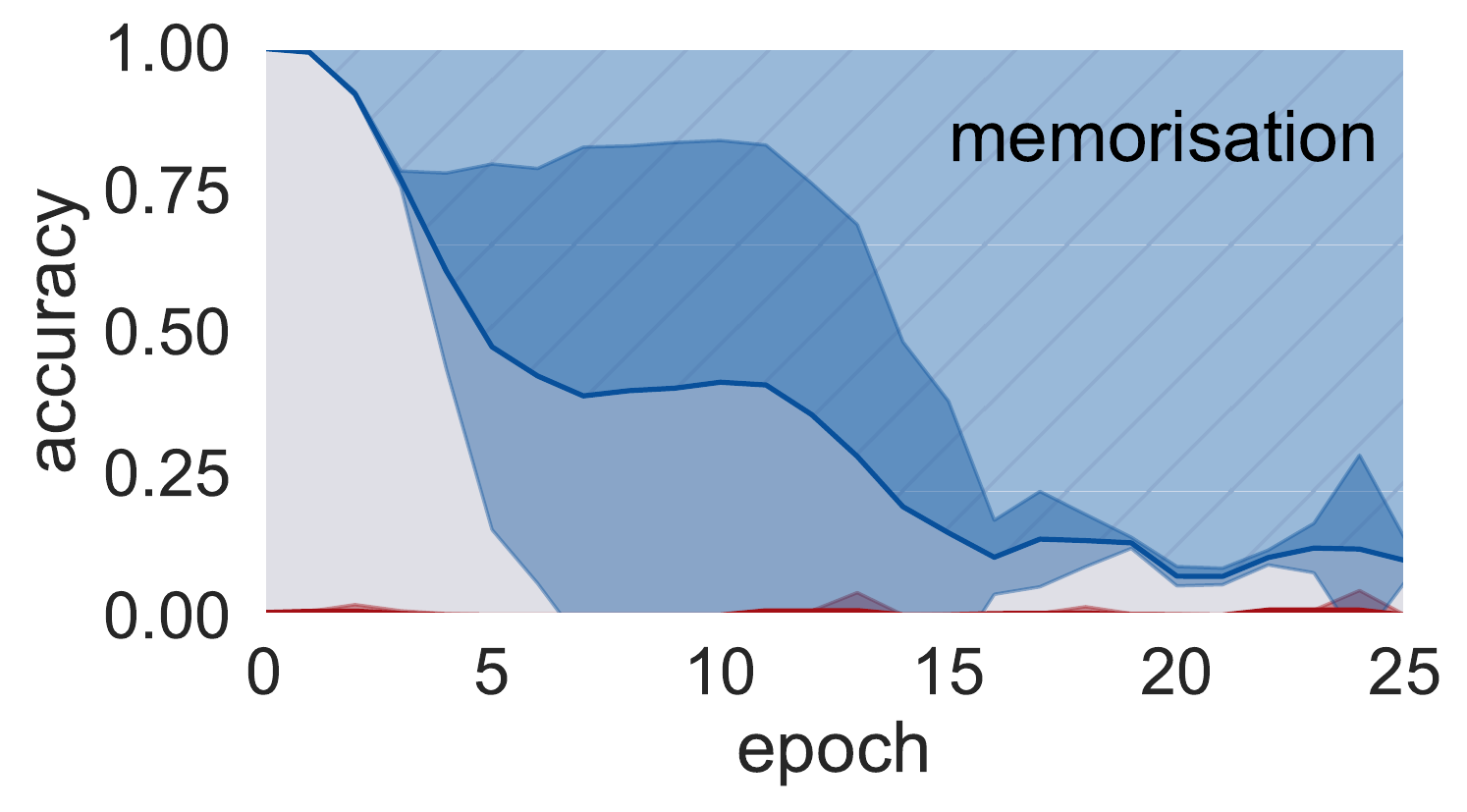}
    \end{subfigure} &
    \begin{subfigure}{0.27\linewidth}
        \includegraphics[width=\linewidth]{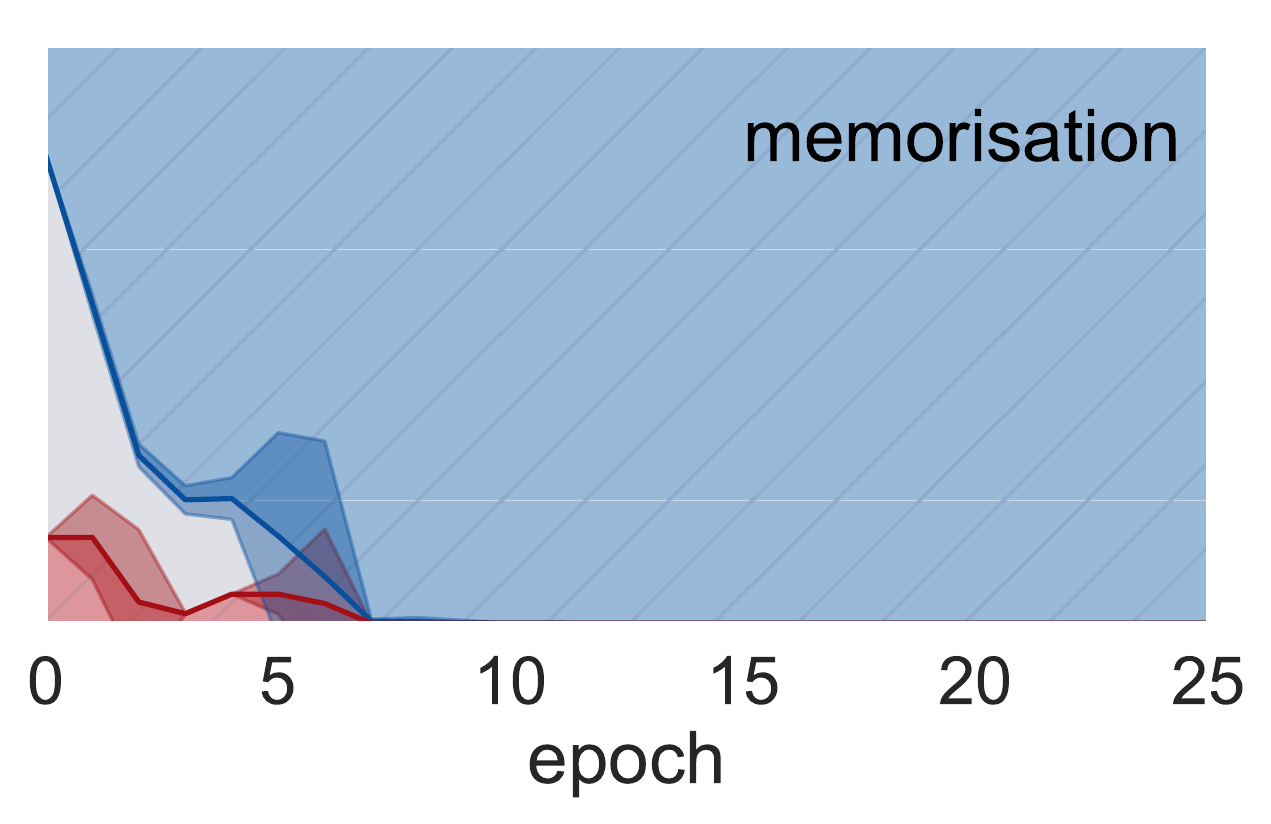}
    \end{subfigure} &
    \begin{subfigure}{0.27\linewidth}
        \includegraphics[width=\linewidth]{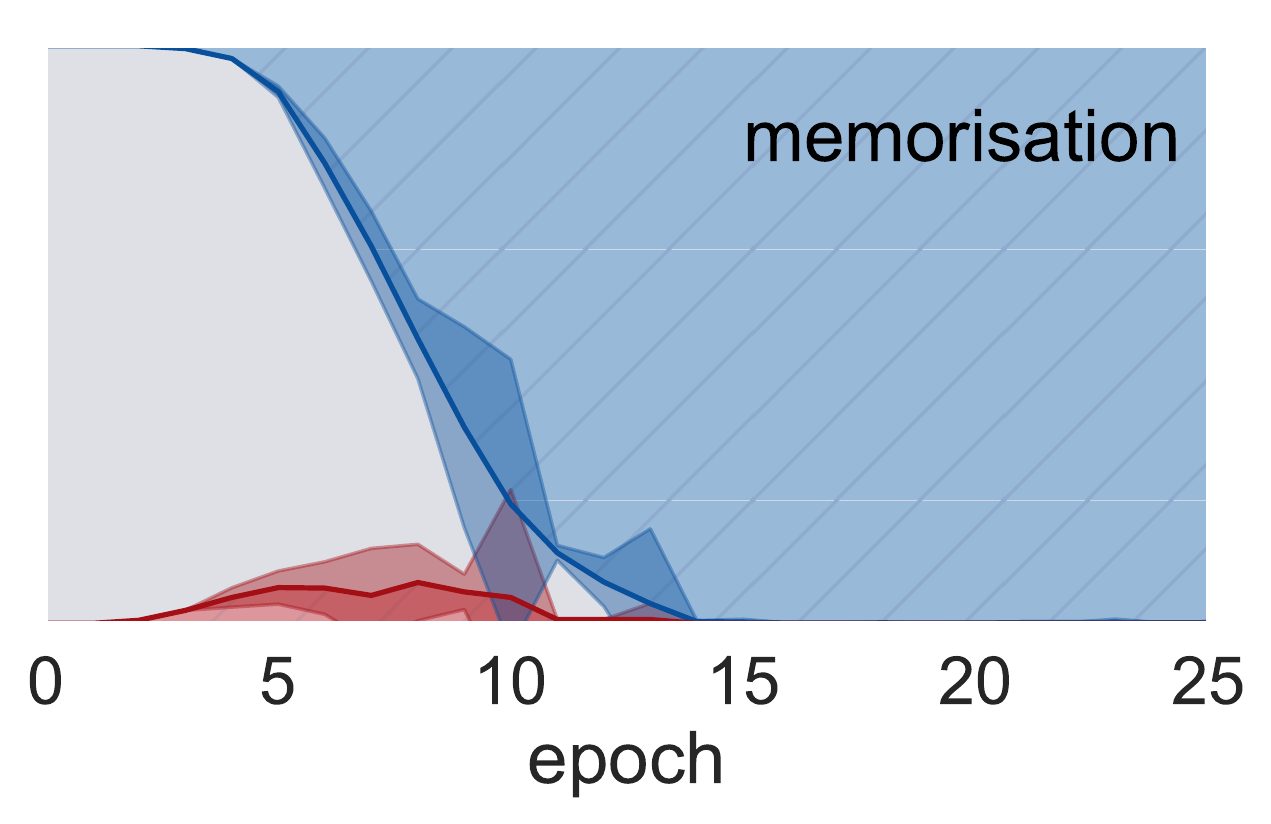}
    \end{subfigure}\\
    \end{tabular}
    \caption{Overgeneralisation profiles over time for LSTMS2S, ConvS2S and Transformer for exception percentages of 0.01\%, 0.05\%, 0.1\% and 0.5\% (in increasing order, from top to bottom).
The lower area of the plots, in red, indicates the mean fraction of exceptions (with standard deviation) for which an overgeneralised output sequence is predicted (i.e.\ not the `correct' exception output for the sequence, but the output that one would construct following the meaning of the functions as observed in the rest of the data).
We denote this area with `overgeneralisation'.
The upper areas, in blue, indicate the mean fraction of the exception sequences (with standard deviation) for which the model generates the true output sequence, which -- as it falls outside of the underlying compositional system -- has to be memorised.
We call this the `memorisation' area.
The grey area in between corresponds to the cases in which a model does not predict the correct output, nor the output that would be expected if the rule were applied.
}\label{fig:overgeneralisation_all}
\end{figure}

\paragraph{Overgeneralisation profile}
More interesting than the height of the peak is the \emph{profile} that different architectures show during learning.
In Figure~\ref{fig:overgeneralisation_all}, we plot this profile for four different exception percentages.
The lower areas (in red), indicate the overgeneralisation strength, whereas the memorisation strength -- the accuracy of a model on the adapted outputs, which can only be learned by memorisation -- is indicated in the upper part of the plots, in blue.
The grey area in between indicates the percentage of exception examples for which a model outputs neither the correct answer nor the rule-based answer.

\paragraph{Exception percentage}
The profiles show that, for all architectures, the degree of overgeneralisation strongly depends on the number of exceptions present in the data.
All architectures show overgeneralisation behaviour for exception percentages lower than 0.5\% (first three rows), but hardly any overgeneralisation is observed when
0.5\% of a function's occurrence is an exception (bottom row).
When the percentage of exceptions becomes too low, on the other hand, all models have difficulties memorising them at all: when the exception percentage is 
0.01\% of the overall function occurrence, only ConvS2S can memorise the correct answers to some extent (middle column, top row). 
LSTMS2S and Transformer keep predicting the rule-based output for the sequences containing exceptions, even after the training converged. 

\paragraph{Learning an exception}
LSTMS2S, in general, appears to find it difficult to accommodate both rules and exceptions at the same time.
Transformer and ConvS2S overgeneralise at the beginning of training, but then, once enough evidence for the exception is accumulated, gradually change to predicting the correct output for the exception sequences.
This behaviour is most strongly present for ConvS2S, as evidenced by the thinness of the grey stripe separating the red and the blue area during training.
For LSTMS2S, on the other hand, the decreasing overgeneralisation strength is not matched by an increasing memorisation strength.
After identifying that a certain sequence is not following the same rule as the rest of the corpus, LSTMS2S does not predict the correct meaning but instead starts generating outputs that match neither the correct exception output nor the original target for the sequence.
After convergence, its accuracy on the exception sequences is substantially lower than the overall corpus accuracy.
As the bottom plot (with an exception percentage of 0.5\%) indicates that LSTMS2S models do not have problems with learning exceptions per se, they appear to struggle with hosting exceptions for words if little evidence for such anomalous behaviour is present in the training data.

\section{Discussion}\label{sec:discussion}

With the rising successes of models based on deep learning, evaluating the compositional skills of neural network models has attracted the attention of many researchers.
Many empirical studies have been presented that evaluate the compositionality of neural models in different ways, but they have not led to a consensus about whether neural models can in fact adequately model compositional data.
We argue that this lack of consensus stems from a deeper issue than the results of the proposed tests: while many researchers have a strong intuition about what it means for a model to be compositional, there is no explicit agreement on what defines compositionality of a model or how it should be tested for in a neural model.

\subsection{An evaluation framework to evaluate compositionality}
In this paper, we proposed an evaluation framework that addresses this problem, with a series of tests that translate theoretical concepts related to compositionality of language into behavioural tests for models of language.
Our evaluation framework contains five independent tests that consider complementary aspects of compositionality that are frequently mentioned in the literature about compositionality. 
These five tests allow us to investigate (i) if models systematically recombine known parts and rules (\emph{systematicity}) (ii) if models can extend their predictions beyond the length they have seen in the training data (\emph{productivity}) (iii) if models' predictions are robust to synonym substitutions (\emph{substitutivity}) (iv) if models' composition operations are local or global (\emph{localism}) and (v) if models favour rules or exceptions during training (\emph{overgeneralisation}).
We formulated these tests on a task-independent level, disentangled from a specific downstream task.
With this, we offer a versatile evaluation paradigm which can be used to evaluate the compositional abilities of a model on five different levels, that can be instantiated for any chosen sequence-to-sequence task.
Importantly, our collection of tests should not be taken as a normative specification of what models should and should not do.
Rather, they are meant to discover which aspects of compositionality a model does or does not implement and learn more about a model's strengths and weaknesses.

To showcase our evaluation paradigm, we instantiated the five tests on a highly compositional artificial data set we dub \pcfg: a sequence-to-sequence translation task which requires to compute meanings of sequences that are generated by a probabilistic context-free grammar by recursively applying string edit operations.
This data set is designed such that modelling it adequately should require a compositional solution, and it is generated such that its length and depth distributions match those of a natural corpus of English.
We then used these instantiated tests to compare three popular sequence-to-sequence architectures: an LSTM-based (\emph{LSTMS2S}), a convolution-based (\emph{ConvS2S}) and an all-attention model (\emph{Transformer}).
For each test, we conducted a number of auxiliary tests that can be used to further increase the understanding of how this aspect is treated by a particular architecture.
Below, we provide a summary of the results of these experiments.\footnote{
At the risk of being redundant, we repeat that these results should not be taken as general claims about LSTMs, convolutional networks or transformers.
Neural models can be sensitive to small changes in hyper-parameters and learning regimes and we did not investigate the effect of changing the hyper-parameters.
Perhaps using a deeper LSTM, a transformer with more attention heads, or convolutions with a wider kernel width would show different patterns.
We leave these questions open for future work.
With the results below, we merely want to show that -- keeping the tests fixed -- interesting differences but also similarities between different models can be found.}

\subsection{Summary of results}
While the overall accuracy on \pcfg was relatively high for all models, a more detailed picture is given by the five compositionality tests.
These tests indicated that, despite our careful data design, high scores do still not necessarily imply that the trained models fully represent the true underlying generative system and illustrated how different models handle different aspects that could be considered important for compositional learning.

Firstly, our \textbf{systematicity} test showed that none of the architectures successfully generalises to pairs of words that were not observed together during training, a result that confirms earlier studies such as the ones from \citet{loula2018rearranging} and \citet{lake2018generalization}. 
The difference between the systematicity scores and the overall task accuracy is quite stark for all models: a drop of 33\%, 34\% and 22\% for LSTMS2S, ConvS2S and Transformer, respectively.
This suggests that the low accuracy on the systematicity test does not stem from poor systematic capacity in general, but that rather from the fact that the models use different segmentations of the input, applying -- for instance -- multiple functions at once, instead of all of the functions in a sequential manner.
While larger chunking to ease processing is not necessarily a bad strategy, it is desirable if models can also maintain a separate representation of the units that make up such chunks, as these units could be useful or needed in other sequences.

With our \textbf{productivity} test, we assessed if models can productively generalise to sequences that are longer than the ones they observed in training.
To evaluate this, we redistributed the training examples such that there is a strict separation of the input sequence lengths in the train and test data.
To tease apart the overall difficulty of modelling longer sequences from the ability to generalise to unseen lengths, we compared the results with the accuracies of models that are trained on data sets that contain at least some evidence for longer sequences.
None of the architectures exhibited strong productive power to sequences of unseen lengths.
By computing how often models' predictions were strictly contained within the true output sequence, we assess if the poor productive power of all models is caused by early emission of the end-of-sequence symbol.
We find that such cases indeed exist, but that early stopping of the generation is not the main cause of the low productivity scores. 

With our \textbf{substitutivity} test, we compared how models react to artificially introduced synonyms occurring in different types of scenarios.
Rather than considering their behaviour in terms of sequence accuracy, in this test, we computed how \textit{consistent} models' predictions are -- correct or incorrect -- when a word is substituted with a synonym.
When synonyms are equally distributed in the input data, both Transformer and ConvS2S obtain high consistency scores, while LSTMS2S is substantially less consistent.
This difference is also reflected in the distance between the embeddings of words and synonyms, which is much lower for Transformer and ConvS2S.
When one of the synonyms is only presented in a few very short sequences, the consistency score of ConvS2S drops to the same level as LSTMS2S, while Transformer still maintains a relatively high synonym consistency.
Also the embeddings of synonyms remain relatively close in Transformer models' embedding space, despite the fact that they are distributionally dissimilar.

To tease apart the ability to learn from very few examples and to infer synonymity, we also considered how consistent models are on \textit{incorrect} outputs.
Here, we observed that none of the models can be said to truly treat words and their counterparts as synonyms.
Transformer is the most consistent, but with a low score of only 0.34.
This test shows an interesting difference between LSTMS2S and ConvS2S: where the former appears to be better at inferring that words are synonyms, the latter is better at few-shot learning a word's meaning from very few examples.

With our \textbf{localism} test, we considered if models apply local composition operations that are true to the syntactic tree of an input sequence, or rather compute the meaning of a sequence in a more global fashion.
In line with the results of the systematicity test, models do not appear to truly follow the syntactic tree of the input to compute its meaning.
In 54\%, 41\% and 46\% of the test samples for LSTMS2S, ConvS2S and Transformer, respectively, enforcing a local computation results in a different answer than the original answer provided by the model.
An error analysis suggests that these results are largely due to function applications to longer string sequences.
With an additional test in which we monitored the accuracy of models on functions applied to increasingly long string inputs, we find evidence that models may not learn general-purpose representations of functions, but instead use different protocols for \emph{copy once} or \emph{copy twice}.
We saw that the accuracy of LSTMS2S immediately drops to 0 when string inputs are longer than the ones observed in training.
The performance of ConvS2S and Transformer, instead, drops rapidly, but remains above 0 for slightly longer string inputs.
These results indicate that LSTMS2S may indeed not have learned a general-purpose representation for functions, while the decreasing accuracy of ConvS2S and Transformer could be related more to performance rather than competence issues.

In our last test, we studied \textbf{overgeneralisation} during training, by monitoring the behaviour of models on artificially introduced \textit{exceptions} to rules for four function pairs.
We found that for small amounts of exceptions (up to 0.1\% of the number of occurrences of the least occurring function of a function pair) all architectures overgeneralise at the beginning of their training.
As overgeneralisation implies that models overextend rules in cases where this is explicitly contradicted by the data, we take this as a clear indication that models in fact capture the underlying rule at that point.
For very small amounts of exceptions (0.01\%), both Transformer and LSTMS2S failed to learn the exception at all: even after their training has converged they overgeneralise on the sequences containing exceptions.
To a lesser extent, also ConvS2S struggles with capturing exceptions that have a low frequency.
LSTMS2S generally appears to have difficulty with accommodating both rules and exceptions.
Often, after learning that a certain rule should not be applied, LSTMS2S models do not memorise the true target but proceed to predict something which matches neither this target nor the general rule.
ConvS2S and Transformer do not show such patterns: when their \emph{overgeneralisation} score goes down, their \emph{memorisation} score goes up.
Aside from at the beginning of their training, they rarely predict something outside of these options.
For larger percentages of exceptions (from 0.5\%), none of the architectures really exhibits overgeneralisation behaviour.

\subsection{Conclusion and future work}
With a proposed collection of tests, we aimed to cover several facets of compositionality.
We believe that as such, this collection of tests can serve as an evaluation paradigm to probe the ability of different neural network architectures in the light of compositionality.
We hope that the tests and their results can help facilitate a general discussion of what it means for neural models to be compositional and what we would like them to represent.
There are, of course, also aspects of compositionality that we did not cover.
We therefore do not consider our evaluation an endpoint, but rather a stepping stone on the way, which we hope can provide the grounds for a clearer discussion concerning the role and importance of compositionality in neural networks, including both aspects that we did and did not include.

We instantiated our tests on an artificial data set that is entirely explainable in terms of compositional phenomena.
This permitted us to focus on the compositional ability of different models in the face of compositional data and allowed us to isolate compositional processing from other signals that are found in more realistic data sets.
However, it leaves open the question of how much the compositional traits we identified are expressed and can be exploited by networks when facing natural data.
Despite the fact that they are not informed by knowledge of language or semantic composition, neural networks have achieved tremendous successes in almost all natural language processing tasks.
While their performance is still far from perfect, it is not evident that their remaining failures stem from their inability to deal with compositionality.
In the future, we plan to instantiate our tests also in natural language domains such as translation and summarisation.
The results of such a study would provide valuable information about how well models pick up compositional patterns in more noisy environments, but -- perhaps even more importantly -- could also provide insights about the importance of these different aspects of compositionality to model natural data.

In summary, we provided an evaluation paradigm that allows a researcher to test the extent to which five distinct, theoretically motivated aspects of compositionality are represented by artificial neural networks.
By instantiating these tests for an artificial data set and applying the resulting tests on three different successful sequence-to-sequence architectures, we shed some light on which aspects of compositionality may prove problematic for different architectures.
These results illustrate well that to test for compositionality in neural networks it does not suffice to consider an accuracy score on a single downstream task, even if this task is designed to be highly compositional.
Models may capture some compositional aspects of this data set very well, but fail to model other aspects that could be considered part of a compositional behaviour.
As such, the results themselves demonstrate the need for the more extensive set of evaluation criteria that we aim to provide with this work.
We hope that future researchers will use our collection of tests to evaluate new models, to investigate the impact of hyper-parameters or to study how compositional behaviour is acquired during training.
To facilitate the usage of our test suite we have made the \pcfg data generator, all test sets and the models trained by us available online.\footnote{\url{https://github.com/i-machine-think/am-i-compositional}}
We further hope that our theoretical motivation, the tests themselves and the analysis that we presented of its application on three different sequence-to-sequence architectures will prove to be a step in the direction of having a clearer discussion about compositionality in the context of deep learning, both from a practical and a theoretical perspective.

\section*{Acknowledgments}

We thank Marco Baroni, Yoav Goldberg, Aureli Herbelot, Louise McNally, Ryan Nefdt, Sandro Pezzelle, Shane Steinert-Threlkeld and Willem Zuidema for taking the time to proofread earlier versions of this paper and giving us feedback.
Furthermore, we thank our anonymous reviewers and editor Stephen Clark for their interesting comments and helpful suggestions.

Dieuwke Hupkes is funded by the Netherlands Organization for Scientific Research (NWO), through a Gravitation Grant 024.001.006 to the Language in Interaction Consortium.
Elia Bruni is funded by the European Union's Horizon 2020 research and innovation program under the Marie Sklodowska-Curie grant agreement No 790369 (MAGIC).

\bibliography{manifesto}

\newpage
\appendix
% Supplementary materials

\section{Naturalisation of artificial data}
\label{appendix:naturalisation}

The artificially generated PCFG SET data are transformed so as to mimic the distribution of a natural language data set according to the following procedure:

\begin{enumerate}

\item Use a natural language data set $\mathcal{D}_N$, define a set of features $F$, and for each $f \in F$, compute the value $f(s)$ for each sentence $s \in \mathcal{D}_N$.

\item Generate a large sample $\mathcal{D}_R$ of PCFG SET data using random probabilities on production rules for each instance. 

\item Transform $\mathcal{D}_R$ as follows:

\begin{enumerate}[label=(\roman*)]
    \item For each feature $f \in F$, specify a \textit{feature increment} $i_f$. 
    
    \item For each $s \in \mathcal{D}_N$, compute the \textit{partitioning vector} $v(s)$, which is the concatenation of the values $\left \lfloor{f(s)} / i_f \right \rfloor$ for each feature $f \in F$.
    
    \item Partition $\mathcal{D}_N$ into subsets by clustering instances with the same partitioning vector. For any such subset $\mathcal{D}_N^i$, let $v(\mathcal{D}_N^i)$  denote the partitioning vector of its members. And for any partitioning vector $\mathbf{v}$, let $v_N^{-1}(\mathbf{v})$ denote the subset $\mathcal{D}_N^i \subseteq \mathcal{D}_N$ whose members have partitioning vector $\mathbf{v}$ (so that $v(\mathcal{D}_N^i ) = \mathbf{v}$).
    
    \item Of the identified subsets, determine the largest set $\mathcal{D}_N^i \subseteq \mathcal{D}_N$. Call this set $\mathcal{D}_N^{\max}$. 
    
    \item Partition $\mathcal{D}_R$ in the same way as $\mathcal{D}_N$, yielding subsets $\mathcal{D}_R^i$. Let the subset $\mathcal{D}_R^i$ such that $v(\mathcal{D}_R^i) = v(\mathcal{D}_N^{\max})$ be $\mathcal{D}_R^{\max}$.
    
    \item Initialise an empty set $\mathcal{D}_{R}'$. 
    
    \item Of each $\mathcal{D}_R^i$, randomly pick 
$\frac{\mid v_N^{-1}(v(\mathcal{D}_R^i)) \mid \times \mid \mathcal{D}_R^{\max}\mid }{\mid \mathcal{D}_N^{\max} \mid}$ members, and assign them to $\mathcal{D}_{R}'$.
    
    \item If necessary, repeat (i) - (vii) for different feature increments $f_i$. For $n$ features, fit an $n$-variate Gaussian to each of the transformed sets $\mathcal{D}_{R}'$. Choose the set with the lowest Kullback-Leibler divergence from the $n$-variate Gaussian approximation of $\mathcal{D}_N$.  

\end{enumerate}

\item Use maximum likelihood estimation to estimate the PCFG parameters of $\mathcal{D}_{R}'$ and generate more PCFG SET data using these parameters. 

\item If necessary, apply step 3 to the data thus generated. 

\end{enumerate}

\end{document}